%% file: _main.tex
\documentclass[fleqn,12pt]{wlscirep}
\usepackage[utf8]{inputenc}
\usepackage[T1]{fontenc}
\usepackage{setspace}

\usepackage{array}
\newcolumntype{L}[1]{>{\raggedright\let\newline\\\arraybackslash\hspace{0pt}}m{#1}}
\newcolumntype{C}[1]{>{\centering\let\newline\\\arraybackslash\hspace{0pt}}m{#1}}
\newcolumntype{R}[1]{>{\raggedleft\let\newline\\\arraybackslash\hspace{0pt}}m{#1}}

\usepackage{caption}
\usepackage{subcaption}
\usepackage{url}

\usepackage{breakurl}
\input{_commands}
\input{_articleinfo}

\begin{document}

\flushbottom
\maketitle

\input{00_summary}
\input{01_introduction}
\input{02_results}
\input{03_discussion}
\input{04_methods}
\input{05_acknowledgements}

\bibliography{_references}

\input{06_supplementary}



\end{document}

%% file: _commands.tex
\newcommand{\xhdr}[1]{\vspace{1.7mm}\noindent{{\bf #1.}}}
\newcommand{\xhdritem}[1]{\noindent{{\bf #1.}}}
\newcommand{\Partner}{\textsc{Partner}}
\newcommand{\oursystem}{\textsc{Hailey}}



%% file: _articleinfo.tex
\title{Human-AI Collaboration Enables More Empathic Conversations in Text-based Peer-to-Peer Mental Health Support}

\author[1]{Ashish Sharma}
\author[1]{Inna W. Lin}
\author[2,3]{Adam S. Miner}
\author[4]{David C. Atkins}
\author[1,*]{Tim Althoff}
\affil[1]{Paul G. Allen School of Computer Science and Engineering, University of Washington, Seattle, WA, USA}
\affil[2]{Department of Psychiatry and Behavioral Sciences, Stanford University, Stanford, CA, USA}
\affil[3]{Center for Biomedical Informatics Research, Stanford University, Stanford, CA, USA}
\affil[4]{Department of Psychiatry and Behavioral Sciences, University of Washington, Seattle, WA, USA}
\affil[*]{althoff@cs.washington.edu}


%% file: 00_summary.tex
\section*{Abstract}


Advances in artificial intelligence (AI) are enabling systems that augment and collaborate with humans to perform simple, mechanistic tasks like scheduling meetings and grammar-checking text. However, such Human-AI collaboration poses challenges for more complex, creative tasks, such as carrying out empathic conversations, due to difficulties of AI systems in understanding complex human emotions and the open-ended nature of these tasks. Here, we focus on peer-to-peer mental health support, a setting in which empathy is critical for success, and examine how AI can collaborate with humans to facilitate peer empathy during textual, online supportive conversations. We develop \oursystem, an AI-in-the-loop agent that provides just-in-time feedback to help participants who provide support (\textit{peer supporters}) respond more empathically to those seeking help (\textit{support seekers}). We evaluate \oursystem~in a non-clinical randomized controlled trial with real-world peer supporters on TalkLife (N=300), a large online peer-to-peer support platform. We show that our Human-AI collaboration approach leads to a 19.60\% increase in conversational empathy between peers overall. Furthermore, we find a larger 38.88\% increase in empathy within the subsample of peer supporters who self-identify as experiencing difficulty providing support. We systematically analyze the Human-AI collaboration patterns and find that peer supporters are able to use the AI feedback both directly and indirectly without becoming overly reliant on AI while reporting improved self-efficacy post-feedback. Our findings demonstrate the potential of feedback-driven, AI-in-the-loop writing systems to empower humans in open-ended, social, creative tasks such as empathic conversations. 



%% file: 01_introduction.tex
\section*{Introduction}

As artificial intelligence (AI) technologies continue to advance, AI systems have started to augment and collaborate with humans in application domains ranging from e-commerce to healthcare\cite{rajpurkar2022ai,hosny2019artificial,Patel2019-td,Tschandl2020-nk,Cai2019-ov,Suh2021-yl,Wen2017-cu,baek2021accurate,jumper2021highly}. In many and especially in high-risk settings, such Human-AI collaboration has proven  more robust and effective than totally replacing humans with AI\cite{Verghese2018-uo,Bansal2021-sr}. However, the collaboration faces dual challenges of developing human-centered AI models to assist humans and designing human-facing interfaces for humans to interact with the AI\cite{Yang2020-nl,Li2020-yt,Gillies2016-vh,Amershi2019-sf,Norman1994-om,Hirsch2017-wp}. For AI-assisted writing, for instance, we must build AI models that generate actionable writing suggestions \textit{and} simultaneously design human-facing systems that help people see, understand and act on those suggestions just-in-time\cite{Clark2018-mi,Roemmele2018-ww,Lee2022-oj,noauthor_undated-tt,Buschek2021-yo,Gero2021-ox,Hirsch2017-wp}. Therefore, current Human-AI collaboration systems have been restricted to simple, mechanistic tasks, like scheduling meetings, checking spelling and grammar, and booking flights and restaurants. Though initial systems have been proposed for tasks like story writing\cite{Clark2018-mi} and graphic designing\cite{Chilton2019-xp}, it remains challenging to develop Human-AI collaboration for a wide range of complex, creative, open-ended, and high-risk tasks.

In this paper, we focus on text-based, peer-to-peer mental health support and investigate how AI systems can collaborate with humans to help facilitate the expression of \textit{empathy} in textual supportive conversations -- in this case by intervening on people in the conversations who provide support (\textit{peer supporters}) as against those who seek support (\textit{support seekers}). \textit{Empathy} is the ability to understand and relate to the emotions and experiences of others and to effectively communicate that understanding\cite{Elliott2011-qr}.  Empathic support is one of the critical factors (along with ability to listen, concrete problem solving, motivational interviewing, etc.) that contributes to successful conversations in mental health support, showing strong correlations with symptom improvement\cite{Elliott2018-na} and the formation of alliance and rapport\cite{Bohart2002-vv,Elliott2011-qr,Watson2002-zx,Sharma2020-dx}. While online peer-to-peer platforms like TalkLife (\href{https://www.talklife.com/}{talklife.com}) and Reddit (\href{https://www.reddit.com/}{reddit.com}) enable such supportive conversations in non-clinical contexts, highly empathic conversations are rare on these platforms\cite{Sharma2020-dx}; peer supporters 
are typically untrained in expressing complex, creative, and open-ended 
skills like empathy\cite{Davis1980-qw,Blease2020-of,Doraiswamy2020-dk,Riess2017-jf} and may lack the required expertise. With an estimated 400 million people suffering from mental health disorders worldwide\cite{noauthor_undated-ra}, combined with a pervasive labor shortage\cite{Kazdin2011-de,Olfson2016-xg}, these platforms have pioneered avenues for seeking social support and discussing mental health issues for millions of people\cite{Naslund2016-dj}. However, the challenge lies in improving conversational quality by encouraging untrained peer supporters to adopt complicated and nuanced 
skills like empathy. 


As shown in prior work\cite{kemp2012challenges,mahlke2014peer}, untrained peer supporters report difficulties in writing supportive, empathic responses to support seekers. Without deliberate training or specific feedback, the difficulty persists over time\cite{Schwalbe2014-ai,Goldberg2016-lu,Sharma2020-dx} and may even lead to a gradual decrease in supporters' effectiveness due to factors such as empathy fatigue\cite{Nunes2011-cz,Hojat2009-qc,Stebnicki2007-qs}. Furthermore, current efforts to improve empathy (e.g., in-person empathy training) do not scale to the millions of peer supporters providing support online. Thus, empowering peer supporters with automated, actionable, just-in-time feedback and training, such as through Human-AI collaboration systems, can help them express higher levels of empathy and, as a result, improve the overall effectiveness of these platforms\cite{Imel2015-sa,Miner2019-ri,Sharma2020-dx, Sharma2021-rq}.




To this end, we develop and evaluate a Human-AI collaboration approach for helping untrained peer supporters write more empathic responses in online, text-based peer-to-peer support. We propose \oursystem~(\textsc{H}uman-\textsc{AI} co\textsc{L}laboration approach for \textsc{E}mpath\textsc{Y}), an AI-in-the-loop agent that offers just-in-time suggestions to express empathy more effectively in conversations (Figure~\ref{figure:1}b, \ref{figure:1}c). We design \oursystem~to be collaborative, actionable and mobile friendly (Methods).

Unlike the AI-only task of empathic dialogue generation (generating empathic responses from scratch)\cite{Lin2019-zw,Majumder2020-bo,Rashkin2018-tp}, we adopt a collaborative
design that edits existing human responses to make them more empathic\cite{Sharma2021-rq}. This design reflects the high-risk setting of mental health, where AI is likely best used to augment, rather than replace, human skills\cite{Miner2019-ri,Chen2017-en}. Furthermore, while current AI-in-the-loop systems are often restricted in the extent to which they can guide humans (e.g., simple classification methods that tell users to be empathic when they are not)\cite{Tanana2019-bj,Peng2020-sg,Hui2018-ty,Kelly2018-fo}, we ensure actionability by guiding peer supporters with concrete steps they may take to respond with more empathy, e.g., through the \textit{insertion} of new empathic sentences or \textit{replacement} of existing low-empathy sentences with their more empathic counterparts (Figure~\ref{figure:1}c). For complex, hard-to-learn skills like empathy, this enables just-in-time suggestions on not just ``what'' to improve but on ``how'' to improve it. 

\input{figures/_figure_1}

We consider the general setting of text-based, asynchronous conversations between a support seeker and a peer supporter (Figure~\ref{figure:1}). In these conversations, the support seeker authors a post for seeking mental health support (e.g., ``\textit{My job is becoming more and more stressful with each passing day}.'') to which the peer supporter writes a supportive response (e.g., ``\textit{Don't worry! I'm there for you.}''). In this context, we support the peer supporters by providing just-in-time AI feedback to improve the empathy of their responses. To do so, \oursystem~prompts the peer supporter through a pop-up (``\textit{Would you like some help with your response?}'') placed above the response text box. On clicking this prompt, \oursystem~shows just-in-time AI feedback consisting of \textit{Insert} (e.g., Insert ``\textit{Have you tried talking to your boss?}'' at the end of the response) and \textit{Replace} (e.g., Replace ``\textit{Don't worry!}'' with ``\textit{It must be a real struggle!}'') suggestions based on the original seeker post and the current peer supporter response. The peer supporter can incorporate these suggestions by directly clicking on the appropriate Insert or Replace buttons, by further editing them, and/or by deriving ideas from the suggestions to indirectly use in their response. 

To evaluate \oursystem, we conducted a randomized controlled trial in a non-clinical, ecologically valid setting 
with peer supporters from a large peer-to-peer support platform, {TalkLife} (\href{https://www.talklife.com/}{talklife.com}) as participants (N=300; Supplementary Table~\ref{supp:tab:rct}). Our study was performed outside the TalkLife platform to ensure platform users' safety 
but adopted an interface similar to TalkLife's chat feature (Figure~\ref{figure:1}; Methods). We employed a between-subjects study design, where each participant was randomly assigned to one of two conditions: Human + AI (treatment; with feedback) or Human Only (control; without feedback). Before the main study procedure of writing supportive, empathic responses (Methods), participants in both groups received basic training on empathy, which included empathy definitions, frameworks, and examples (Supplementary Figure~\ref{supp:figure:empathy_training}).

While peer supporters on these platforms do not typically receive empathy training, we trained both Human + AI (treatment) and Human Only (control) groups just before the main study procedure of writing supportive, empathic responses; this let us conservatively estimate the effect of just-in-time feedback beyond traditional, offline feedback or training (Discussion). During the study, each participant was asked to write supportive, empathic responses to a unique set of 10 existing seeker posts (one at a time) that were sourced at random from a subset of TalkLife posts; we filtered out posts related to critical settings of suicidal ideation and self-harm to ensure participant safety (Methods; Discussion). While writing responses, participants in the Human + AI (treatment) group received feedback via \oursystem~(Figure~\ref{figure:1}b, \ref{figure:1}c). Participants in the Human Only (control) group, on the other hand, wrote responses but received no feedback, reflecting the current status quo on online peer-to-peer support platforms (Figure~\ref{figure:1}a). After completing responses to the 10 posts, participants were asked to assess \oursystem~by answering questions about the challenges they experienced while writing responses and the effectiveness of our approach.

Our primary hypothesis was that Human-AI collaboration would lead to more empathic responses, i.e., responses in the Human + AI (treatment) group would show higher empathy than the Human Only (control) group responses. We evaluated  this hypothesis using both human and automatic evaluation, which helped us capture platform users' perceptions and provided a theory-based assessment of empathy in the collected responses respectively (Methods). Note that due to the sensitive mental health context and for reasons of safety, our evaluation of empathy was only based on empathy that was \textit{expressed} in responses and not the empathy that might have been \textit{perceived} by the support seeker of the original seeker post\cite{Barrett-Lennard1981-ji}. Psychotherapy research indicates a strong correlation between expressed empathy and positive therapeutic outcomes and commonly uses it as a credible alternative\cite{Elliott2011-qr} (Methods; Discussion).

Furthermore, we conducted multiple post hoc evaluations to assess whether the participants who self-reported challenges in writing supportive responses could benefit more from our system, to investigate the differences in how participants collaborated with the AI, and to assess the participants' perceptions of our approach.

%% file: figures/_figure_1.tex
\begin{figure}
    \caption{We performed a randomized controlled trial with 300 TalkLife peer supporters as participants. We randomly divided participants into Human Only (control) and Human + AI (treatment) groups and asked them to write supportive, empathic responses to seeker posts without feedback and with feedback, respectively. 
    To identify whether just-in-time Human-AI collaboration helped increase expressed empathy beyond potential (but rare) traditional training methods, 
    participants in both groups received initial empathy training before starting the study (Methods; Supplementary Figure~\ref{supp:figure:empathy_training}). \textbf{(a)} Without AI, human peer supporters are presented with an empty chatbox to author their response (the current status quo). As peer supporters are typically untrained on best-practices in therapy -- such as empathy -- they rarely conduct highly empathic conversations. \textbf{(b)} Our feedback agent (\oursystem) prompts peer supporters for providing just-in-time AI feedback as they write their responses. \textbf{(c)} \oursystem~then suggests changes that can be made to the response to make it more empathic. These suggestions include new sentences that can be \textit{inserted} and options for \textit{replacing} current sentences with their more empathic counterparts. Participants can accept these suggestions by clicking on the \textit{Insert} and \textit{Replace} buttons and continue editing the response or get more feedback, if needed.}
    \centering
         \includegraphics[width=\textwidth]{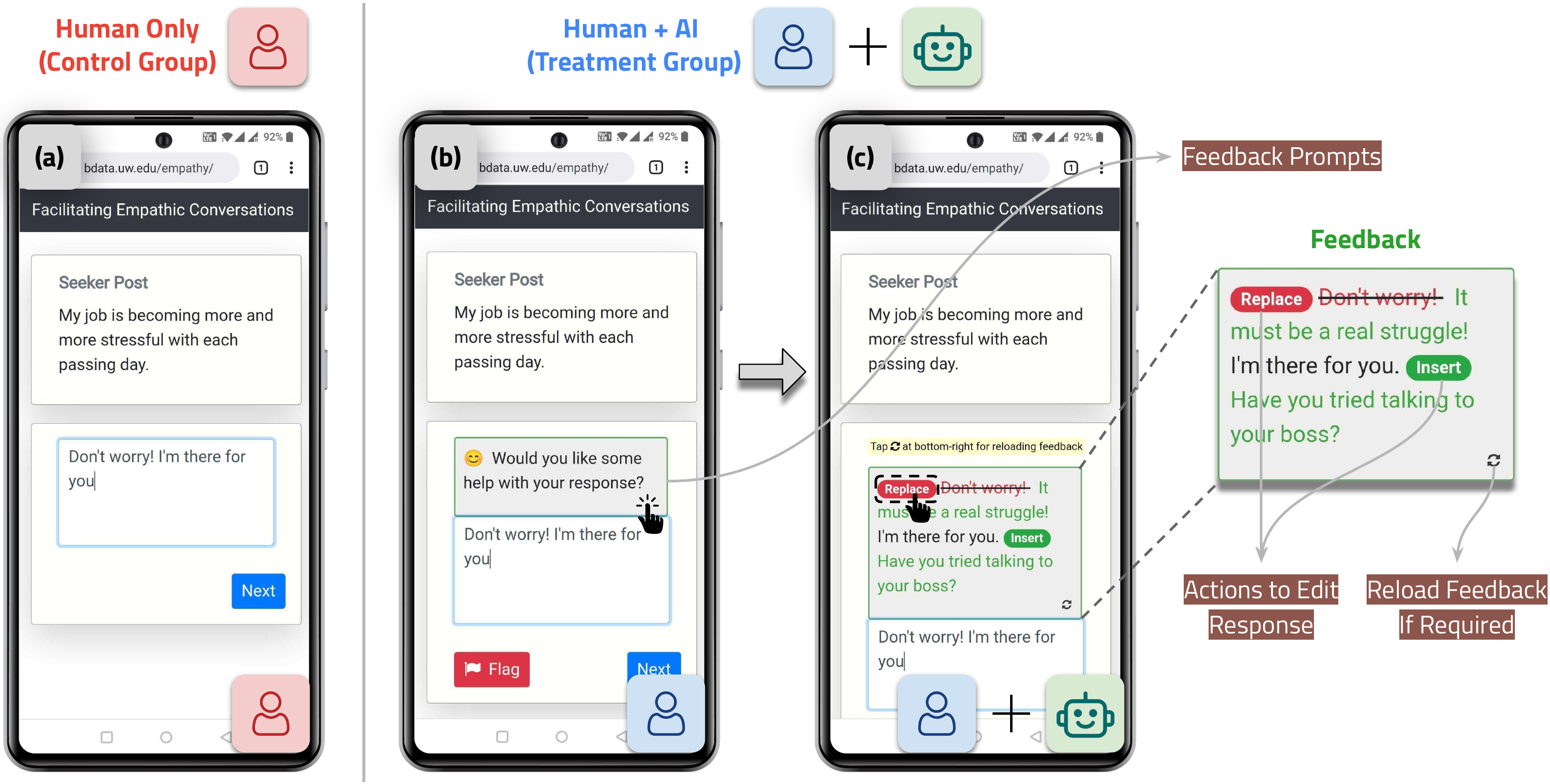}
    \label{figure:1}
\end{figure}

%% file: 02_results.tex
\section*{Results}

\subsection*{Increase in expressed empathy due to the Human-AI collaboration}
Our primary finding is that providing just-in-time AI feedback to participants leads to more empathic responses (Figure~\ref{figure:2}). Specifically, through human evaluation from an independent set of TalkLife users (Methods), we found that the Human + AI responses were rated as being more empathic than the Human Only responses 46.87\% of the time and were rated equivalent in empathy to Human Only responses 15.73\% of the time, whereas the Human Only responses were preferred only 37.40\% of the time (p < 0.01; Two-sided Student's t-test; Figure~\ref{figure:2}a). In addition, by automatically estimating empathy levels of responses using a previously validated empathy classification model on a scale from 0 to 6 (Methods), we found that the Human + AI approach led to 19.60\% higher empathic responses compared to the Human Only approach (1.77 vs. 1.48; p < $10^{-5}$; Two-sided Student's t-test; Figure~\ref{figure:2}b).

\input{figures/_figure_2}

\subsection*{Significantly higher gains for participants who reported challenges in writing responses and participants with little to no experience with peer support}
Prior work has shown that online peer supporters find it extremely challenging to write supportive and empathic responses\cite{kemp2012challenges,mahlke2014peer,Sharma2020-dx}. Some participants have little to no prior experience with peer support (e.g., if they are new to the platform; N=95/300; Methods). Even as the participants gain more experience, in the absence of explicit training or feedback, the challenge of writing supportive responses persists over time and may even lead to a gradual decrease in empathy levels due to factors such as empathy fatigue\cite{Schwalbe2014-ai,Goldberg2016-lu,Stebnicki2007-qs,Nunes2011-cz,Hojat2009-qc}, as also observed during the course of our 30-minute study (Supplementary Figure~\ref{supp:figure:empathy_over_time}). Therefore, it is particularly important to better assist the many participants who struggle with writing responses. 

For the subsample of participants who self-reported challenges in writing responses at the end of our study (N=91/300; Methods), a post hoc analysis revealed significantly higher empathy gains using the Human-AI collaboration approach. For such participants, we found a 4.50\% stronger preference for the Human + AI responses (49.12\% vs. 44.62\%; p < 0.01; Two-sided Student's t-test; Figure~\ref{figure:2}c) and a 27.01\% higher increase in expressed empathy using the Human + AI approach (38.88\% vs. 11.87\%; p < $10^{-5}$; Two-sided Student's t-test; Figure~\ref{figure:2}d) compared to participants who did not report any challenges. For the subsample of participants who self-reported no previous experience with online peer support at the start of our study (N=95/300; 37 of these participants also self-reported challenges), we found a 8.14\% stronger preference for the Human + AI responses (51.82\% vs. 43.68\%; p < 0.01; Two-sided Student's t-test;) and a 21.15\% higher increase in expressed empathy using the Human + AI approach (33.68\% vs. 12.53\%; p < $10^{-5}$; Two-sided Student's t-test; Supplementary Figure~\ref{supp:figure:background_empathy}d) compared to participants who reported experience with online peer support.


\subsection*{Hierarchical taxonomy of Human-AI collaboration strategies reveals key AI consultation and usage patterns}
The collaboration between humans and AI can take many forms since humans can apply AI feedback in a variety of ways depending on their trust in the shown feedback\cite{Bansal2021-sr}. Investigating how humans collaborate with our AI can help us better understand the system’s use-cases and inform better design decisions. Here, we analyzed collaboration patterns of participants both over the course of the study as well as during a single response instance. We leveraged this analysis to derive a hierarchical taxonomy of Human-AI collaboration patterns based on how often the AI was consulted during the study and how AI suggestions were used (Figure~\ref{figure:3}a; Methods). 

Our analysis revealed several categories of collaboration. For example, some participants chose to always rely on the AI feedback, whereas others only utilized it as a source of inspiration and rewrote it in their own style. Based on the number of posts in the study for which AI was consulted (out of the 10 posts for each participant), we found that participants consulted AI either always (15.52\%), often (56.03\%), once (6.03\%), or never (22.41\%). Very few participants always consulted and used the AI (2.59\%),  indicating that they did not rely excessively on AI feedback. A substantial number of participants also chose to never consult the AI (22.41\%). Such participants, however, also expressed the least empathy in their responses (1.13 on average out of 6; Figure~\ref{figure:3}b), suggesting that consulting the AI could have been beneficial.

Furthermore, based on how AI suggestions were used, we found that participants used the suggestions either directly (64.62\%), indirectly (18.46\%), or not at all (16.92\%). As expected given our system’s design, the most common way of usage was direct,  which entailed clicking on the suggested actions to incorporate them in the response. In contrast, participants who indirectly used AI (Methods) drew ideas from the suggested feedback and rewrote it in their own words in the final response. Some participants, however, chose not to use suggestions at all (16.92\%); a review of these instances by the researchers, as well as the subjective feedback from participants, suggested that reasons included 
the feedback not being helpful, the feedback not being personalized, or their response already being empathic and leaving little room for improvement. Finally, multiple types of feedback are possible for the same combination of seeker post and original response, and some participants (16.92\%) used our reload functionality (Methods) to read through these multiple suggestions before they found something they preferred.  

In general, participants who consulted and used AI more often expressed higher empathy as evaluated through our automatic expressed empathy score (Figure~\ref{figure:3}b). This trend of increased empathy with increased AI consultation or usage was similar but less pronounced in our human evaluation (Supplementary Figure~\ref{supp:figure:hai_empathy}).  

\input{figures/_figure_3}

\subsection*{Positive perceptions of participants}
At the end of our study, we collected study participants' perceptions about the usefulness and actionability of the feedback and their intention to adopt the system. We observed that 63.31\% of participants found the feedback they received helpful, 60.43\% found it actionable, and 77.70\% of participants wanted this type of feedback system to be deployed on TalkLife or other similar peer-to-peer support platforms (Supplementary Figure~\ref{supp:figure:exit_survey_perceptions}), indicating the overall effectiveness of our approach. We also found that 69.78\% of participants self-reported feeling more confident at providing support after our study; this indicates the potential value of our system for training and increased self-efficacy (Supplementary Figure~\ref{supp:figure:exit_survey_perceptions}).

%% file: figures/_figure_2.tex
\begin{figure}
\centering
\caption{\textbf{(a)} Human evaluation from an independent set of TalkLife users showed that the Human + AI responses were strictly preferred 46.88\% of the time relative to a 37.39\% strict preference for the Human Only responses. \textbf{(b)} Through automatic evaluation using an AI-based expressed empathy score\cite{Sharma2020-dx}, we found that the Human + AI responses had 19.60\% higher empathy than the Human Only responses (1.77 vs. 1.48; p < $10^{-5}$; Two-sided Student's t-test). \textbf{(c)} For the participants who reported challenges in writing responses after the study, we found a stronger preference for the Human + AI responses vs. Human Only responses (49.12\% vs. 35.96\%), compared to participants who did not report challenges (44.62\% vs. 41.54\%). \textbf{(d)} For participants who reported challenges in writing responses after the study, we found a higher improvement in expressed empathy scores of the Human + AI responses vs. Human Only responses (38.88\%; 1.74 vs. 1.25), compared to participants who did not report challenges (11.87\%; 1.79 vs. 1.60). In \textbf{c} and \textbf{d}, the sample size varied to ensure comparable conditions (Methods). Error bars indicate bootstrapped 95\% confidence intervals.}
    \includegraphics[width=0.9\textwidth]{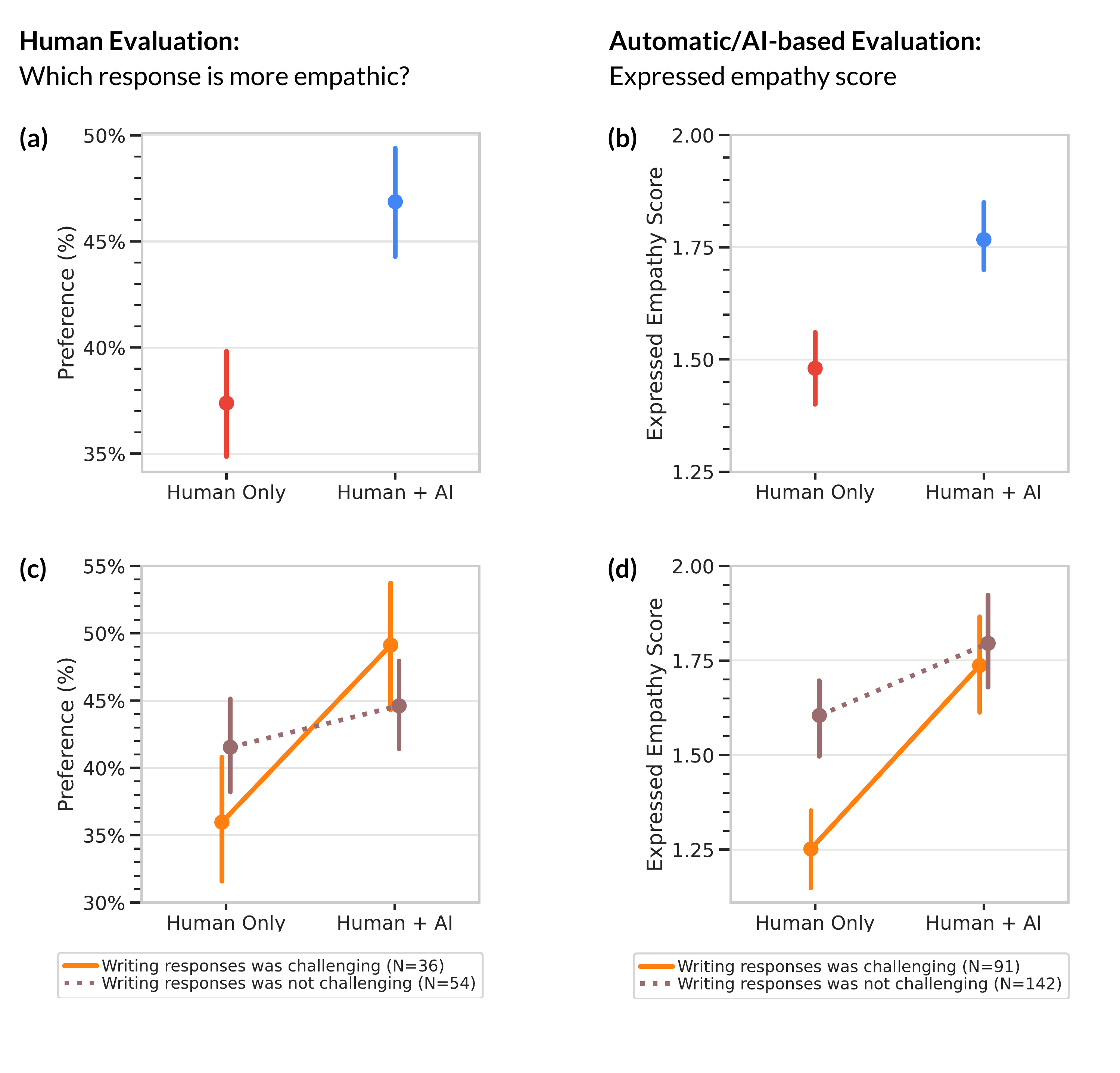}
\label{figure:2}
\end{figure}

%% file: figures/_figure_3.tex
\begin{figure}
    \caption{We derived a hierarchical taxonomy of Human-AI collaboration categories. \textbf{(a)} We clustered the interaction patterns of Human + AI (treatment) participants based on how often the AI was consulted during the study and how the AI suggestions were used (N=116/139). We excluded participants who belonged to multiple clusters (N=23/139). Very few participants always consulted and used AI (2.59\%), indicating that participants did not rely excessively on AI feedback. Participants could use AI feedback directly through suggested actions (64.62\%) or indirectly by drawing ideas from the suggested feedback and rewriting it in their own words in the final response (18.46\%). \textbf{(b)} Empathy increased when participants consulted and used AI more frequently, with those who did not consult AI (22.41\%) having the lowest empathy levels (1.13 on average out of 6). The area of the points is proportional to the number of participants in the respective human-AI collaboration categories. Error bars indicate bootstrapped 95\% confidence intervals. }
    \centering
	\includegraphics[width=0.8\columnwidth]{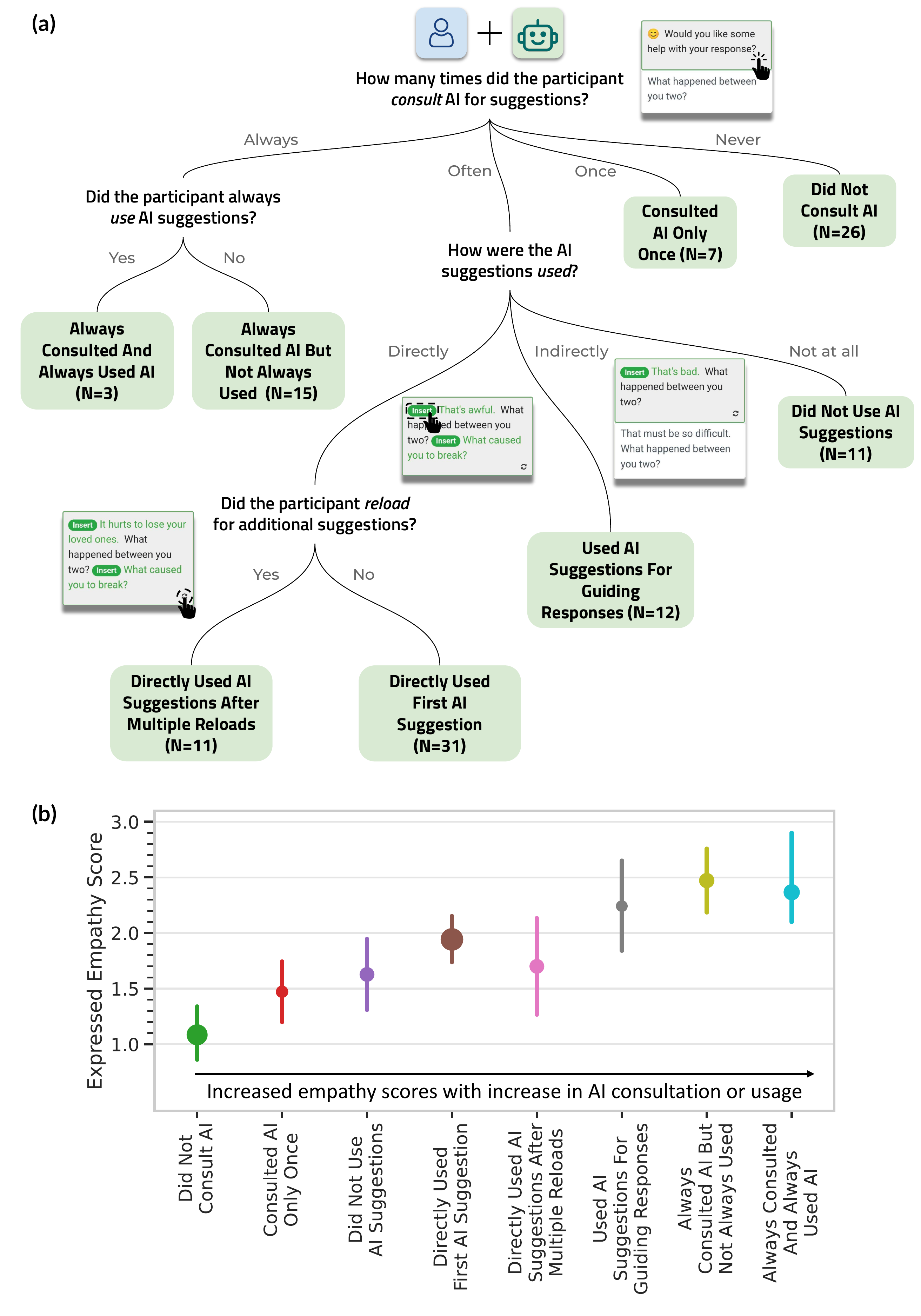}
    \label{figure:3}
\end{figure}

%% file: 03_discussion.tex
\section*{Discussion}
Our work demonstrates how humans and AI might collaborate on open-ended, social, creative tasks 
such as conducting empathic conversations. Empathy is complex and nuanced\cite{Davis1980-qw,Riess2017-jf,Blease2020-of,Doraiswamy2020-dk} and is thus more challenging for AI than many other Human-AI collaboration tasks, such as scheduling meetings, therapy appointments and checking grammar in text. We show how the joint effects of humans and AI can be leveraged to help peer supporters, especially those who have difficulty providing support, converse more empathically with those seeking mental health support.

Our study has implications for addressing barriers to mental health care, where existing resources and interventions are insufficient to meet the current and emerging need. According to a WHO report, over 400 million people globally suffer from a mental health disorder, with approximately 300 million suffering from depression\cite{noauthor_undated-ra}. Overall, mental illness and related behavioral health problems contribute 13\% to the global burden of disease, more than both cardiovascular diseases and cancer\cite{Collins2011-vg}. Although psychotherapy and social support\cite{Kaplan1977-rv} can be effective treatments, many vulnerable individuals have limited access to therapy and counseling\cite{Kazdin2011-de,Olfson2016-xg}. For example, most countries have less than one psychiatrist per 100,000 individuals,
indicating widespread shortages of workforce and inadequate in-person treatment options\cite{Rathod2017-lo}. 

One way to scale up support is by connecting millions of peer supporters online on platforms like TalkLife (\href{https://www.talklife.com/}{talklife.com}), YourDost (\href{https://yourdost.com/}{yourdost.com}) or Mental Health Subreddits (\href{https://reddit.com/}{reddit.com}) \cite{Naslund2016-dj} to those with mental health issues. However, a key challenge in doing so lies in enabling effective and high-quality conversations between untrained peer supporters and those in need at scale. We show that Human-AI collaboration can considerably increase empathy in peer supporter responses, a core component of effective and quality support that ensures improved feelings of understanding and acceptance\cite{Bohart2002-vv,Elliott2011-qr,Watson2002-zx,Sharma2020-dx}. While fully replacing humans with AI for empathic care has previously drawn skepticism from psychotherapists\cite{Blease2020-of,Doraiswamy2020-dk}, 
our results suggest that it is feasible to empower untrained peer supporters with appropriate AI-assisted technologies in relatively lower-risk settings, 
such as peer-to-peer support\cite{Lee2021-kg,Vaidyam2021-hu,Miner2019-ri,Imel2015-sa,Kazdin2011-de}. 

Our findings also point to potential secondary gain for peer supporters in terms of (1) increased self-efficacy, as indicated by 69.78\% of participants feeling more confident in providing support after the study, and (2) gained  experience and expertise by multiple example learning when using reload functionality to scroll through multiple types of responses for the same seeker post. This has implications for helping untrained peer supporters beyond providing them just-in-time feedback. One criticism of AI is that it may steal or dampen opportunities for training more clinicians and workforce\cite{Blease2020-of,Doraiswamy2020-dk}. We show that Human-AI collaboration can actually enhance, rather than diminish, these training opportunities. This is also reflected in the subjective feedback from participants (Methods), with several participants reporting different types of learning after interacting with the AI (e.g., one participant wrote, ``\textit{I realized that sometimes I directly jump on to suggestions rather than being empathic first. I will have to work on it.}'', while another wrote, ``\textit{Feedback in general is helpful. It promotes improvement and growth.}'').

Further, we find that participants not only directly accept suggestions but also draw higher-level inspiration from the suggested feedback (e.g., a participant wrote ``\textit{Sometimes it just gave me direction on what [should] be said and I was able to word it my way. Other times it helped add things that made it sound more empathic...}''), akin to having access to a therapist's internal brainstorming, which participants can use to rewrite responses in their own style.

In our study, many participants (N=91) reported challenges in writing responses (e.g., several participants reported not knowing what to say: ``\textit{I sometimes have a hard time knowing what to say.}''), which is characteristic of the average user on online peer-to-peer support platforms\cite{kemp2012challenges,mahlke2014peer,Sharma2020-dx}. We demonstrate a significantly larger improvement in empathy for these users, suggesting that we can provide significant assistance in writing more empathic responses, thereby improving empathy awareness and expression of the typical platform user\cite{Sharma2020-dx,Sharma2021-rq}. Through qualitative analysis of such participants’ subjective feedback on our study, we find that \oursystem~can guide someone who is unsure about what to say (e.g., a participant wrote, ``\textit{Feedback gave me a point to start my own response when I didn't know how to start.}'') and can help them frame better  responses (e.g., one participant wrote, ``\textit{Sometimes I didn't really knew [sic] how to form the sentence but the feedback helped me out with how I should incorporate the words.}'', while another wrote, ``\textit{Sometimes I do not sound how I want to and so this feedback has helped me on sounding more friendly...}''; Methods). 
One concern is that AI, if used in practice, may harm everyone in healthcare, from support seekers to care providers and peer supporters\cite{Richardson2021-dg,Blease2020-of,Doraiswamy2020-dk}.
According to our findings, however, the individuals struggling to do a good job are the ones who benefit the most, which forms an important use case of AI in healthcare.

The reported differences in empathy between treatment and control groups conservatively estimates the impact of our AI-in-the-loop feedback system due to (1) additional initial empathy training to both Human + AI and Human Only groups (Supplementary Figure~\ref{supp:figure:empathy_training}), and (2) a potential selection effect that may have attracted TalkLife users who care more about supporting others (Supplementary Figure~\ref{supp:figure:talklife_response}). In practice, training of peer supporters is very rare, and the effect of training typically diminishes over time\cite{Nunes2011-cz,Hojat2009-qc}. We included this training to understand whether just-in-time AI feedback is helpful beyond traditional training methods. Moreover, the Human Only responses in our study had 34.52\% higher expressed empathy than existing Human Only responses to the corresponding seeker posts on the TalkLife platform (1.11 vs. 1.48; p $\ll$ $10^{-5}$; Two-sided Student's t-test; Supplementary Figure~\ref{supp:figure:talklife_response}), reflecting the effects of additional training as well as a potential selection effect. 
We show here that Human-AI collaboration improves empathy expression even for participants who already express empathy more often; practical gains for the average user of the TalkLife platform could be even higher than the intentionally conservative estimates presented here.


\subsection*{Safety, Privacy, and Ethics}
Developing computational methods for intervention in high-risk settings such as mental health care involves ethical considerations related to safety, privacy, and bias\cite{Collings2012-ec,Li2020-yt,Luxton2012-io,Martinez-Martin2018-nd}. There is a risk that in attempting to help, AI may have the opposite effect on the potentially vulnerable support seeker or peer supporter\cite{Richardson2021-dg}. The present study included several measures to reduce such risks and unintended consequences. First, our collaborative, AI-in-the-loop writing approach ensured that the primary conversation remains between two humans, with AI offering feedback only when it appears useful, and allowing the human supporter to accept or reject it. Providing such human agency is safer than relying solely on AI, especially in a high-risk mental health context\cite{Miner2019-ri}. Moreover, using only AI results in loss of authenticity in responses; hence, our Human-AI collaboration approach leads to responses with high empathy as well as high authenticity (Supplementary Figure~\ref{supp:figure:AI_only}). 

Second, our approach intentionally assists only peer supporters, not support seekers in crisis, since they are likely to be at a lower risk 
and more receptive to the feedback. Third, we filtered posts related to suicidal ideation and self-harm by using pre-defined unsafe regular expressions (e.g., ``{.*(commit suicide).*}'', ``{.*(cut).*}''). Such posts did not enter our feedback pipeline, but instead we recommended escalating them to therapists. We applied the same filtering to every generated feedback, as well, to try and ensure that \oursystem~did not suggest unsafe text as responses. Fourth, such automated filtering may not be perfect; therefore, we included a mechanism to flag inappropriate/unsafe posts and feedback by providing our participants with an explicit ``Flag Button'' (Supplementary Figure~\ref{supp:figure:study_design:flag}). In our study, 1.58\% posts (out of 1390 in the treatment group) and 2.88\% feedback instances (out of 1939 requests) were flagged as inappropriate or unsafe. While the majority of them were concerned with unclear seeker posts or irrelevant feedback, we found six cases (0.18\%) that warranted further attention. One of these cases involved the post containing intentionally misspelled self-harm content (e.g., ``\textit{c u t}'' with spaces between letters in order to circumvent safety filters); another related to feedback containing a self-harm related term; three addressed the post or feedback containing a swear word that may not directly be a safety concern (e.g., ``\textit{You are so f**king adorable}''); and one contained toxic/offensive feedback (``\textit{It's a bad face}''). 

Future iterations of our system could address these issues by leveraging more robust filtering methods and toxicity/hate speech classifiers (e.g., Perspective API (\href{https://www.perspectiveapi.com/}{perspectiveapi.com})). Several platforms, including TalkLife, already have systems in place to prevent triggering content from being shown, which can be integrated into our system on deployment. Finally, we removed all personally identifiable information (user and platform identifiers) from the TalkLife dataset prior to training the AI model.

\subsection*{Limitations}
While our study results reveal the promise of Human-AI collaboration in open-ended and even high-risk settings, the study is not without limitations. Some of our participants indicated that empathy may not always be the most helpful way to respond (e.g., when support seekers are looking for concrete actions). However, as demonstrated repeatedly in the clinical psychology literature\cite{Elliott2011-qr,Bohart2002-vv,Watson2002-zx,Sharma2020-dx}, empathy is a critical, foundational approach to all evidence-based mental health support, plays an important role in building alliance and relationship between people, and is highly correlated with symptom improvement. It has consistently proven to be an important aspect of responding, but support seekers may sometimes benefit from additional responses involving different interventions (e.g., concrete problem solving, motivational interviewing\cite{Tanana2016-jk}). Future work should investigate when such additional responses are helpful or necessary.

Some participants may have been apprehensive about using our system, as indicated by the fact that many participants did not consult or use it (N=37). Qualitatively analyzing the subjective feedback from these participants suggested that this might be due to feedback communicating incorrect assumptions about the preferences, experience, and background of participants (e.g., assuming that a participant is dealing with the same issues as the support seeker: ``\textit{Not sure this can be avoided, but the feedback would consistently assume I've been through the same thing.}''). Future work should personalize prompts and feedback to individual participants. This could include personalizing the content and the frequency of the prompt as well as personalizing the type of feedback that is shown from multiple possible feedback options. 

Our assessment includes validated yet automated and imperfect measures. Specifically, our evaluation of empathy is based only on empathy that was \textit{expressed} in responses, not empathy that might have been \textit{perceived} by the support seeker\cite{Barrett-Lennard1981-ji}. In sensitive contexts like ours, however, obtaining perceived empathy ratings from support seekers is challenging and involves ethical risks. 
We attempted to reduce the gap between expressed and perceived empathy in our human evaluation by recruiting participants from TalkLife who may be seeking support on the platform (Methods). Nevertheless, studying the effects of Human-AI collaboration on perceived empathy in conversations is a vital future research direction. However, note that psychotherapy research indicates a strong correlation between expressed empathy and positive therapeutic outcomes and commonly uses it as a credible alternative\cite{Elliott2011-qr}.

Furthermore, we acknowledge that a variety of social and cultural factors might affect the dynamics of the support and the expression of empathy\cite{de2017gender,cauce2002cultural,satcher2001mental}. As such, our Human-AI collaboration approach must be adapted and evaluated in various socio-cultural contexts, including underrepresented communities and minorities. While conducting randomized controlled trials on specific communities and investigating heterogeneous treatment effects across demographic groups is beyond the scope of our work, our study was deployed globally and included participants of various gender identities, ethnicities, ages, and countries (Methods; Supplementary Figure~\ref{supp:figure:background}, \ref{supp:figure:background_empathy}). However, this is a critical area of research, and ensuring equitable access and support requires further investigation.




Our study evaluated a single Human-AI collaboration interface design, and there could have been other potential interface designs, as well. Additionally, as a secondary exploration, we analyzed a classification-based interface design, which provided participants with the option to request automatic expressed empathy scores\cite{Sharma2020-dx} for their responses (Supplementary Figure~\ref{supp:figure:study_design:classification}). We assigned this secondary classification-based AI treatment to 10\% of the incoming participants at random (N=30). Due to conflicting human and automatic evaluation results, we observed that the effects of this secondary treatment on empathy of participants were ambiguous (Supplementary Figure~\ref{supp:figure:classification_treatment}a,~\ref{supp:figure:classification_treatment}b); however, the design was perceived as being less actionable than our primary rewriting-based interface (Supplementary Figure~\ref{supp:figure:classification_treatment}c). This poses questions on what types of design are optimal and how best to provide feedback.

Finally, we recruited participants from a single platform (TalkLife) and only for providing empathic support in the English language. We further note that this study focuses on empathy expression in peer support and does not investigate long-term clinical outcomes.

\subsection*{Conclusion}
We developed and evaluated \oursystem, a Human-AI collaboration system that led to a 19.60\% increase in empathy in peer-to-peer conversations overall and a 38.88\% increase in empathy for mental health supporters who experience difficulty in writing responses in a randomized controlled trial on a large peer-to-peer mental health platform. Our findings demonstrate the potential of feedback-driven, AI-in-the-loop writing systems to empower online peer supporters to improve the quality of their responses without increasing the risk of harmful responses. 

%% file: 04_methods.tex
\section*{Methods}
\subsection*{Study Design}

We employed a between-subjects study design in which each participant was randomly assigned to one of Human + AI (treatment; N=139) or Human Only (control; N=161) conditions. Participants in both groups were asked to write supportive, empathic responses to a unique set of 10 existing seeker posts (one at a time), sourced at random from a subset of TalkLife posts.
The Human + AI (treatment) group participants were given the option of receiving feedback through prompts as they typed their responses. Participants in the Human Only (control) group, in contrast, wrote responses with no option for feedback. 

\xhdr{Participant Recruitment} We worked with a large peer-to-peer support platform, TalkLife, to recruit participants directly from their platform. Because users on such platforms are typically untrained in best-practices of providing mental health support, their work offers a natural place to deploy feedback systems like ours. To recruit participants, we advertised our study on TalkLife (Supplementary Figure~\ref{supp:figure:advertisement}). Recruitment started in Apr 2021 and continued until Sep 2021. The study was approved by the University of Washington's Institutional Review Board (determined to be exempt; IRB ID STUDY00012706). 

\xhdr{Power Analysis} We used a power analysis to estimate the number of participants required for our study. For an effect size of 0.1 difference in empathy, a power analysis with a significance level of 0.05, powered at 80\%, indicated that we required 1,500 samples of (seeker post, response post) pairs each for treatment and control groups. To meet the required sample size, we collected 10 samples per participant and therefore recruited from 300 participants in total (with the goal of 150 participants per condition), for a total of 1,500 samples each. 

\xhdr{Dataset of Seeker Posts} We obtained a unique set of 1500 seeker posts, sampled at random with consent from the TalkLife platform, in the observation period from May 2012 to June 2020. Prior to sampling, we filtered posts related to (1) critical settings of suicidal ideation and self-harm to ensure participant safety (Discussion), and (2) common social media interactions not related to mental health (e.g., ``\textit{Happy mother's day}'')\cite{Sharma2021-rq}. We randomly divided these 1500 posts into 150 subsets of 10 posts each. 
We used the same 150 subsets for both treatment and control conditions for consistent context for both groups of participants.
 
\xhdr{Participant Demographics} In our study, 54.33\% of the participants identified as female, 36.67\% as male, 7.33\% as non-binary, and the remaining 1.67\% preferred not to report their gender. The average age of participants was 26.34 years (std = 9.50). 45.67\% of the participants identified as White, 20.33\% as Asians, 10.67\% as Hispanic or Latino, 10.33\% as Black or African American, 0.67\% as Pacific Islander or Hawaiian, 0.33\% as American Indian or Alaska Native, and the remaining 12.00\% preferred not to report their race/ethnicity. 62.33\% of the participants were from the United States, 13.67\% were from India, 2.33\% were from United Kingdom, 2.33\% were from Germany, and the remaining 19.33\% were from 36 different countries (spanning six of seven continents excluding Antarctica). Moreover, 31.67\% of the participants reported having no experience with peer-to-peer support despite having been recruited from the TalkLife platform, 26.33\% as having less than one year of experience, and 42.00\% reported having greater than or equal to one year of experience with peer-to-peer support. 

\xhdr{RCT Group Assignment} On clicking the advertised pop-up used for recruitment, a TalkLife user was randomly assigned to one of the Human + AI (treatment) or Human Only (control) conditions for the study duration. 

\xhdr{Study Workflow} We divided our study into four phases:
\begin{itemize}
    \item \xhdritem{Phase I: Pre-Intervention Survey} First, both control and treatment group participants were asked the same set of survey questions describing their demographics, background and experience with peer-to-peer support (Supplementary Figure~\ref{supp:figure:study_design:background}, \ref{supp:figure:study_design:onboarding}).
    \item \xhdritem{Phase II: Empathy Training and Instructions} Next, to address whether participants held similar understandings of empathy, both groups received the same initial empathy training, which included empathy definitions, frameworks, and examples based on psychology theory, before starting the main study procedure of writing empathic responses (Supplementary Figure~\ref{supp:figure:empathy_training}). Participants were also shown instructions on using our study interface in this phase (Supplementary Figure~\ref{supp:figure:study_design:instructions_control_1}, \ref{supp:figure:study_design:instructions_control_2}, \ref{supp:figure:study_design:instructions_treatment_1}, \ref{supp:figure:study_design:instructions_treatment_2}, \ref{supp:figure:study_design:instructions_treatment_3}, \ref{supp:figure:study_design:instructions_treatment_4}, \ref{supp:figure:study_design:instructions_treatment_5}, \ref{supp:figure:study_design:instructions_treatment_6}). 
    \item \xhdritem{Phase III: Write Supportive, Empathic Responses} Participants then started the main study procedure and wrote responses to one of the 150 subsets of 10 existing seeker posts (one post at a time). For each post, participants in both the groups were prompted ``\textit{Write a supportive, empathic response here}''. The Human + AI (treatment) group participants were given the option of receiving feedback through prompts as they typed their responses (Supplementary Figure~\ref{supp:figure:study_design:interface_treatment}). Participants in the Human Only (control) group wrote responses without any option for feedback (Supplementary Figure~\ref{supp:figure:study_design:interface_control}). 
    \item \xhdritem{Phase IV: Post-Intervention Survey} After completing the 10 posts, participants in both groups were asked to assess the study by answering questions about the difficulty they faced while writing responses, the helpfulness and actionability of the feedback, their self-efficacy after the study, and the intent to adopt the system (Supplementary Figure~\ref{supp:figure:study_design:exit_survey_control}, \ref{supp:figure:study_design:exit_survey_treatment_1}, \ref{supp:figure:study_design:exit_survey_treatment_2}). \end{itemize}

If participants dropped out of the study before completing it, their data was removed from our analyses. Participants took 20.62 minutes on average to complete the study. US citizens and permanent US residents were compensated with a 5 USD Amazon gift card. Furthermore, the top-2 participants in the human evaluation (Evaluation) received an additional 25 USD Amazon gift card. Based on local regulations, we were unable to pay non-US participants. This was explicitly highlighted in the participant consent form on the first landing page of our study (Supplementary Figure~\ref{supp:figure:study_design:consent}, \ref{supp:figure:study_design:feedback_eval:consent}).

\subsection*{Design Goals}
\oursystem~is designed (1) with a collaborative ``AI-in-the-loop'' approach, (2) to provide actionable feedback, and (3) to be mobile friendly.

\xhdr{Collaborative AI-in-the-loop Design} In the high-risk setting of mental health support, AI is best used to augment, rather than replace, human skill and knowledge\cite{Miner2019-ri,Chen2017-en}. Current natural language processing technology -- including language models, conversational AI methods, and chatbots -- continue to pose risks related to toxicity, safety, and bias, which can be life-threatening in contexts of suicidal ideation and self-harm\cite{Wolf2017-xf,Bolukbasi2016-wo,Daws2020-ds,Richardson2021-dg}. To mitigate these risks, researchers have called for Human-AI collaboration methods, where primary communication remains between two humans with an AI system ``in-the-loop'' to assist humans in improving their conversation\cite{Miner2019-ri,Chen2017-en}. In \oursystem, humans remain at the center of the interaction, receive suggestions from our AI ``in-the-loop,'' and retain full control over which suggestions to use in their responses (e.g., by selectively choosing the most appropriate Insert or Replace suggestions and editing them as needed).

\xhdr{Actionable Feedback} Current AI-in-the-loop systems are often limited to addressing ``what'' (rather than "how") participants should improve\cite{Tanana2019-bj,Peng2020-sg,Hui2018-ty,Kelly2018-fo}. For such a goal, it is generally acceptable to design simple interfaces that prompt participants to leverage strategies for successful supportive conversations (e.g., prompting ``\textit{you may want to empathize with the user}'') without any instructions on how  to concretely apply those strategies. However, for complex, hard-to-learn constructs such as empathy\cite{Davis1980-qw,Elliott2011-qr}, there is a need to  address the more actionable goal of steps to take for participants to improve. \oursystem, designed to be actionable, suggests concrete actions (e.g., sentences to insert or replace) that participants may take to make their current response more empathic. 

\xhdr{Mobile Friendly Design} Online conversations and communication are increasingly mobile based. This is also true for peer-to-peer support platforms, which generally provide their services through a smartphone application. Therefore, a mobile friendliness design is critical for the adoption of conversational assistive agents like ours. However, the challenge here relates to the complex nature of the feedback and the smaller, lower-resolution screen on a mobile device as compared to a desktop. We therefore designed a compact, minimal interface that works equally well on desktop and mobile platforms. We created a conversational experience based on the mobile interface of peer-to-peer support platforms that was design minimal, used responsive prompts that adjusted in form based on screen sizes, placed AI feedback compactly above the response text box for easy access, and provided action buttons that were easy for mobile users to click on.

\subsection*{Feedback Workflow}
Through \oursystem, we showed prompts to participants that they could click on to receive feedback. Our feedback, driven by a previously validated Empathic Rewriting model, consists of actions that users can take to improve the empathy of their responses (Supplementary Figure~\ref{supp:figure:study_design:interface_treatment}). 

\xhdr{Prompts to Trigger Feedback} We showed the prompt ``\textit{Would you like some help with your response?}'' to participants, which was placed above the response text box (Figure~\ref{figure:1}b). Participants could at any point click on the prompt to receive feedback on their current response (including when it is still empty). When this prompt is clicked, \oursystem~acts on the seeker post and the current response to suggest changes that will make the response more empathic. Our suggestions consisted of Insert and Replace operations generated through empathic rewriting of the response. 

\xhdr{Generating Feedback through Empathic Rewriting} The goal of empathic rewriting, originally proposed in Sharma et al.\cite{Sharma2021-rq}, is to transform low empathy text to higher empathy. The authors proposed \Partner, a deep reinforcement learning model that learns to take sentence-level edits as actions in order to increase the expressed level of empathy while maintaining conversational quality. \Partner's learning policy is based on a transformer language model (adapted from GPT-2\cite{Radford_undated-yv}), which performs the dual task of generating candidate empathic sentences and adding those sentences at appropriate positions. Here, we build on \Partner~by further improving training data quality through additional filtering, supporting multiple generations for the real-world use-case of multiple types of feedback for the same post, and evaluating a broader range of hyperparameter choices.

\xhdr{Showing Feedback as Actions} We map the rewritings generated by our optimized version of \Partner~to suggestions to \textit{Insert} and \textit{Replace} sentences. These suggestions are then shown as actions to edit the response. To incorporate the suggested changes, the participant clicks on the respective Insert or Replace buttons. Continuing our example from Figure~\ref{figure:1}, given the seeker post ``\textit{My job is becoming more and more stressful with each passing day.}'' and the original response ``\textit{Don't worry! I'm there for you}.'', \Partner~takes two insert actions – Replace ``\textit{Don't worry!}'' with ``\textit{It must be a real struggle!}'' and Insert ``\textit{Have you tried talking to your boss?}'' at the end of the response. These actions are shown as feedback to the participant. See Supplementary Figure~\ref{supp:figure:qualitative} for more qualitative examples.

\xhdr{Reload Feedback If Required} 
For the same combination of seeker post and original response, multiple feedback suggestions are possible. In the Figure 1 example, instead of suggesting the insert ``\textit{Have you tried talking to your boss?}'', we could also propose inserting ``\textit{I know how difficult things can be at work}''. These feedback variations can be sampled from our model and, if the initial sampled feedback does not meet participant needs, iterated upon to help participants find better-suited feedback. \oursystem~provides an option to \textit{reload} feedback, allowing participants to navigate through different feedback and suggestions if necessary.

\subsection*{Evaluation}
\xhdr{Empathy Measurement} We evaluated empathy using both human and  automated methods. For our human evaluation, we recruited an independent set of participants from the TalkLife platform and asked them to compare responses written with feedback to those written without feedback given the same seeker post (Supplementary Figure~\ref{supp:figure:study_design:feedback_eval:consent}, \ref{supp:figure:study_design:feedback_eval:instructions}, \ref{supp:figure:study_design:feedback_eval:interface}). When analyzing strata of participants based on challenges in writing responses (Figure~\ref{figure:1}c), we considered only those seeker post instances for which the respective Human Only and Human + AI participants both indicated writing as challenging or not challenging. Since our human evaluation involves comparing Human Only and Human + AI responses, this ensures that participants in each strata belong to only one of challenging or not challenging categories.

Though our human evaluation captures platform users' perceptions of empathy in responses, it is unlikely to measure empathy from the perspective of psychology theory given the limited training of TalkLife users. Therefore, we conducted a second complementary evaluation by applying the theory-based empathy classification model proposed by Sharma et al.\cite{Sharma2020-dx}, which assigns a score between 0 and 6 to each response and has been validated and used in prior work\cite{Sharma2021-rq,zheng2021comae,wambsganss2021supporting,majumder2021exemplars}. Note that this approach evaluates empathy expressed in responses and not the empathy perceived by support seekers of the original seeker post (Discussion).

\subsection*{Deriving a Hierarchical Taxonomy of Human-AI Collaboration Patterns}
We wanted to understand how different participants collaborated with \oursystem. To derive collaboration patterns at the participant level, we aggregated and clustered post-level interactions for each participant over the 10 posts in our study. First, we identified three dimensions of interest that were based on the design and features of \oursystem~as well as by qualitatively analyzing the interaction data: (1) the number of posts in the study for which the AI was consulted, (2) the way in which AI suggestions were used (direct vs. indirect vs. not at all), and (3) whether the participant looked for additional suggestions for a single post (using the reload functionality). 

Direct use of AI was defined as directly accepting the AI's suggestions by clicking on the respective Insert or Replace buttons. Indirect use of AI, in contrast, was defined as making changes to the response by drawing ideas from the suggested edits. We operationalized indirect use as a cosine similarity of more than $95\%$ between the BERT-based embeddings\cite{Devlin2018-cl} of the final changes to the response by the participant and the edits suggested by the AI. Next, we used k-means to cluster the interaction data of all participants on the above dimensions (k=20 based on the Elbow method\cite{Wikipedia_contributors2022-ol}). We manually analyzed the distribution of the 20 inferred clusters, merged similar clusters, discarded the clusters that were noisy (e.g., too small or having no consistent interaction behavior), and organized the remaining 8 clusters in a top-down approach to derive the hierarchical taxonomy of Human-AI collaboration patterns (Figure~\ref{figure:3}a; Results). Finally, for the collaboration patterns with simple rule-based definitions (e.g., participants who never consulted AI), we manually corrected the automatically inferred cluster boundaries to make the patterns more precise, e.g., by keeping only the participants who had never consulted AI in that cluster. 

%% file: 05_acknowledgements.tex
\section*{Acknowledgements}
We would like to thank TalkLife and Jamie Druitt for supporting this work, for advertising the study on their platform, and for providing us access to a TalkLife dataset. We also thank members of the UW Behavioral Data Science Group, Microsoft AI for Accessibility team, and Daniel S. Weld for their suggestions and feedback. T.A., A.S, and I.W.L were supported in part by NSF grant IIS-1901386, NSF grant CNS-2025022, NIH grant R01MH125179, Bill \& Melinda Gates Foundation (INV-004841), the Office of Naval Research (\#N00014-21-1-2154), a Microsoft AI for Accessibility grant, and a Garvey Institute Innovation grant. A.S.M. was supported by grants from the National Institutes of Health, National Center for Advancing Translational Science, Clinical and Translational Science Award (KL2TR001083 and UL1TR001085) and the Stanford Human-Centered AI Institute. 

\xhdr{Conflict of Interest Disclosure} D.C.A. is a co-founder with equity stake in a technology company, Lyssn.io, focused on tools to support training, supervision, and quality assurance of psychotherapy and counseling.

%% file: 06_supplementary.tex
\appendix
\newpage

\setcounter{table}{0}
\renewcommand{\thetable}{S\arabic{table}}
\setcounter{figure}{0}
\renewcommand{\thefigure}{S\arabic{figure}}

\section*{Supplementary Materials}
\subsection*{List of supplementary materials}
Table S1\\
Figures S1 to S39\\

\input{suppTables/_supp_table_rct}
\include{suppFigures/_supp_figure_empathy_training}
\include{suppFigures/_supp_figure_AI_only}
\include{suppFigures/_supp_figure_exit_survey_perceptions}
\include{suppFigures/_supp_figure_classification}
\include{suppFigures/_supp_figure_classification_interface}
\include{suppFigures/_supp_figure_empathy_over_time}
\include{suppFigures/_supp_figure_talklife_responses}
\include{suppFigures/_supp_figure_qualitative_examples}
\include{suppFigures/_supp_figure_hai_empathy}

\include{suppFigures/_supp_figure_background}

\include{suppFigures/_supp_figure_background_empathy}

\include{suppFigures/_supp_figure_onboarding}
\include{suppFigures/_supp_figure_exit_survey_challenging}
\include{suppFigures/_supp_figure_exit_survey_feedback_would_improve}
\include{suppFigures/_supp_figure_exit_survey_treatment}
\include{suppFigures/_supp_figure_exit_survey_vs_empathy}
\include{suppFigures/_supp_figure_hai_perceptions}
\include{suppFigures/_supp_figure_advertisement}
\include{suppFigures/_supp_figure_study_design}
\include{suppFigures/_supp_figure_feedback_eval_design}

\include{suppFigures/_supp_figure_empathic_rewriting_model}

%% file: suppTables/_supp_table_rct.tex
\begin{table}
\small
\centering
\caption{Description of our randomized controlled trial (RCT) study population, setting and model, following reporting standards for artificial intelligence in health care from Hernandez-Boussard et al.\cite{hernandez2020minimar} }
\label{supp:tab:rct}
\begin{tabular}{p{3.5cm}p{12.5cm}} 
\toprule
Feature & Description  \\ 
\midrule
\multicolumn{2}{l}{\textbf{Study population and setting}} \\
- Population: & 300 TalkLife users; 161 in Human Only (control); 139 in Human + AI (treatment).  \\
- Study setting: & Non-clinical, online platform outside of TalkLife to ensure platform users' safety, through an interface similar to TalkLife's chat feature. \\
- Data collected in RCT: & Participants responded to 10*300=3000 seeker posts (1500 unique seeker posts duplicated across control and treatment), generating 3000 responses (1610 in control, 1390 in treatment). An independent set of 50 participants rated 1390 pairs of control and treatment responses on empathy preference.   \\
- Cohort selection: & Participants were sent a recruitment request after they submitted a response on the TalkLife platform, with an aim of targeting active peer supporters. Participants were excluded if they dropped out of the study before completion. \\
- Registration: & We did not pre-register on \href{clinicaltrials.gov}{ClinicalTrials.gov} because our study was conducted in a non-clinical setting. \\
\midrule
\multicolumn{2}{l}{\textbf{Participant demographic characteristics}} \\
- Age: & Mean=26.34 years; Std=9.50 years \\
- Gender: & Female: 54.33\%; Male: 36.67\%; Non-binary: 7.33\%; Preferred not to say: 1.67\% \\
- Race/Ethnicity: & White: 45.67\%; Asian: 20.33\%; Hispanic or Latino: 10.67\%; Black or African American: 10.33\%; Pacific Islander or Hawaiian: 0.67\%; American Indian or Alaska Native: 0.33\%; Preferred not to say: 12.00\% \\
\midrule
\multicolumn{2}{l}{\textbf{\oursystem's modeling components}} \\ 
- Model output: & Empathic rewriting of the response post  \\
- Target user: & Peer supporter (users who provide peer-to-peer support to the support seeker) \\
- Data splitting: &  Training: 3.2M; Test: 0.1M; Validation: 0.1M (seeker post, response post) pairs  \\
- Gold standard: &  180 empathic rewritings from human therapy experts used for evaluation of the original \Partner~ model\cite{Sharma2021-rq}.  \\ 
- Model task: & Text generation \\ 
- Model architecture: & Deep reinforcement learning with a transformer based language model as its policy. \\
- Optimization: & Based on reward functions to increase empathy in posts and maintain text fluency, sentence coherence, context specificity, and diversity. \\
- Internal validation: & Automatic and human evaluation on hold-out test set. \\
- External validation: &  The empathy scale used in \oursystem~and \Partner\cite{Sharma2021-rq} has previously been shown to correlate with ``likes'' from the support seeker and the forming of relationships b/w support seekers and peer supporters\cite{Sharma2020-dx}, consistent with empathy theory\cite{Bohart2002-vv,Elliott2011-qr,Watson2002-zx}. Our present randomized controlled trial represents an external evaluation of the rewriting modeling components of \oursystem~and \Partner\cite{Sharma2021-rq}. \\
- Transparency: & Data is available from TalkLife through a Data License Agreement; code is available via GitHub (\href{https://github.com/behavioral-data/PARTNER}{github.com/behavioral-data/PARTNER}).\\
\bottomrule
\end{tabular}
\end{table}

%% file: suppFigures/_supp_figure_empathy_training.tex
\begin{figure}
    \caption{Empathy training used in our study. Participants in both the Human + AI (treatment) and Human Only (control) groups received the same training. The training included the empathy definition, a framework of common ways of expressing empathy in responses, and examples of empathic responses. This ensures that participants were working under similar understandings of empathy. In practice, such training is very rare and the effect of training typically diminishes over time. The identified difference in empathy between treatment and control groups in our study therefore conservatively estimates the impact of our AI-in-the-loop feedback system, and not baseline differences in empathy definitions. The effect in practice may be larger than the intentionally conservative estimates produced here, as such training is uncommon on current mental health platforms.}
    \centering
         \includegraphics[width=\textwidth]{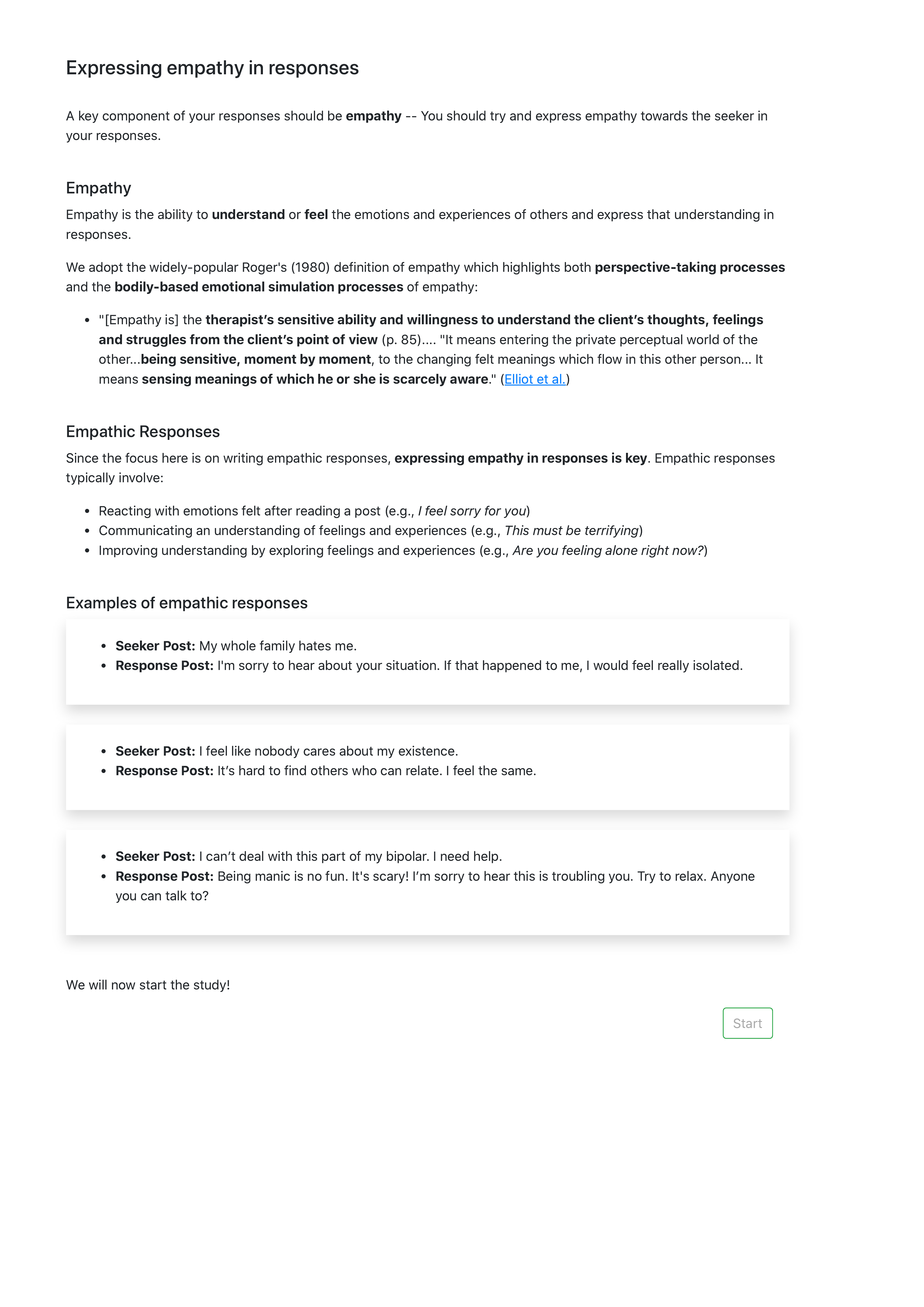}
    \label{supp:figure:empathy_training}
\end{figure}

%% file: suppFigures/_supp_figure_AI_only.tex
\begin{figure}
\centering
\caption{Comparison of Human Only (control) and Human + AI (treatment) responses with AI Only responses (generated directly from \textsc{Partner}\cite{Sharma2021-rq}, the deep reinforcement learning model for empathic rewriting, used as a foundation for \oursystem~(Methods)). \textbf{(a)} Through human evaluation from an independent set of TalkLife users, we found that AI Only responses have a similar preference as the Human + AI responses (48.20\% vs. 46.87\%; p=0.23; Two-sided Student's t-test) but a higher preference than the Human Only responses (48.20\% vs. 37.40\%; p < $10^{-5}$; Two-sided Student's t-test). \textbf{(b)} Automatic estimation of empathy, on the contrary, suggested that AI Only responses have a higher expressed empathy score compared to Human + AI responses (2.10 vs. 1.77; p < $10^{-5}$; Two-sided Student's t-test). Importantly however, note that the AI Only responses were optimized on the same scoring function that we use to automatically estimate empathy, which likely explains the high scores of the AI Only approach. \textbf{(c)} However, while the authenticity of Human Only and Human + AI responses was comparable (69.55\% vs. 65.38\%; p=0.01; Two-sided Student's t-test), the authenticity of AI Only responses was significantly lower (36.49\% vs. 65.38\%; p < $10^{-5}$; Two-sided Student's t-test). This highlights the key issue of authenticity with using AI Only, alongside safety, privacy, bias and other unintended consequences in the high-risk setting of mental health. To summarize, we find that Human + AI is the only approach that leads to both high empathy and high authenticity. Error bars indicate bootstrapped 95\% confidence intervals. }
\subfloat[\textbf{Human Evaluation:} Which response is more empathic?]{
	\includegraphics[width=0.45\columnwidth]{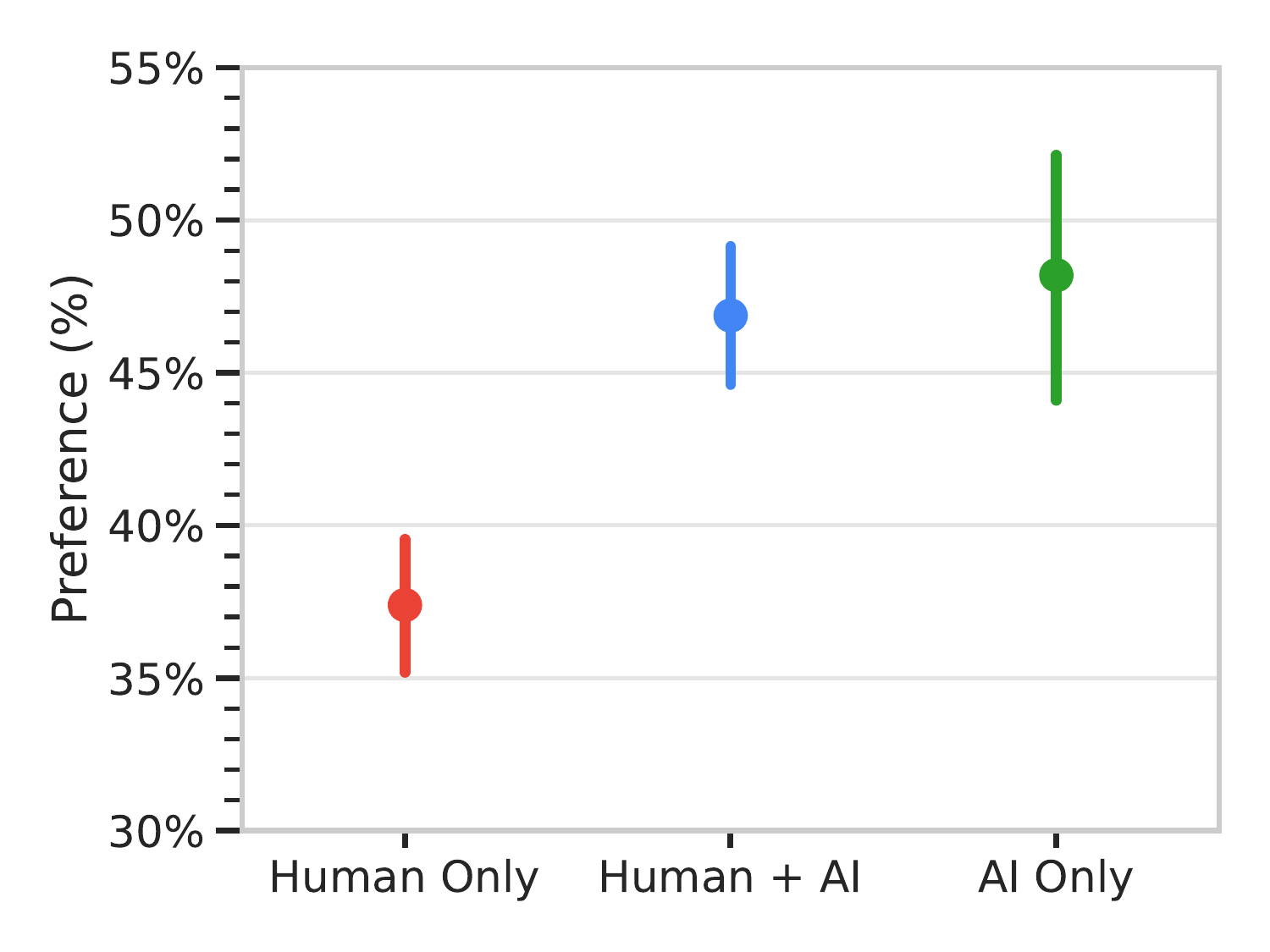} } 
\hfill
\subfloat[\textbf{Automatic/AI-based Evaluation:} Expressed empathy score]{
	\includegraphics[width=0.45\columnwidth]{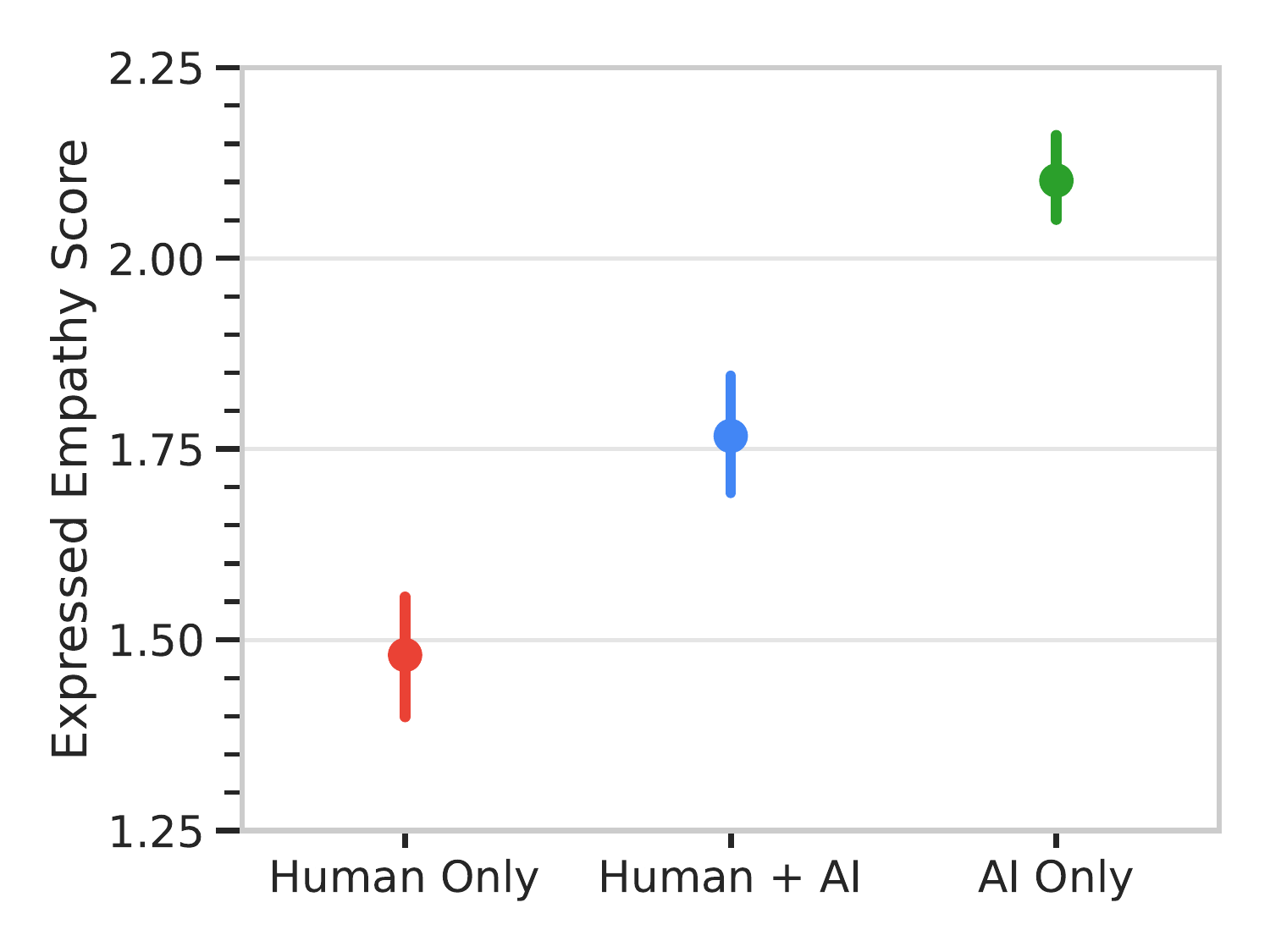} }
\hfill
\subfloat[\textbf{Authenticity:} Is the response human-written or computer-generated?]{
	\includegraphics[width=0.47\columnwidth]{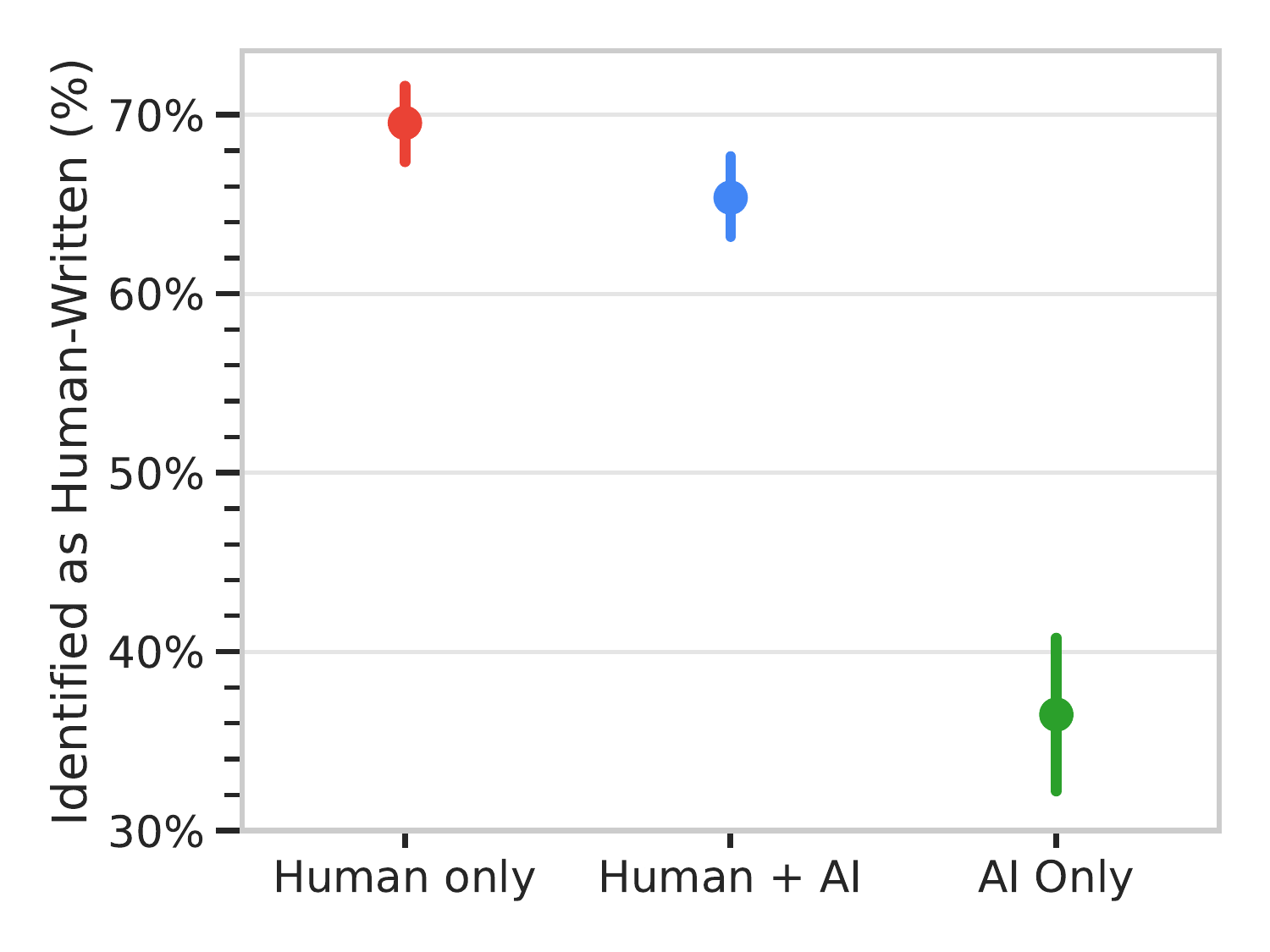} }
\label{supp:figure:AI_only}
\end{figure}

%% file: suppFigures/_supp_figure_exit_survey_perceptions.tex
\begin{figure}
    \caption{Perceptions of Human + AI (treatment) group participants as reported in phase IV (post-intervention survey). We observed that more than 63.31\% of participants found the current feedback helpful, 60.43\% found it actionable and 69.78\% of participants self-reported feeling more confident at providing support after our study. Also, 77.70\% of participants wanted this type of feedback system to be deployed on TalkLife or other similar peer-to-peer support platforms, indicating potential opportunities for deployment in real-world.}
    \centering
         \includegraphics[width=0.9\textwidth]{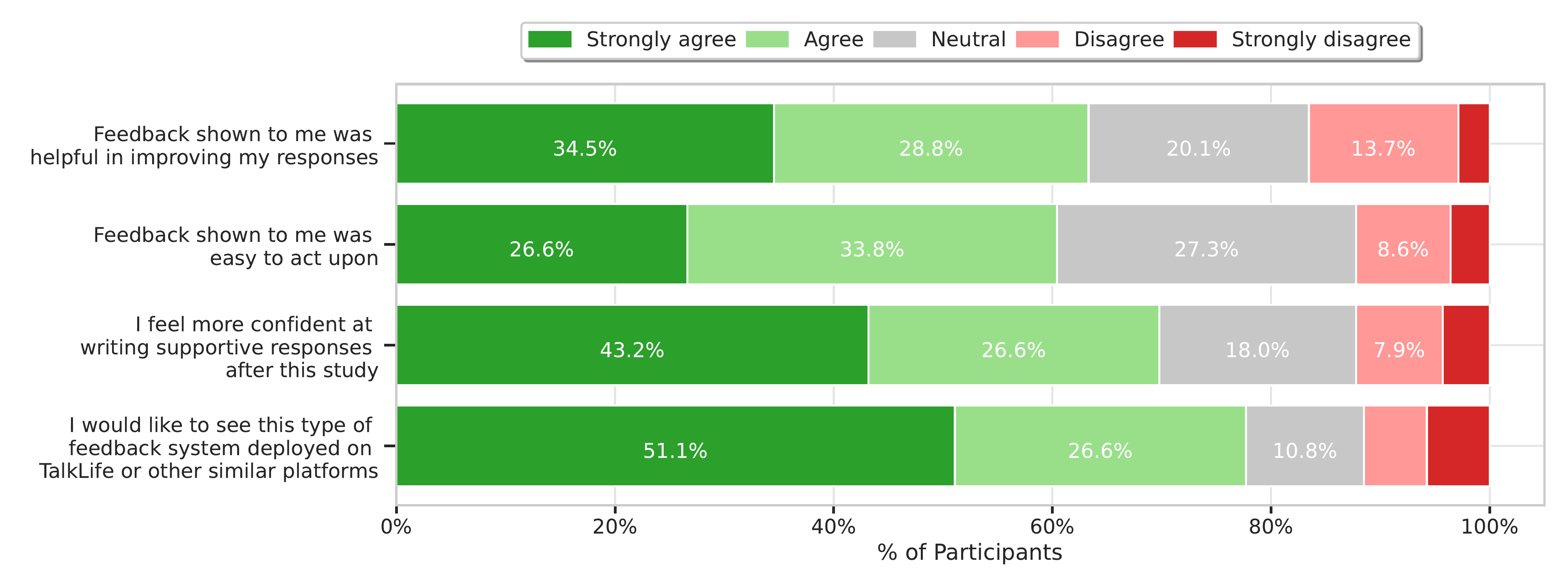}
    \label{supp:figure:exit_survey_perceptions}
\end{figure}


%% file: suppFigures/_supp_figure_classification.tex
\begin{figure}
\centering
\caption{Comparison of our rewriting-based AI treatment with a secondary classification-based AI treatment. A classification-based AI treatment provided participants with an option to request empathy classification scores for their responses, as opposed to the more granular feedback consisting of concrete suggestions to edit responses in our primary rewriting-based approach (Supplementary Figure~\ref{supp:figure:study_design:classification}). Our hypothesis was that such a treatment should be less actionable and is likely to lead to less empathic responses than the rewriting-based treatment. In our study, we assigned a secondary classification-based treatment to 10\% of the incoming participants at random (N=30). \textbf{(a)} Through human evaluation from an independent set of TalkLife users, we found that the Human + Classification responses have a significantly lower preference than the Human + Rewriting responses (37.94\% vs. 47.84\%; p < $0.01$; Two-sided Student's t-test). \textbf{(b)} Automatic estimation of empathy, on the contrary, suggested that the Human + Classification responses have a higher expressed empathy score compared to Human + Rewriting responses (2.24 vs. 1.77; p < $10^{-5}$; Two-sided Student's t-test). As the same score is also exposed to participants just-in-time in the classification-based treatment, it may have led participants to be put particular emphasis on a high expressed empathy score, which participants in the rewriting-based treatment feedback didn't have direct access to. \textbf{(c)} We found that less participants in the classification-based treatment group agree on deploying the system on TalkLife than the rewriting-based treatment (63.33\% vs. 77.70\%; p=0.0998; Two-sided Student's t-test; Supplementary Figure~\ref{supp:figure:exit_survey_perceptions}). Also, we observed that more participants in the classification-based treatment disagree on its actionability than participants in the rewriting-based treatment, but the difference may not be statistically significant due to the limited power (23.33\% vs. 12.23\%; p=0.1154; Two-sided Student's t-test). The area of the points in the plots is proportional to the number of participants in the respective control/treatment conditions. Error bars indicate bootstrapped 95\% confidence intervals.} 
\subfloat[\textbf{Human Evaluation:} Which response is more empathic?]{
	\includegraphics[width=0.45\columnwidth]{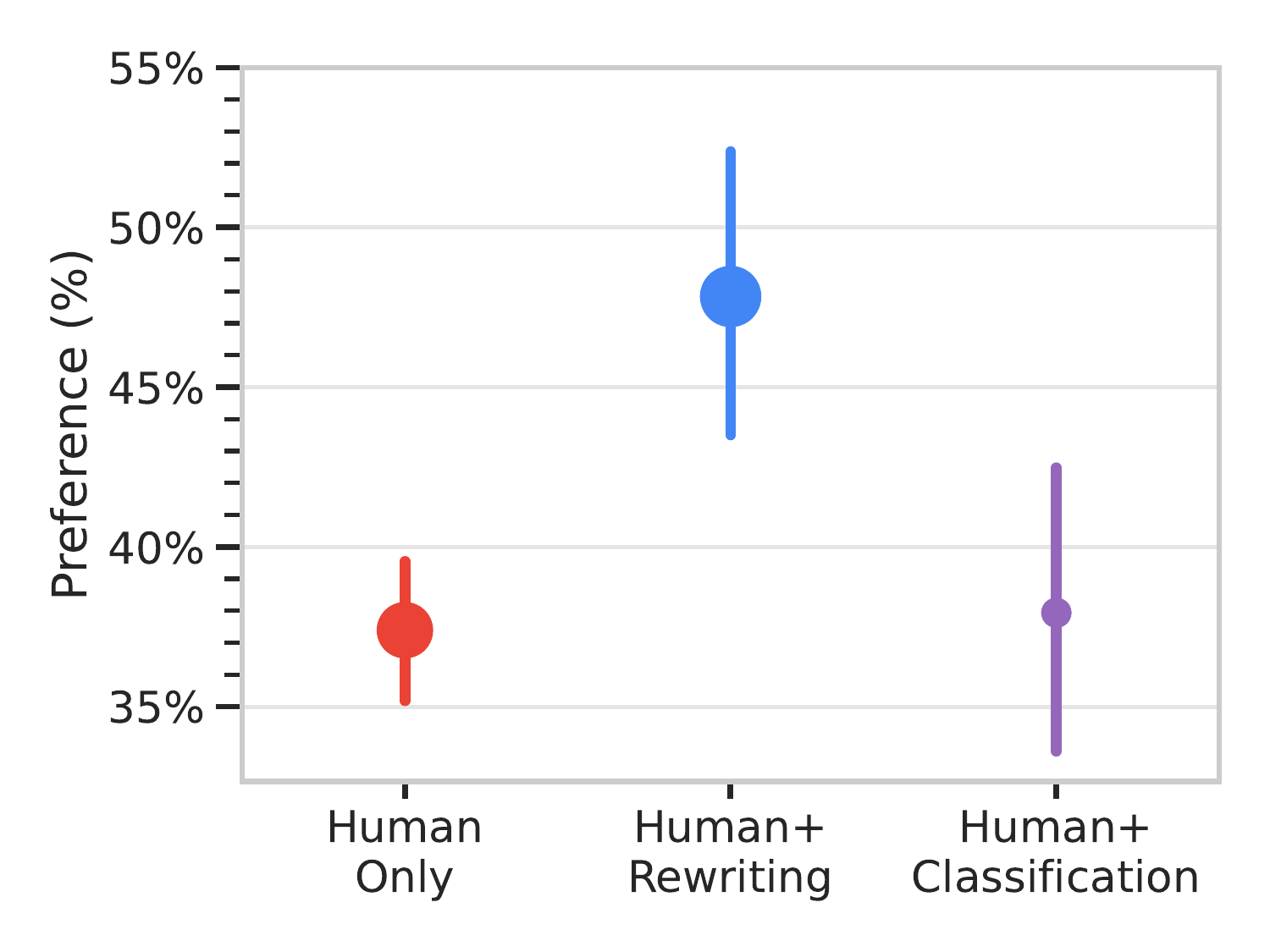} } 
\hfill
\subfloat[\textbf{Automatic/AI-based Evaluation:} Expressed empathy score]{
	\includegraphics[width=0.45\columnwidth]{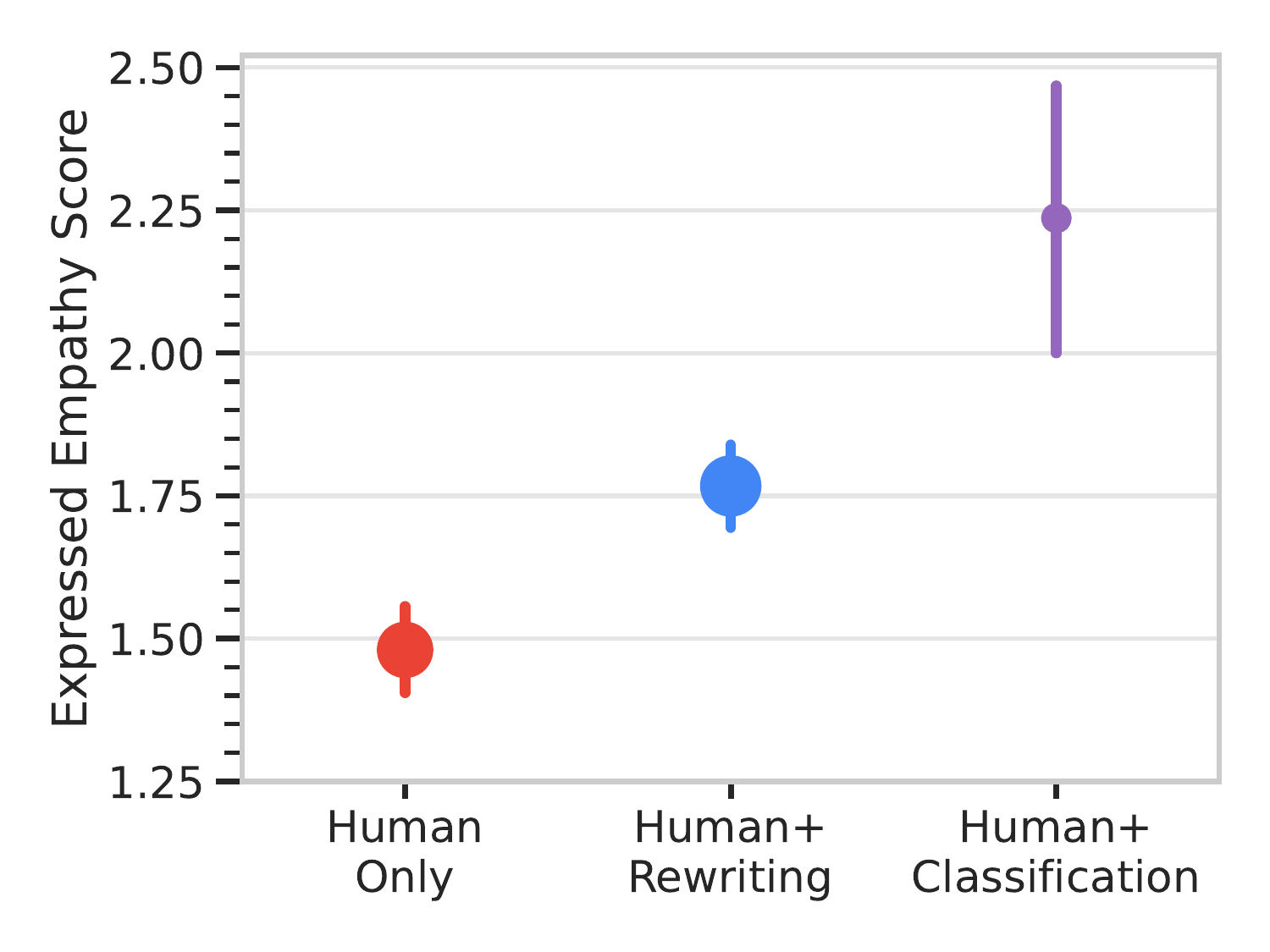} }
\hfill
\subfloat[Study participants' perceptions (classification-based treatment group)]{
	\includegraphics[width=0.9\columnwidth]{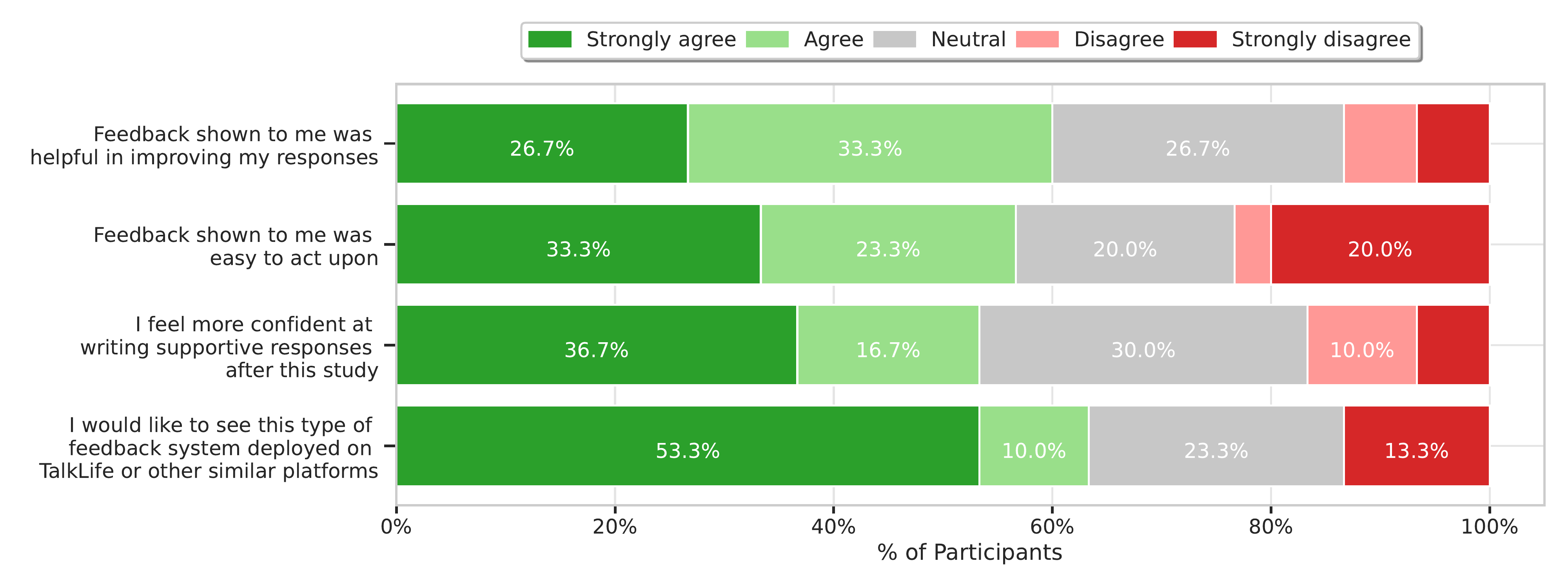} }
\label{supp:figure:classification_treatment}
\end{figure}


%% file: suppFigures/_supp_figure_classification_interface.tex
\begin{figure}
\centering
\caption{Interface of our classification-based AI treatment (Supplementary Figure~\ref{supp:figure:classification_treatment}). \textbf{(a)} Participant is asked to write a supportive, empathic response and given an option to receive feedback. \textbf{(b)} Participant starts writing the response. \textbf{(c)} Participant clicks on the ``Get Feedback'' button to request classification-based feedback. The feedback consists of classification scores on three empathy communication mechanisms -- Emotional Reactions, Interpretations, and Explorations\cite{Sharma2020-dx}. \textbf{(d)} Participant edits the response based on the classification scores, often improving on the communication mechanisms with low scores and requests ``More Feedback'' if needed.}
\subfloat[]{
	\includegraphics[width=0.25\columnwidth]{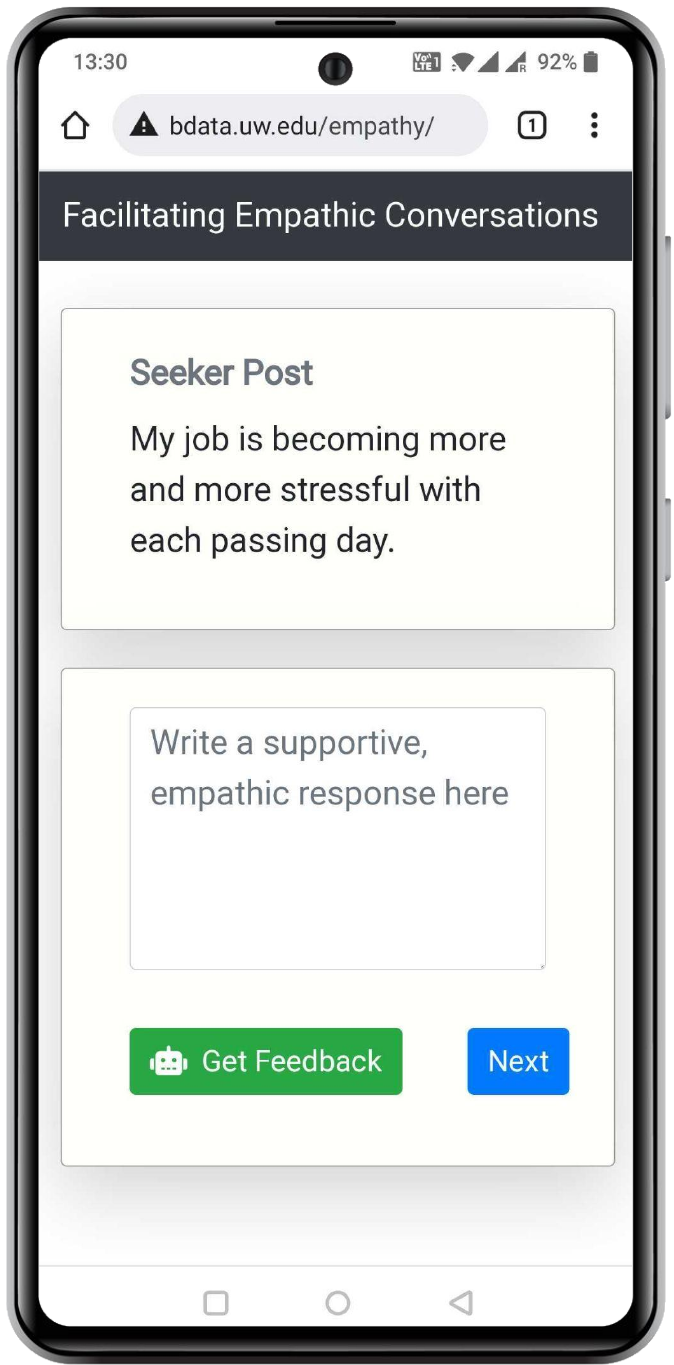} } 
\hfill
\subfloat[]{
	\includegraphics[width=0.25\columnwidth]{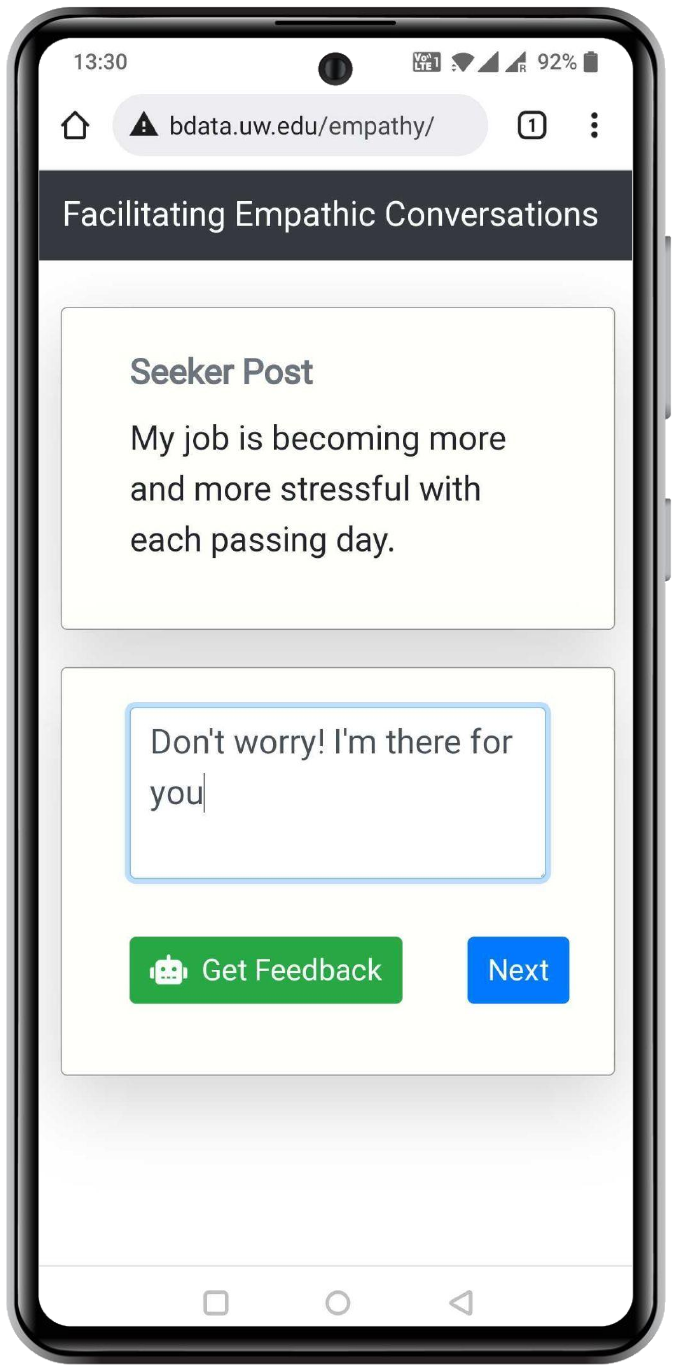} }
\hfill
\subfloat[]{
	\includegraphics[width=0.25\columnwidth]{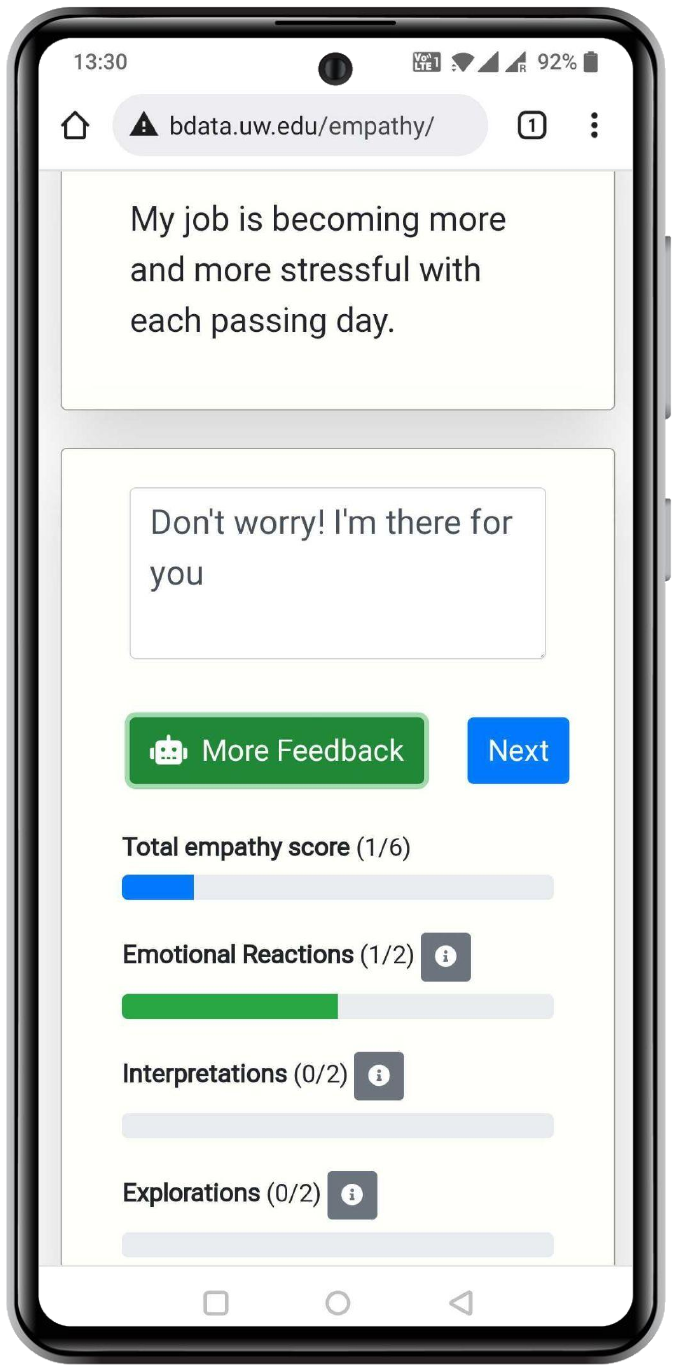} }
\hfill
\vspace{10pt}
\subfloat[]{
	\includegraphics[width=0.25\columnwidth]{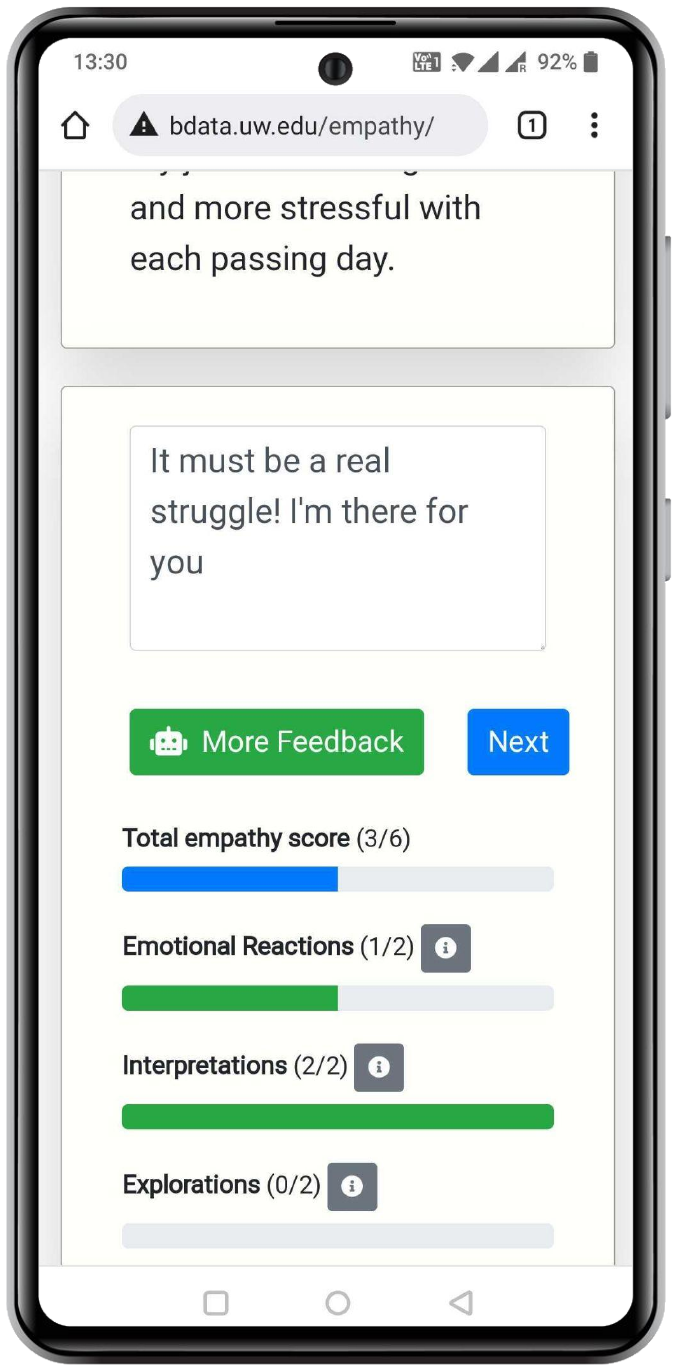} }

\label{supp:figure:study_design:classification}
\end{figure} 

%% file: suppFigures/_supp_figure_empathy_over_time.tex
\begin{figure}
    \caption{Both Human Only (control) and Human + AI (treatment) group participants showed a significant drop in empathy levels in the last 5 responses of our study. With Human + AI, however, we observed a significantly lower drop in empathy (5.34\% vs. 25.99\%; p=0.0062; Two-sided Student's t-test).  This indicates the effectiveness of just-in-time AI feedback in alleviating challenges like empathy fatigue, associated with providing mental health support. The empathy differences between Human Only and Human + AI responses are statistically significant for both first 5 and last 5 responses (p < $10^-5$; Two-sided Student's t-test). Error bars indicate bootstrapped 95\% confidence intervals. }
    \centering
         \includegraphics[width=0.45\textwidth]{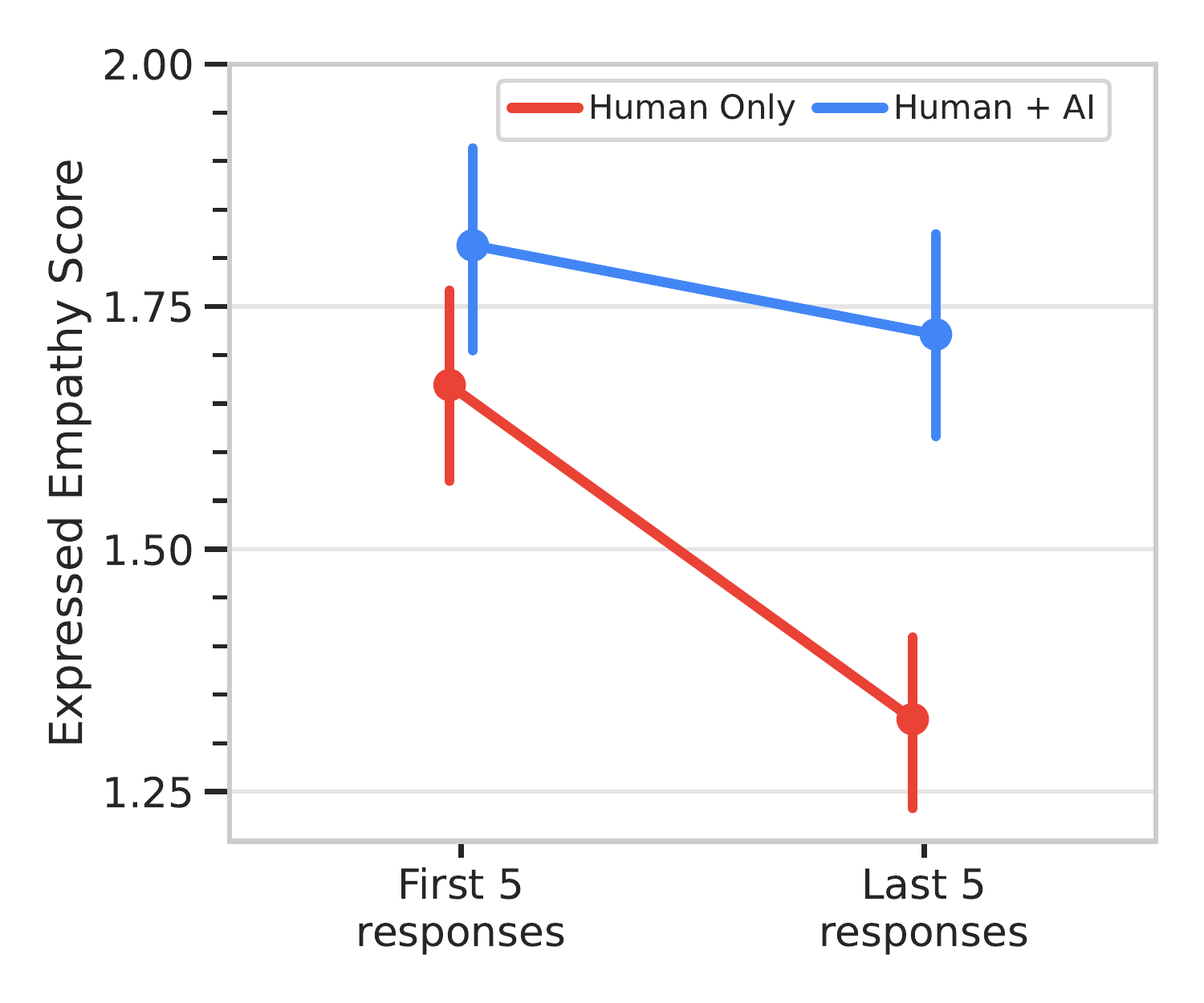}
    \label{supp:figure:empathy_over_time}
\end{figure}

%% file: suppFigures/_supp_figure_talklife_responses.tex
\begin{figure}
\centering
\caption{Comparison of existing Human Only responses on TalkLife with Human Only and Human + AI responses in our study. Human Only responses on TalkLife had significantly lower preference for empathy (18.42\% vs. 37.40\% vs. 46.87\%; p < $10^{-5}$) and significantly lower expressed empathy score (1.11 vs. 1.48 vs. 1.77; p $\ll$ $10^{-5}$). This difference might be attributed to the additional initial empathy training provided to participants, as well as a potential selection effect in our study that may have attracted Talklife users who particularly care about expressing empathy in supporting others. As our study shows that Human-AI collaboration improves empathy expression even for those participants who already express empathy more often, practical gains for the average user of the Talklife platform could be even higher. Error bars indicate bootstrapped 95\% confidence intervals.}
\subfloat[\textbf{Human Evaluation:} Which response is more empathic?]{
	\includegraphics[width=0.45\columnwidth]{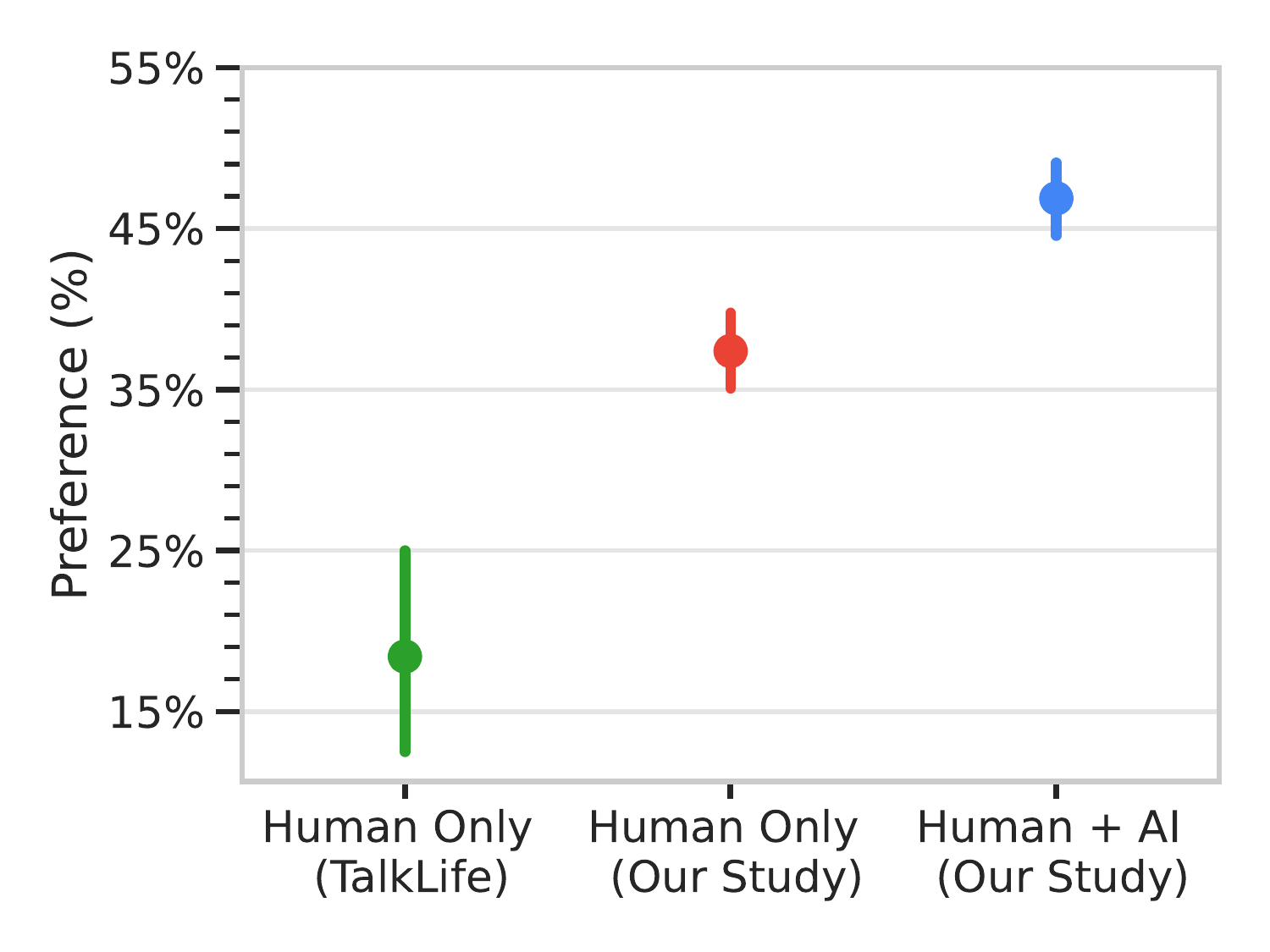} } 
\hfill
\subfloat[\textbf{Automatic/AI-based Evaluation:} Expressed empathy score]{
	\includegraphics[width=0.45\columnwidth]{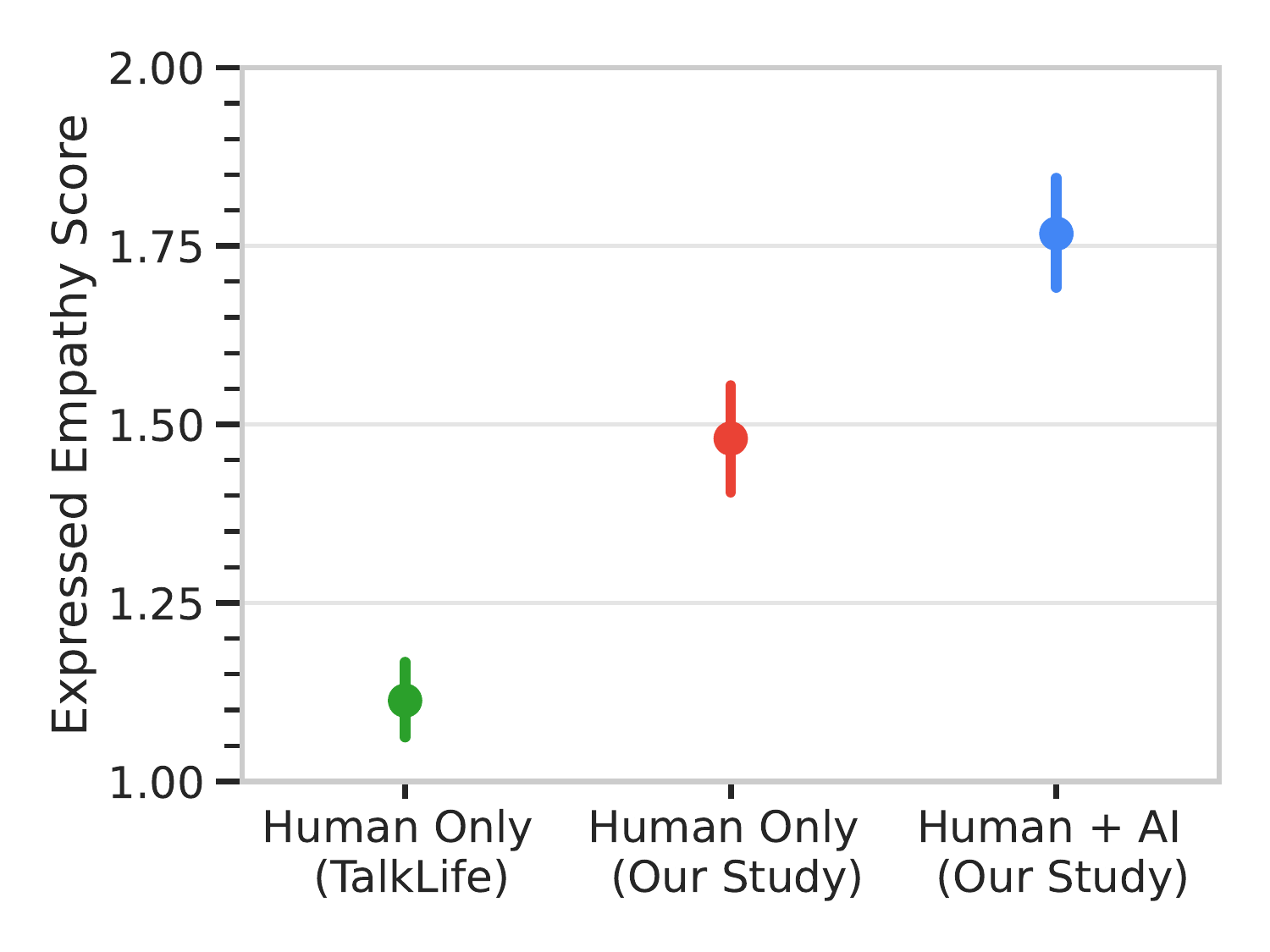} }
\hfill
\label{supp:figure:talklife_response}
\end{figure}

%% file: suppFigures/_supp_figure_qualitative_examples.tex
\begin{figure}
\centering
\caption{Qualitative examples of just-in-time AI feedback provided to participants by \oursystem. In \textbf{(b)} and \textbf{(c)}, the original peer supporter response was empty. Seeker posts in these examples have been paraphrased for anonymization.}
\subfloat[]{
	\includegraphics[width=0.25\columnwidth]{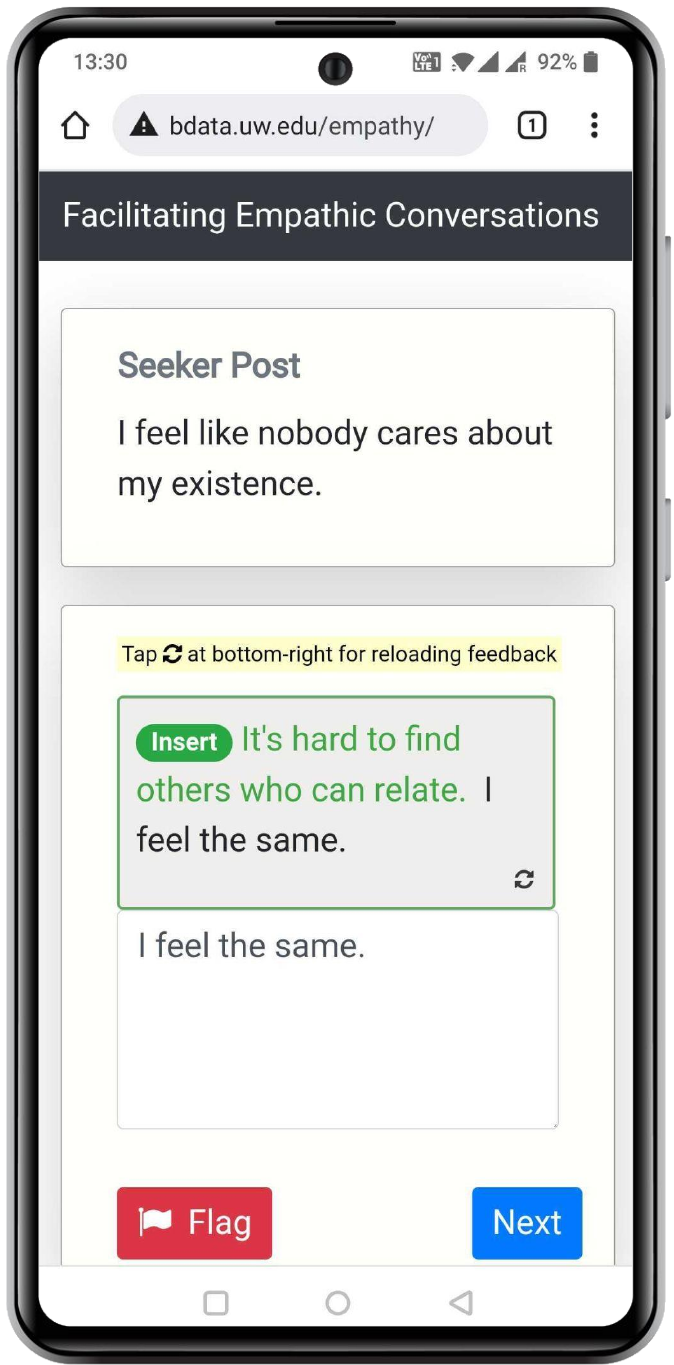} } 
\hfill
\subfloat[]{
	\includegraphics[width=0.25\columnwidth]{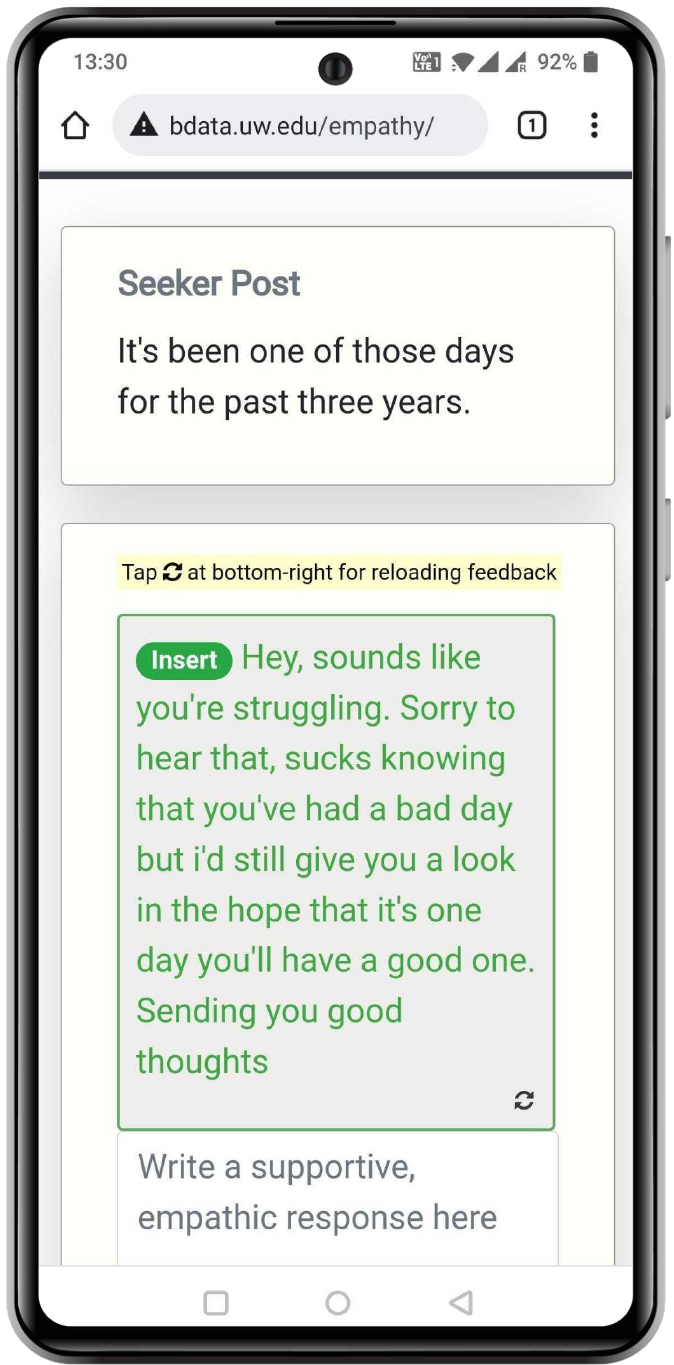} }
\hfill
\subfloat[]{
	\includegraphics[width=0.25\columnwidth]{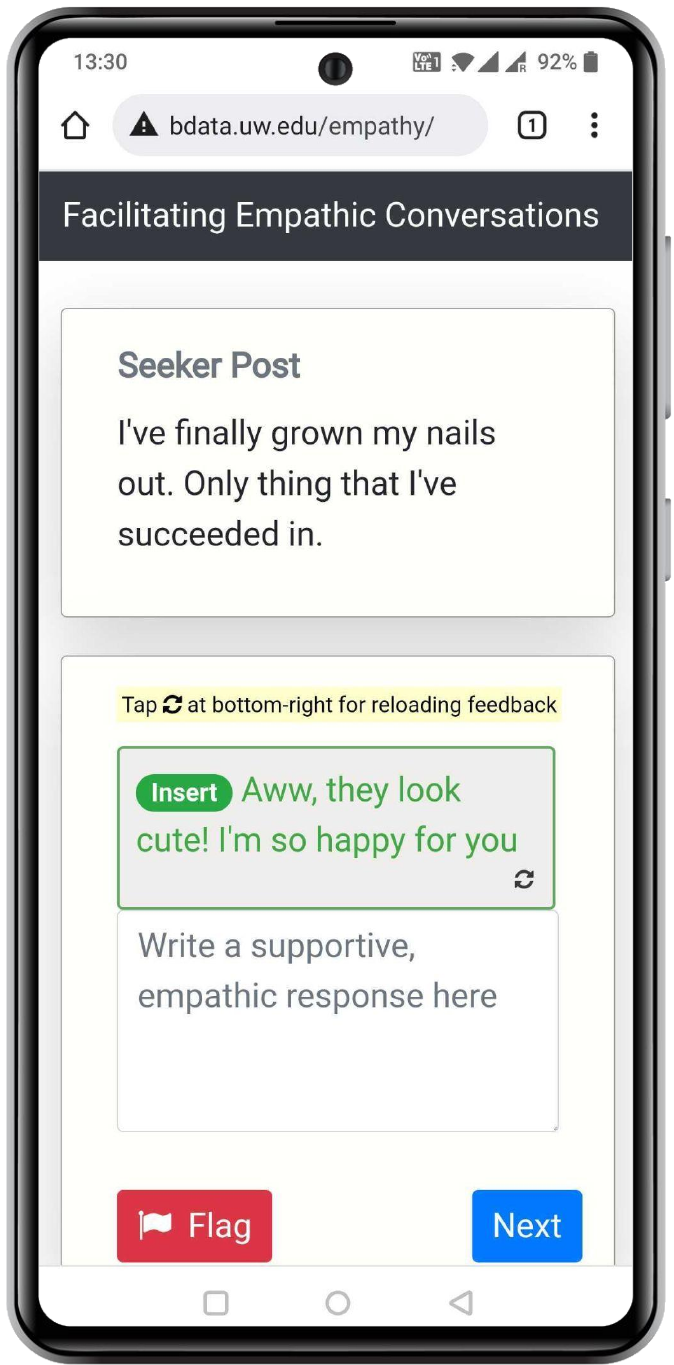} }
\hfill
\vspace{10pt}
\subfloat[]{
	\includegraphics[width=0.25\columnwidth]{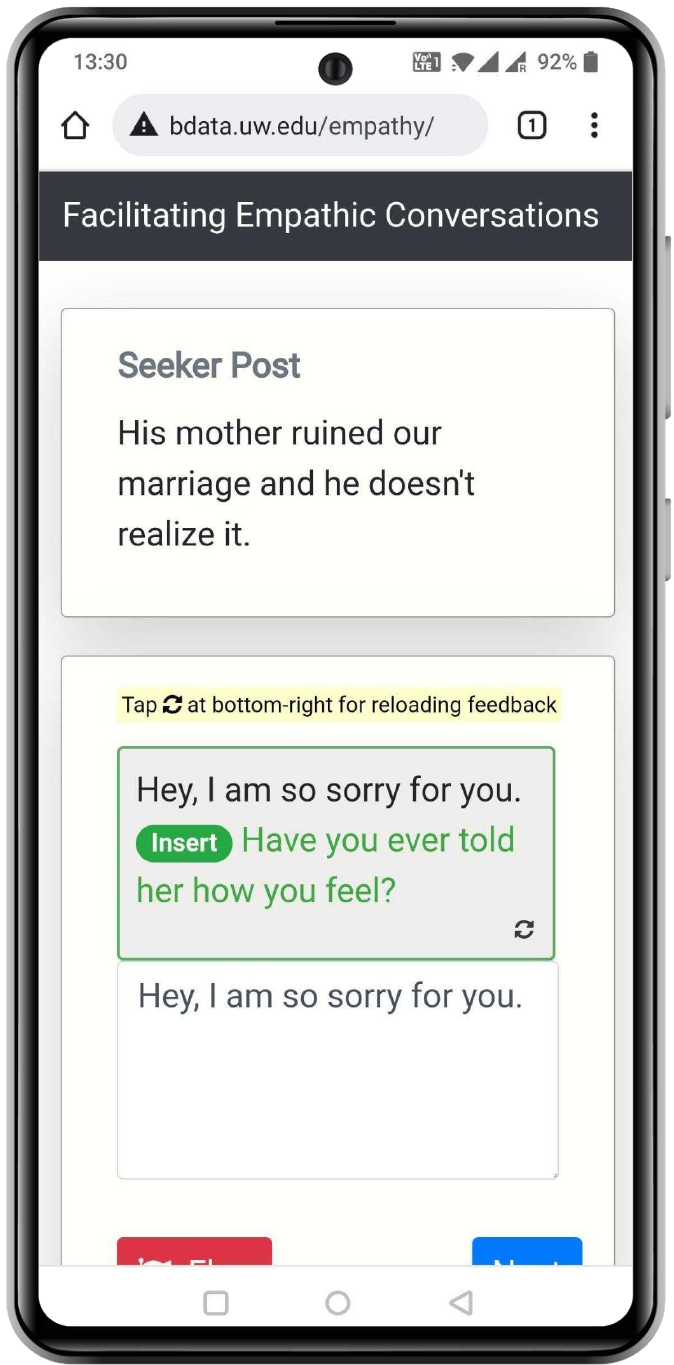} }
\hfill
\subfloat[]{
	\includegraphics[width=0.25\columnwidth]{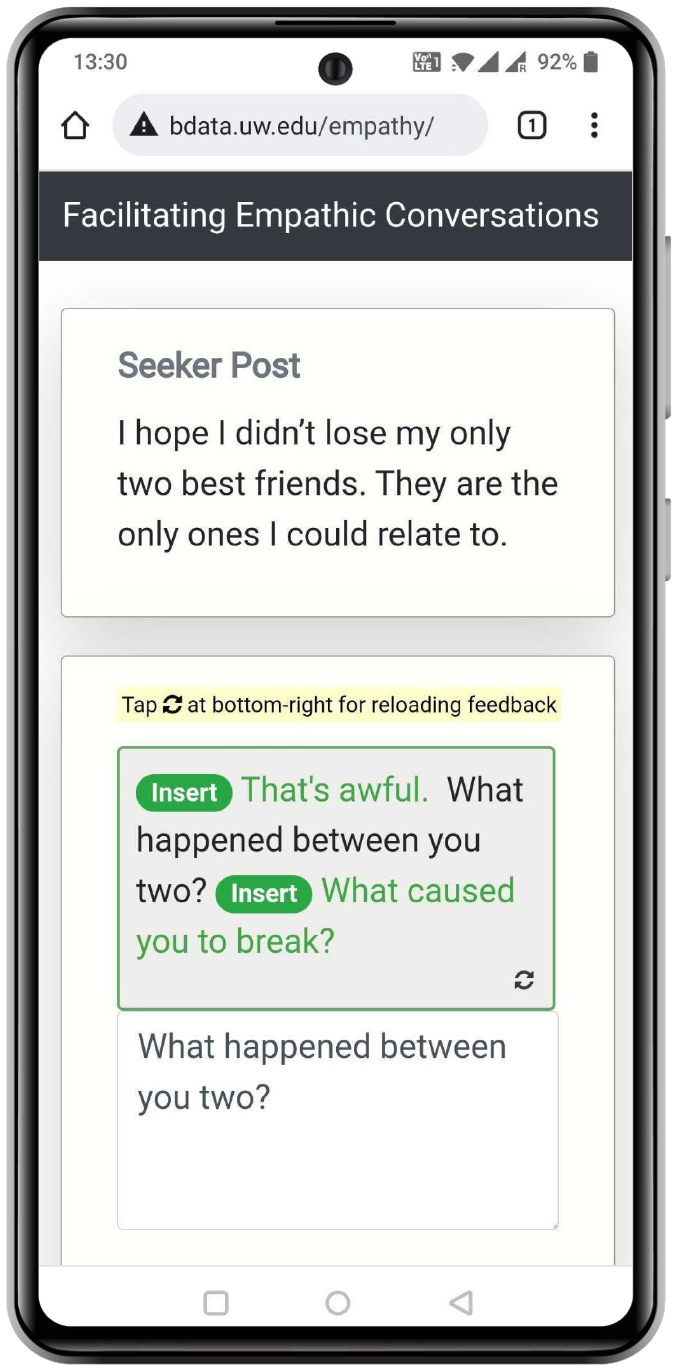} }
\hfill
\subfloat[]{
	\includegraphics[width=0.25\columnwidth]{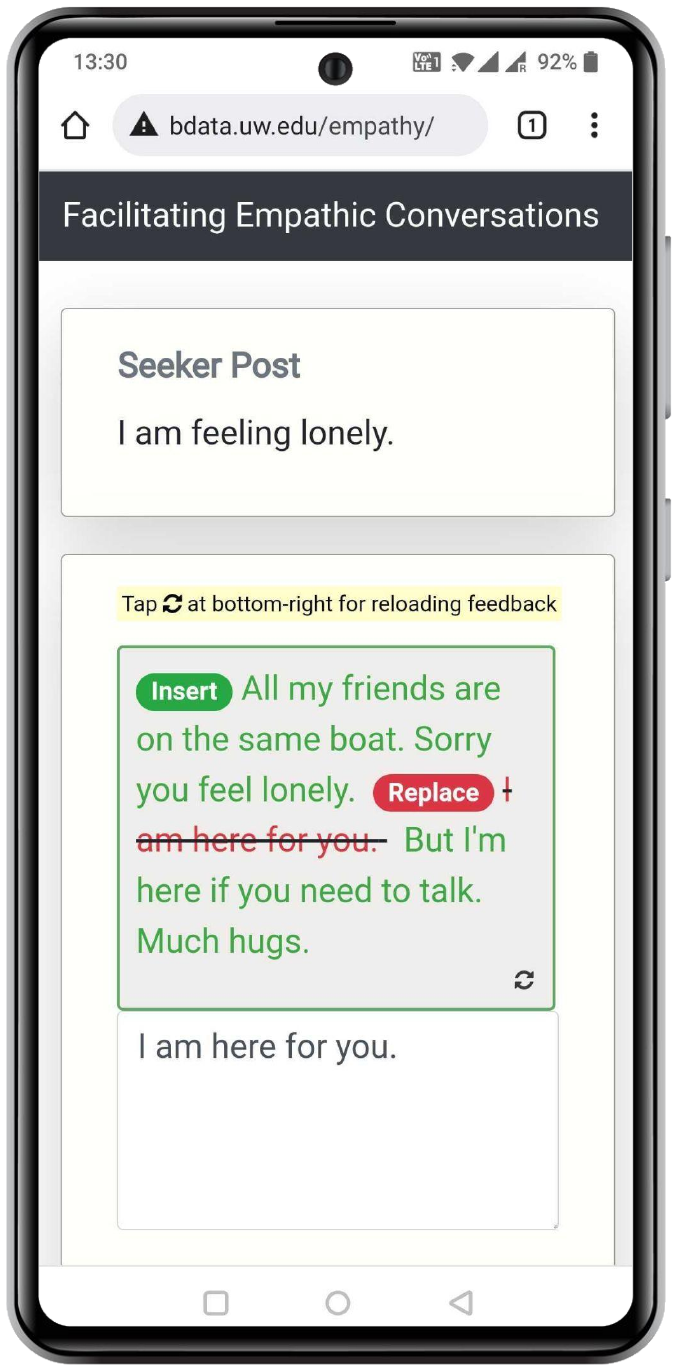} }
\hfill
\label{supp:figure:qualitative}
\end{figure} 

%% file: suppFigures/_supp_figure_hai_empathy.tex
\begin{figure}
    \caption{Human evaluation of empathy of  participants with different human-AI collaboration categories. Responses from participants who consulted or used AI more, broadly, had a higher empathy preference compared to those who rarely, if ever, expressed it (confirming the findings from the automatic evaluation in Figure~\ref{figure:3}b). The area of the points is proportional to the number of participants in the respective human-AI collaboration categories. Error bars indicate bootstrapped 95\% confidence intervals.}
    \centering
         \includegraphics[width=0.7\textwidth]{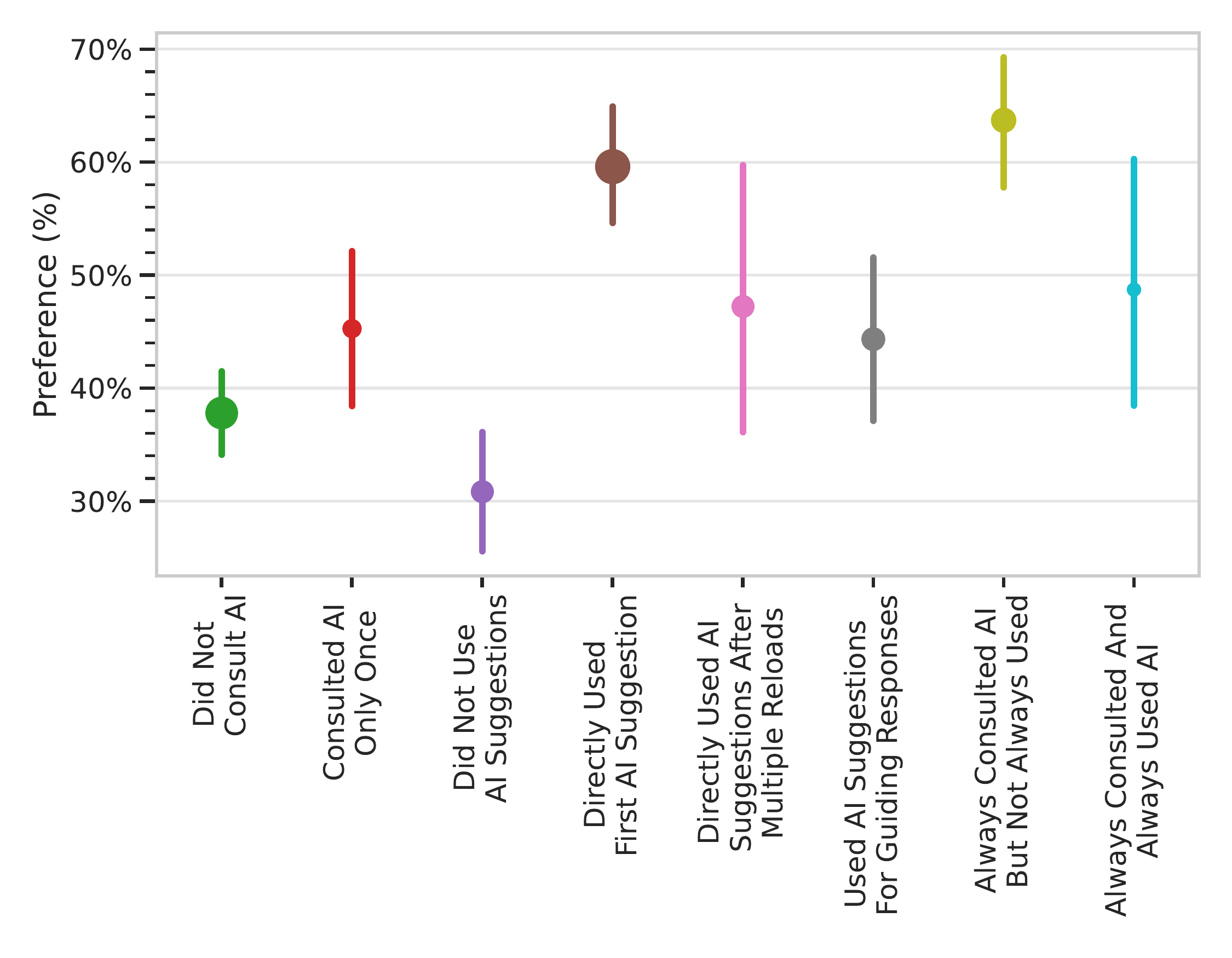}
    \label{supp:figure:hai_empathy}
\end{figure}

%% file: suppFigures/_supp_figure_background.tex
\begin{figure}
\centering
\caption{Background and demographics of participants in Human Only (control) and Human + AI (treatment) groups, as reported in phase I (pre-intervention survey).}
\subfloat[Gender]{
	\includegraphics[width=0.45\columnwidth]{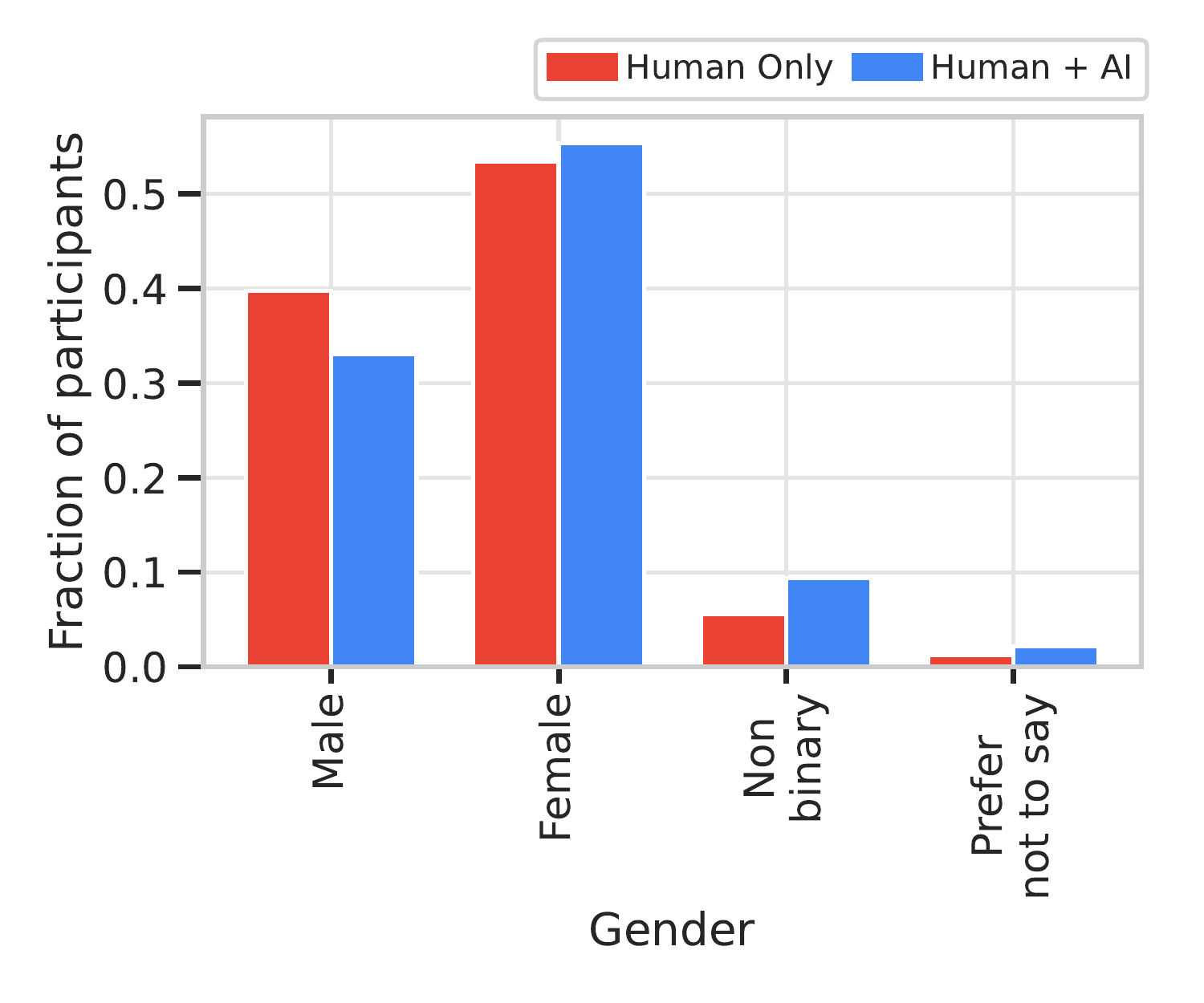} } 
\hfill
\subfloat[Age]{
	\includegraphics[width=0.45\columnwidth]{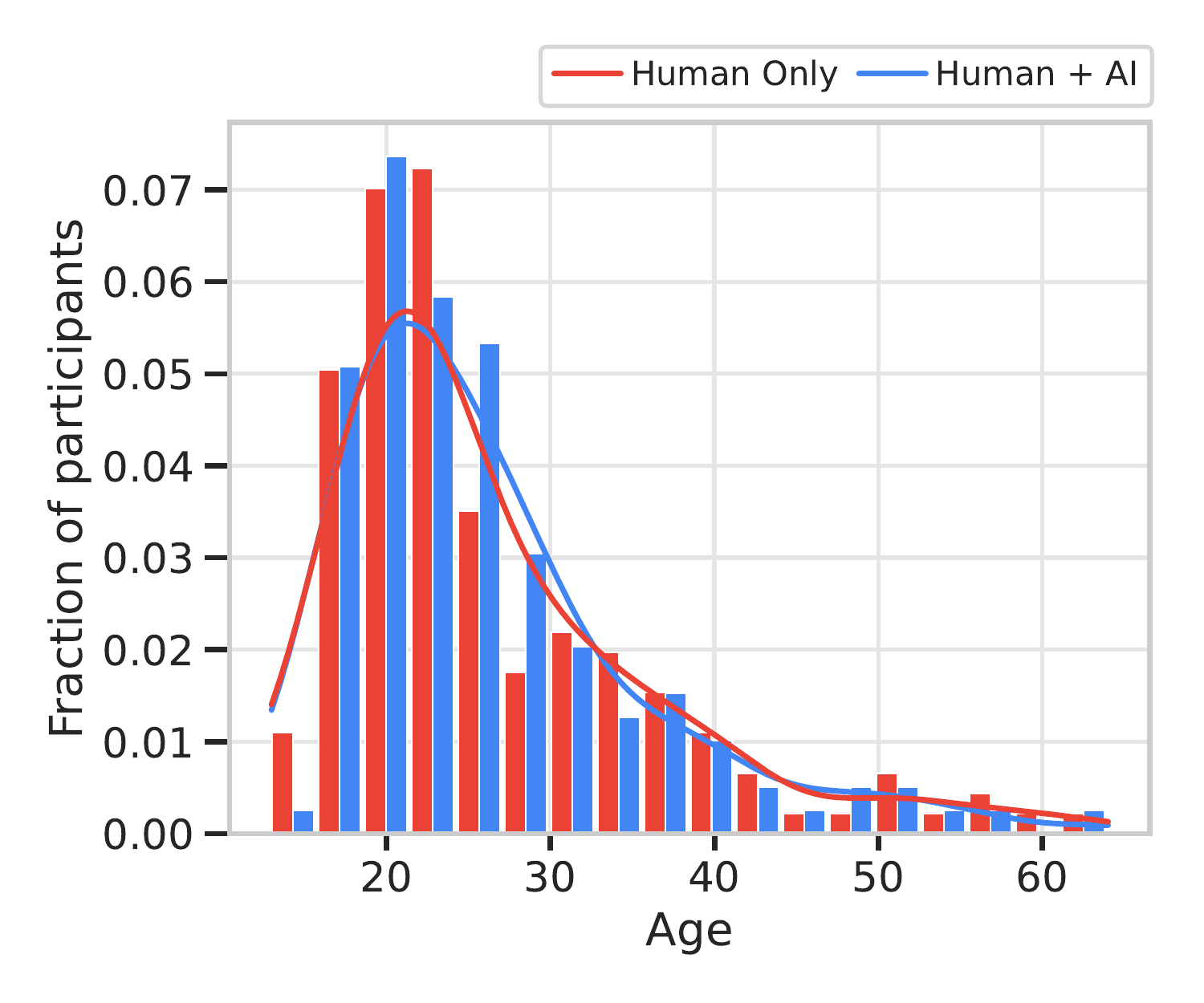} } 
\hfill
\subfloat[Race/Ethnicity]{
	\includegraphics[width=0.45\columnwidth]{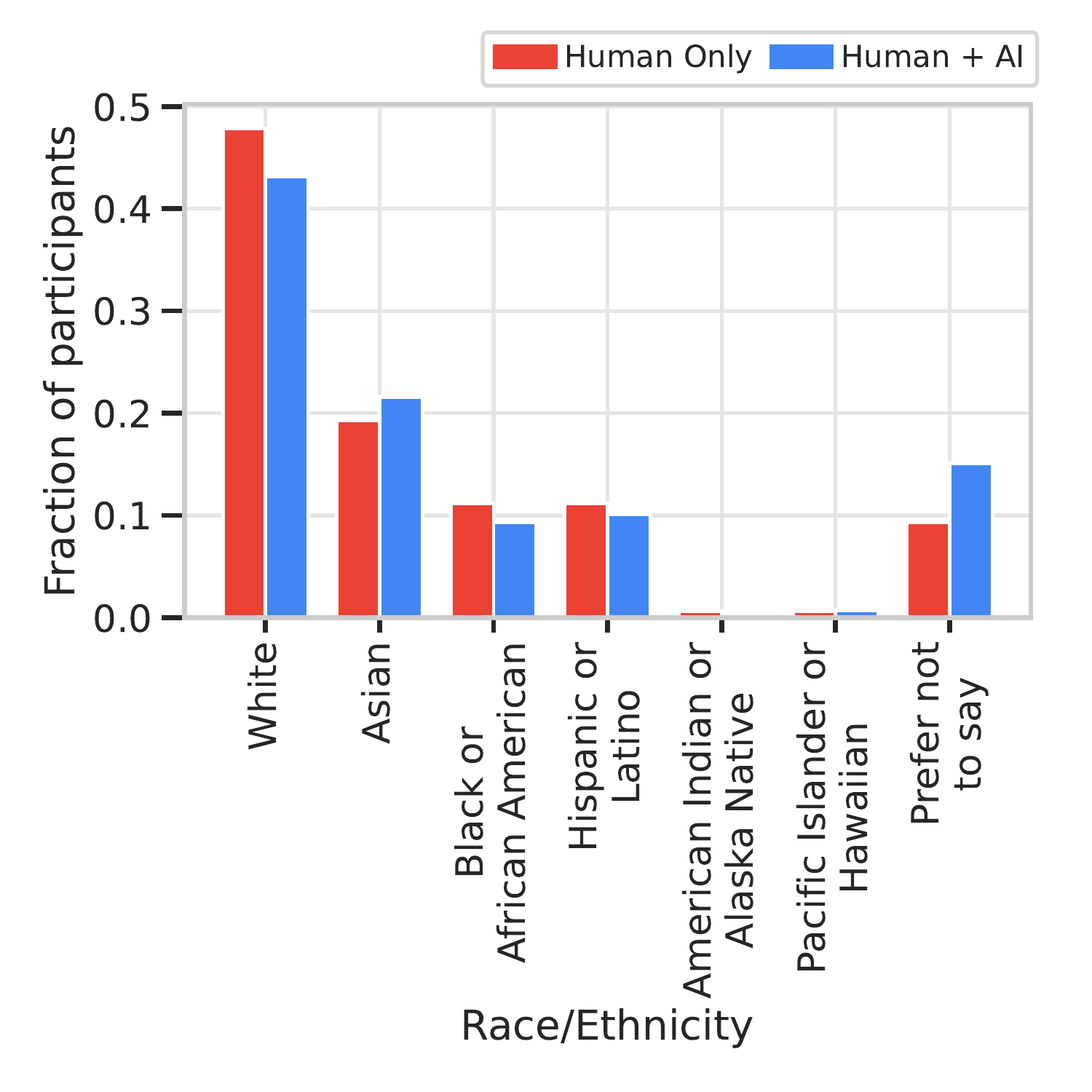} }
\hfill
\subfloat[Experience with online peer support]{
	\includegraphics[width=0.45\columnwidth]{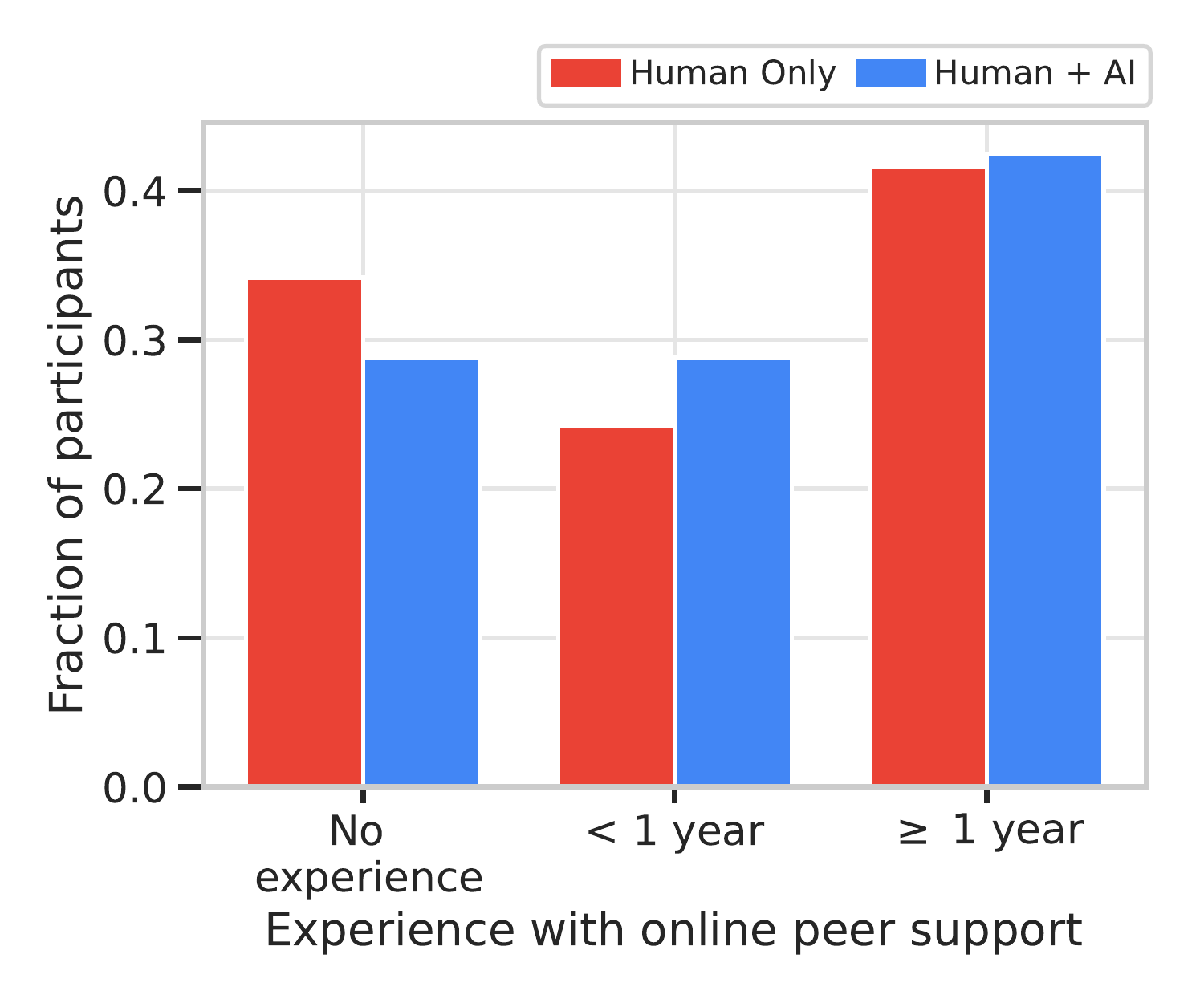} }
\label{supp:figure:background}
\end{figure}

%% file: suppFigures/_supp_figure_background_empathy.tex
\begin{figure}
\centering
\caption{Differences between expressed empathy scores of participants in Human Only (control) and Human + AI (treatment) groups, stratified by demographics of participants and their prior experience with online peer support. The area of the points is proportional to the number of participants in the respective human-AI collaboration categories. Error bars indicate bootstrapped 95\% confidence intervals.}
\subfloat[Gender]{
	\includegraphics[width=0.45\columnwidth]{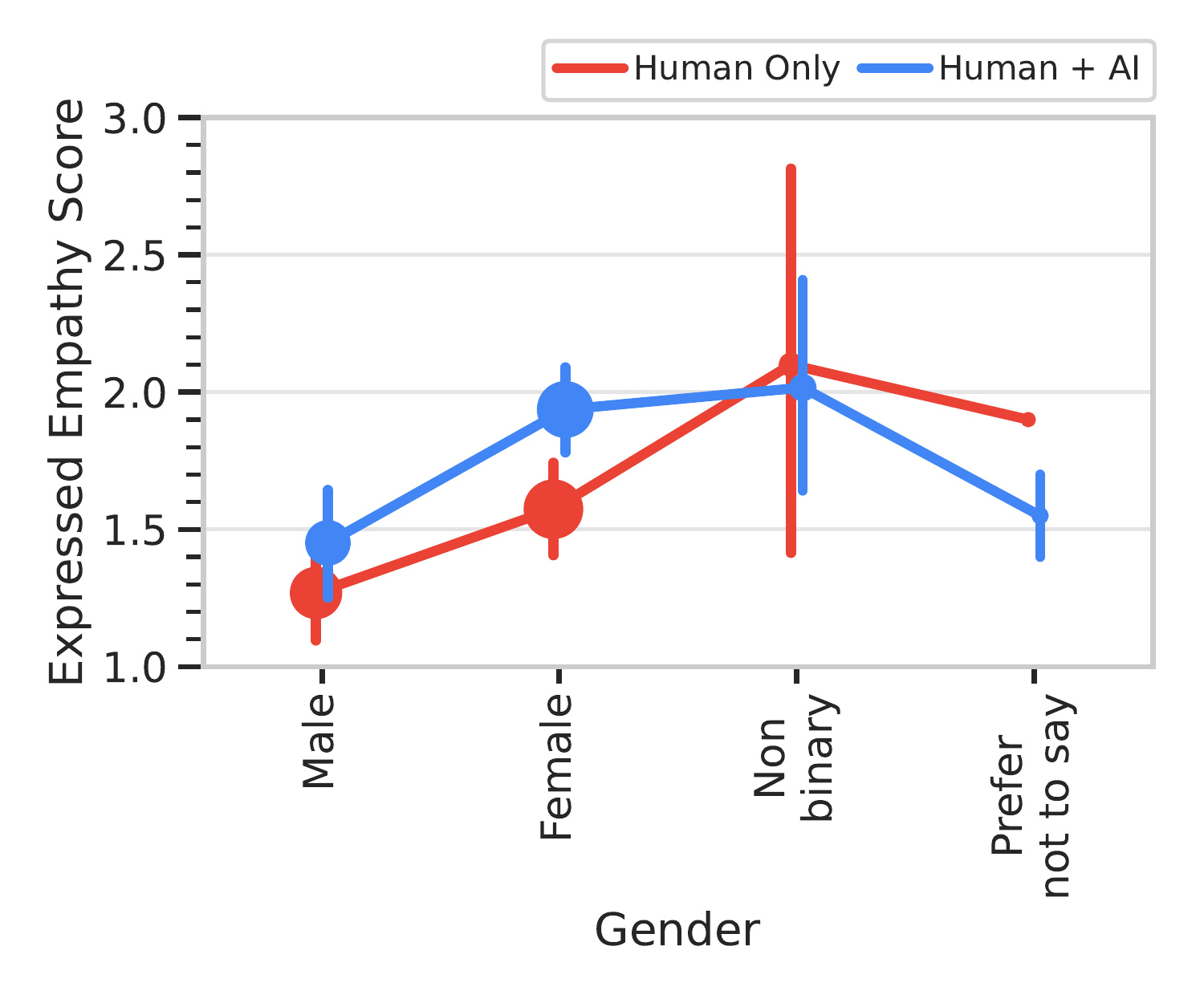} } 
\hfill
\subfloat[Age]{
	\includegraphics[width=0.45\columnwidth]{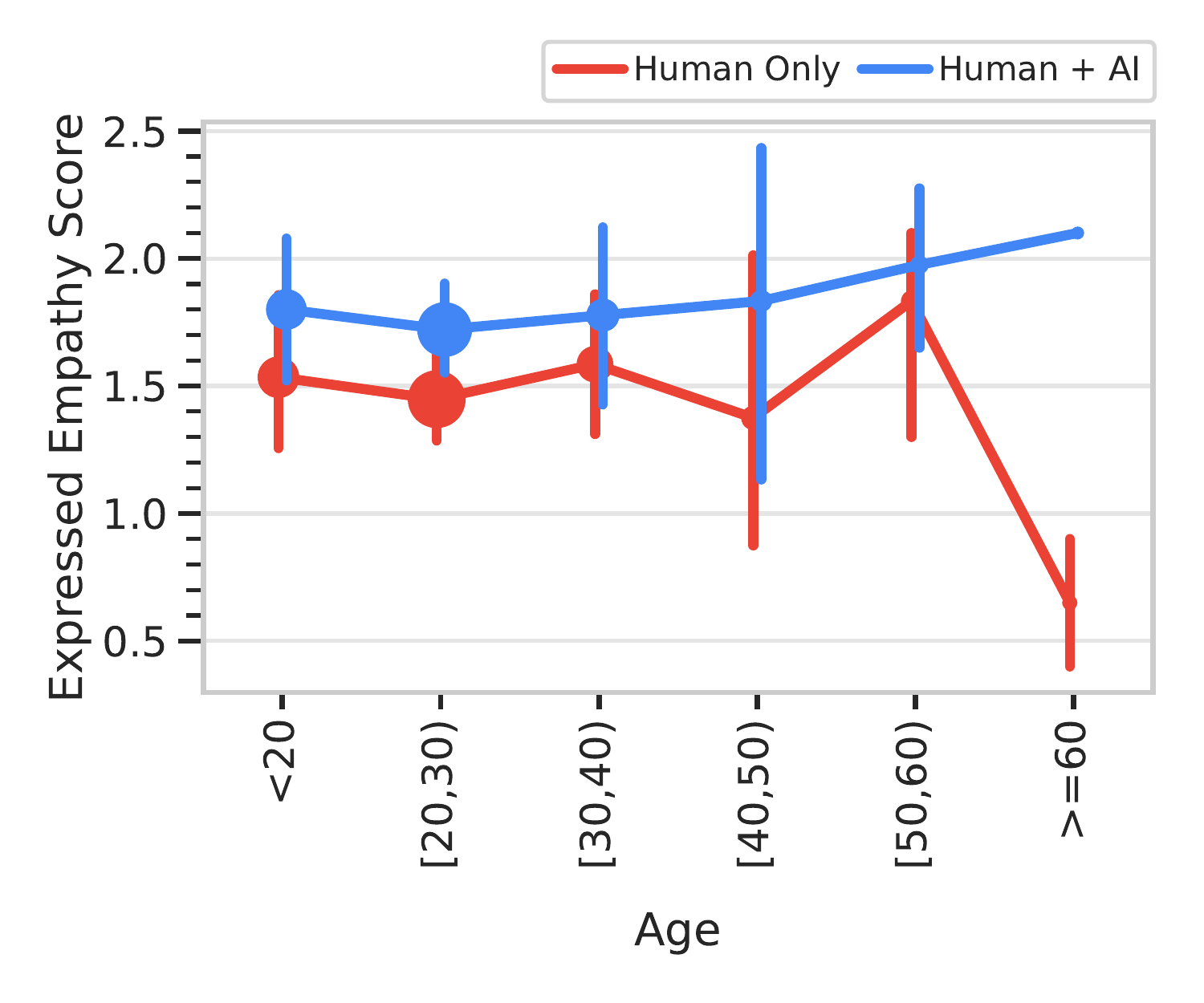} } 
\hfill
\subfloat[Race/Ethnicity]{
	\includegraphics[width=0.45\columnwidth]{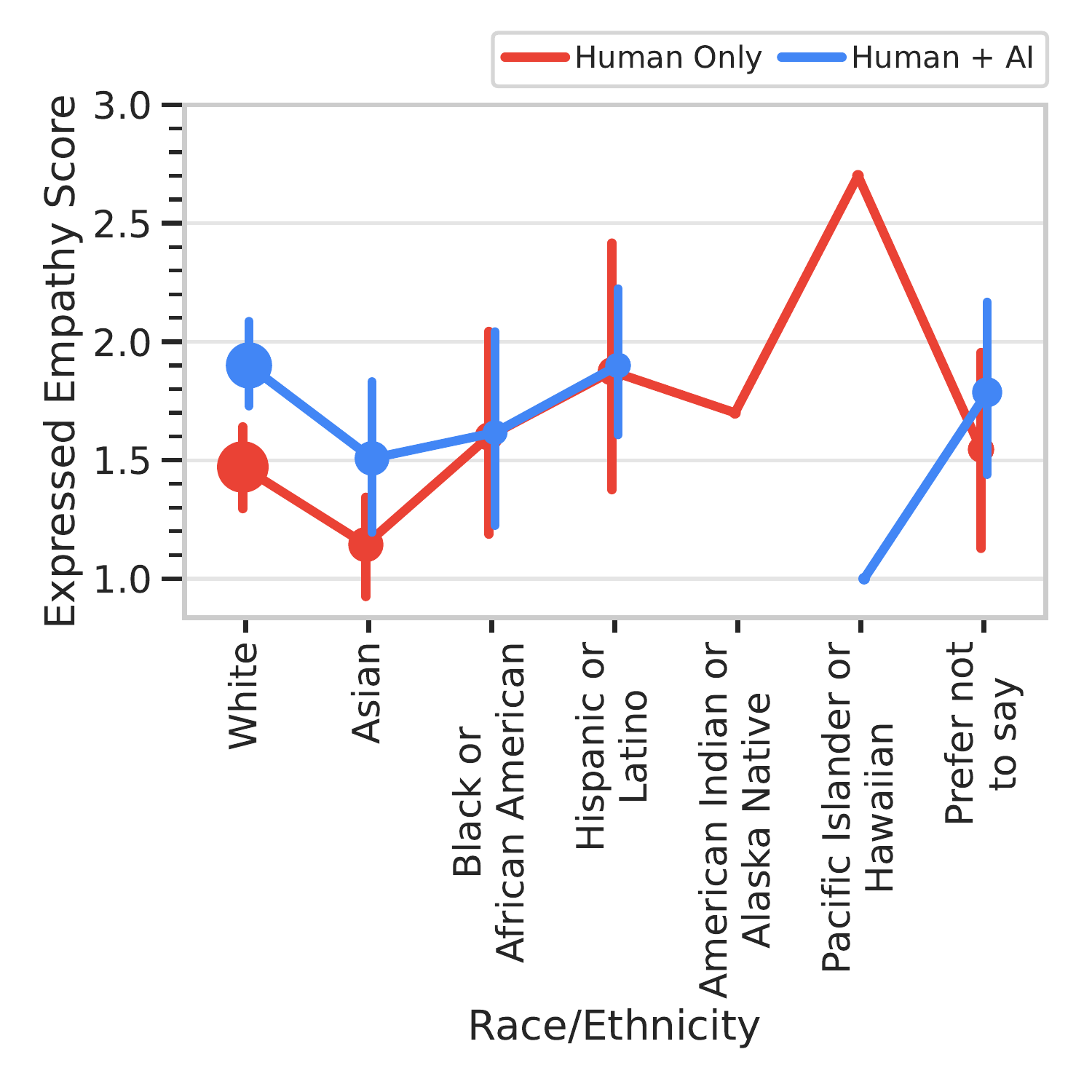} }
\hfill
\subfloat[Prior experience with online peer support]{
	\includegraphics[width=0.45\columnwidth]{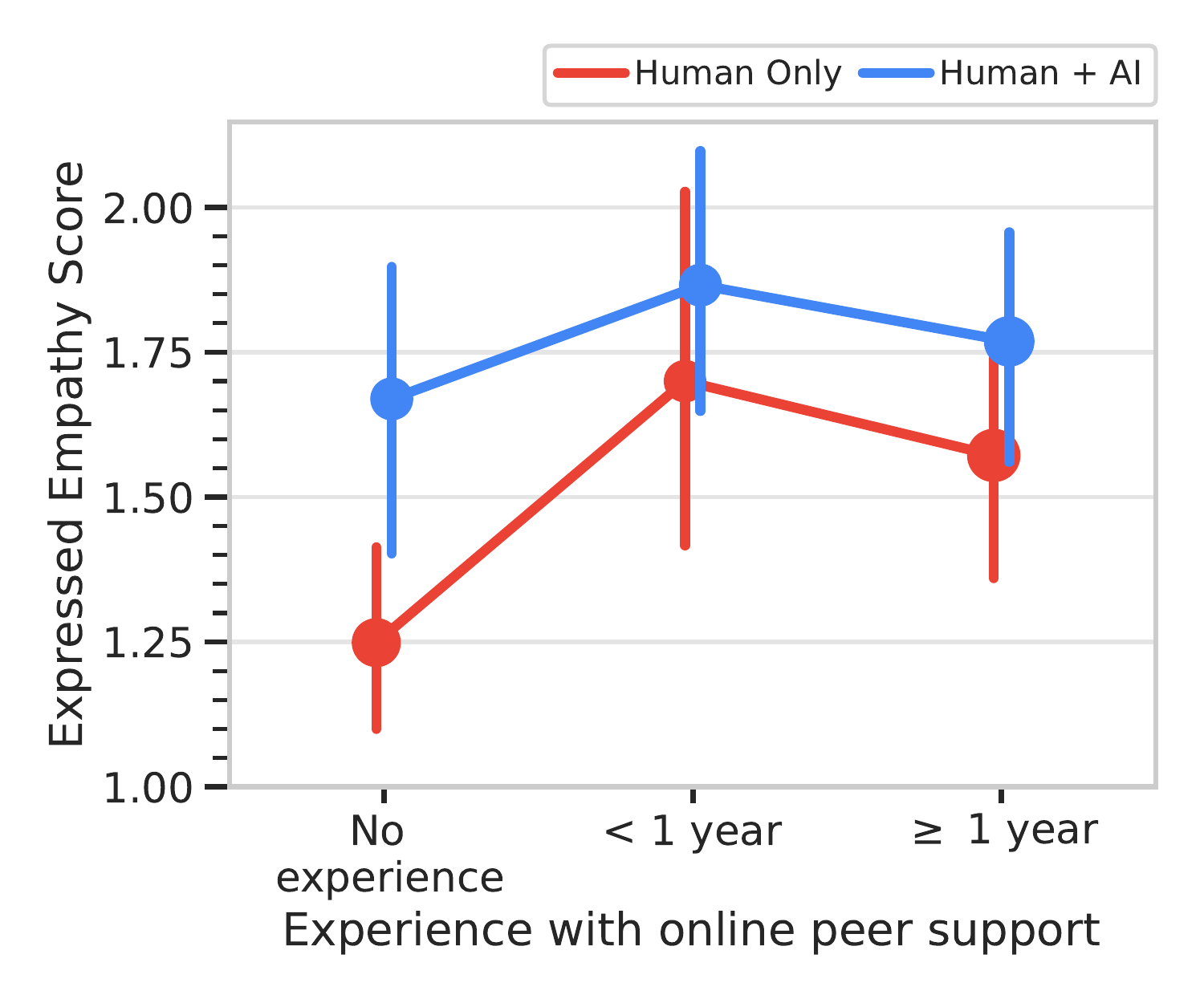} }
\label{supp:figure:background_empathy}
\end{figure}

%% file: suppFigures/_supp_figure_onboarding.tex
\begin{figure}
\centering
\caption{Perceptions of participants in Human Only (control) and Human + AI (treatment) groups, as reported in phase I (pre-intervention survey).}
\subfloat[Self-efficacy in writing good, effective, or helpful responses]{
	\includegraphics[width=0.45\columnwidth]{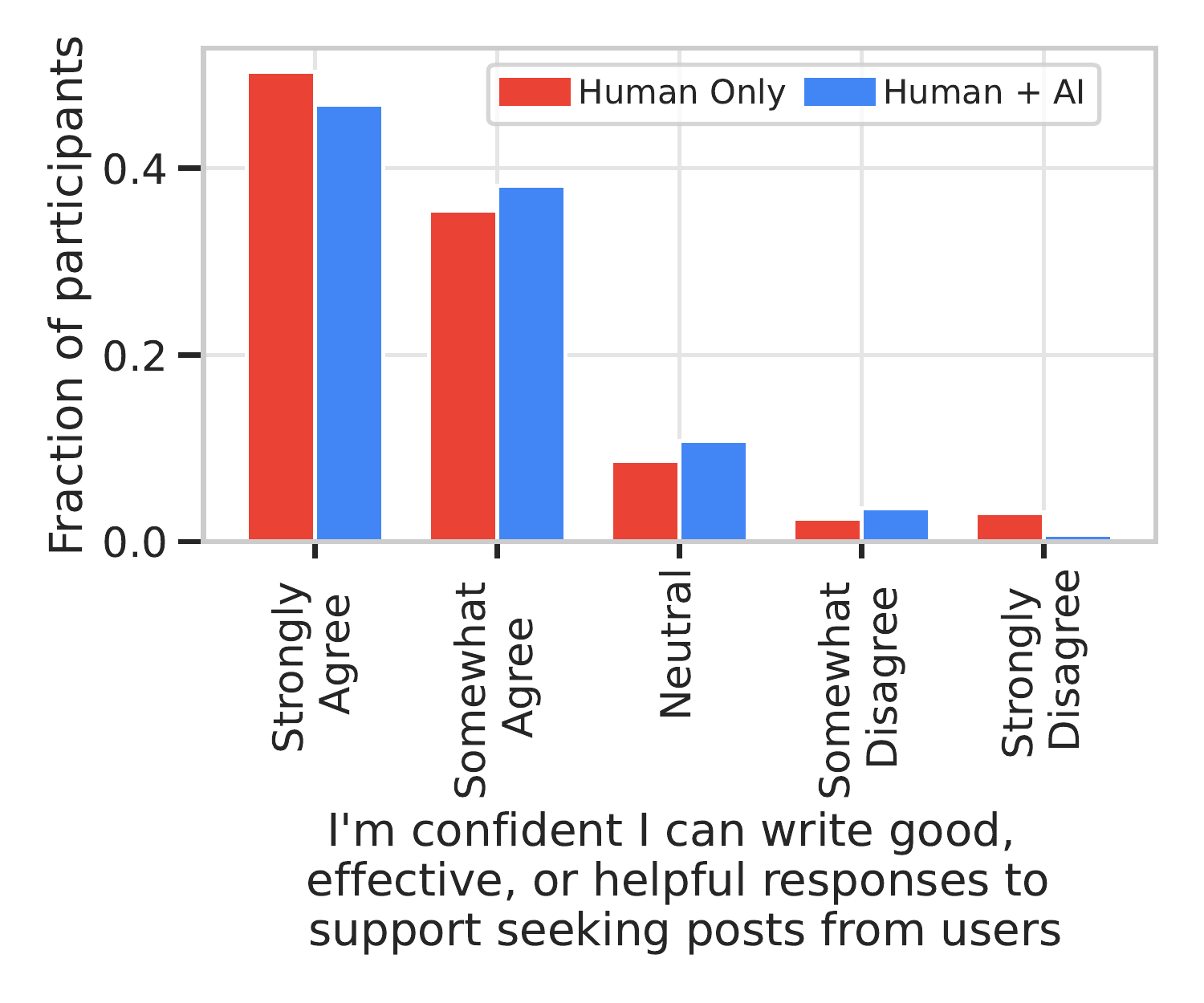} } 
\hfill
\subfloat[Self-efficacy in writing empathic responses]{
	\includegraphics[width=0.45\columnwidth]{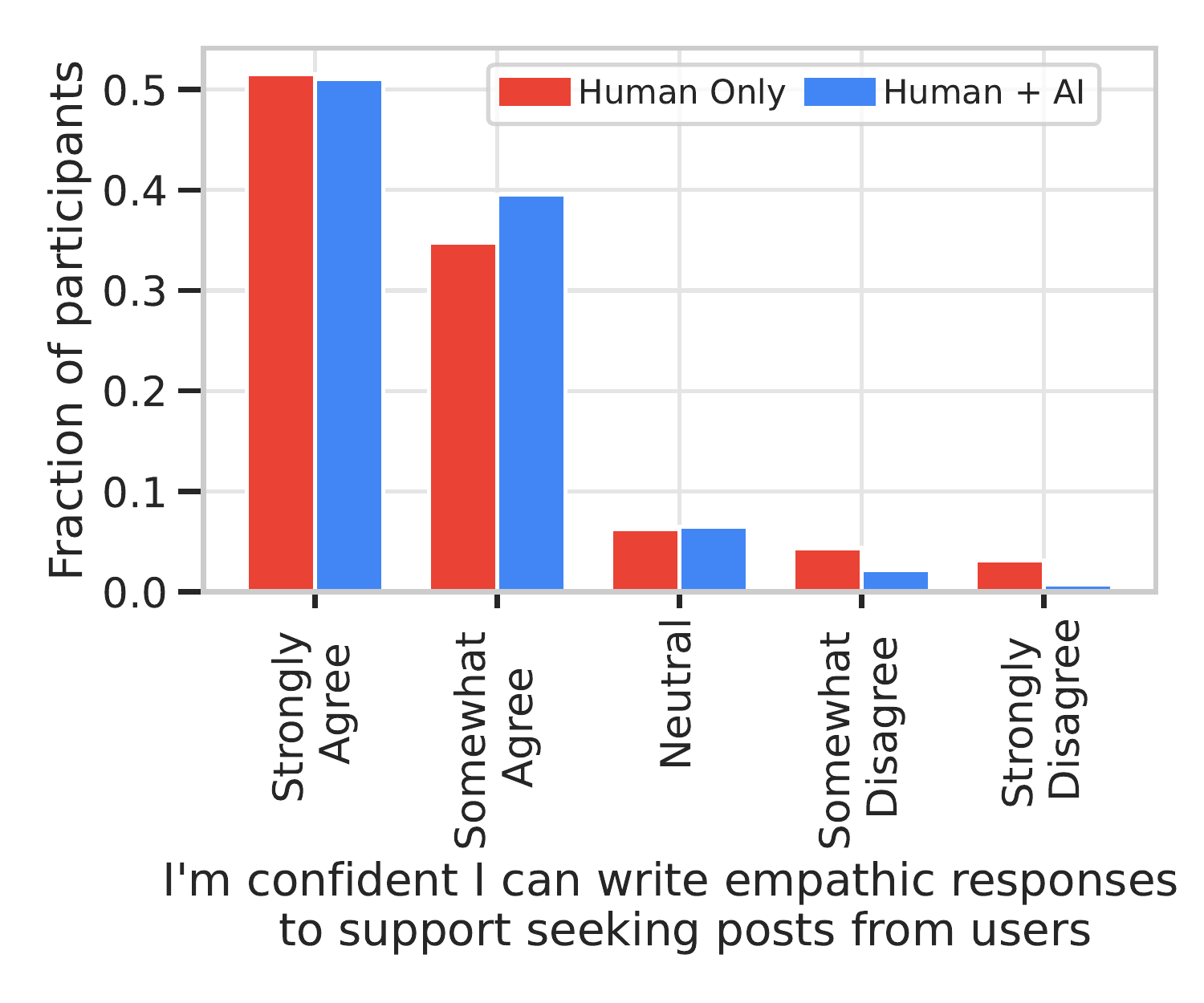} } 
\hfill
\subfloat[Could feedback be helpful?]{
	\includegraphics[width=0.45\columnwidth]{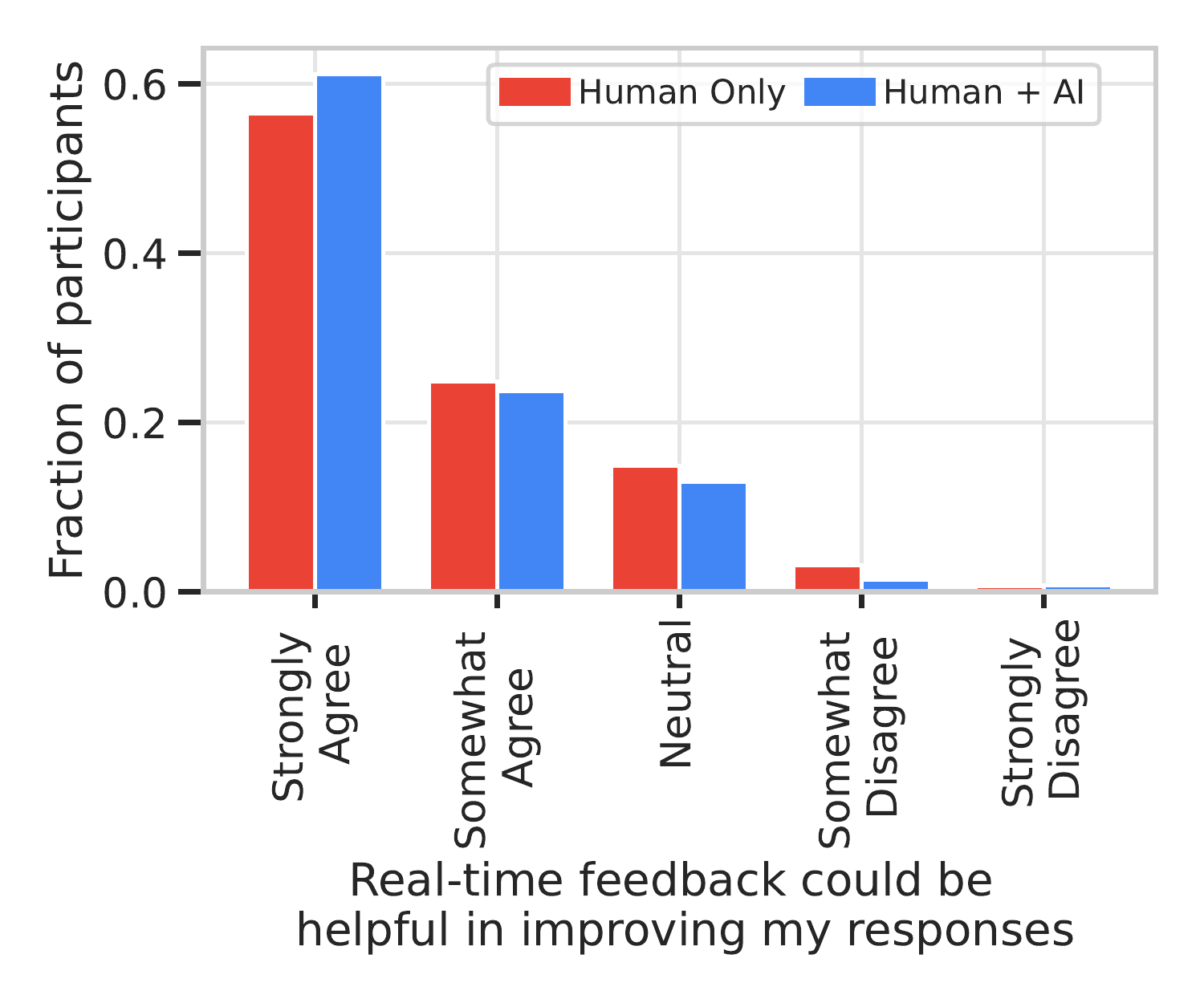} }
\label{supp:figure:onboarding}
\end{figure} 

%% file: suppFigures/_supp_figure_exit_survey_challenging.tex
\begin{figure}
\centering
\caption{Distribution of participants in Human Only (control) and Human + AI (treatment) groups who report writing responses as challenging or stressful, as reported in phase IV (post-intervention survey).}
	\includegraphics[width=0.45\columnwidth]{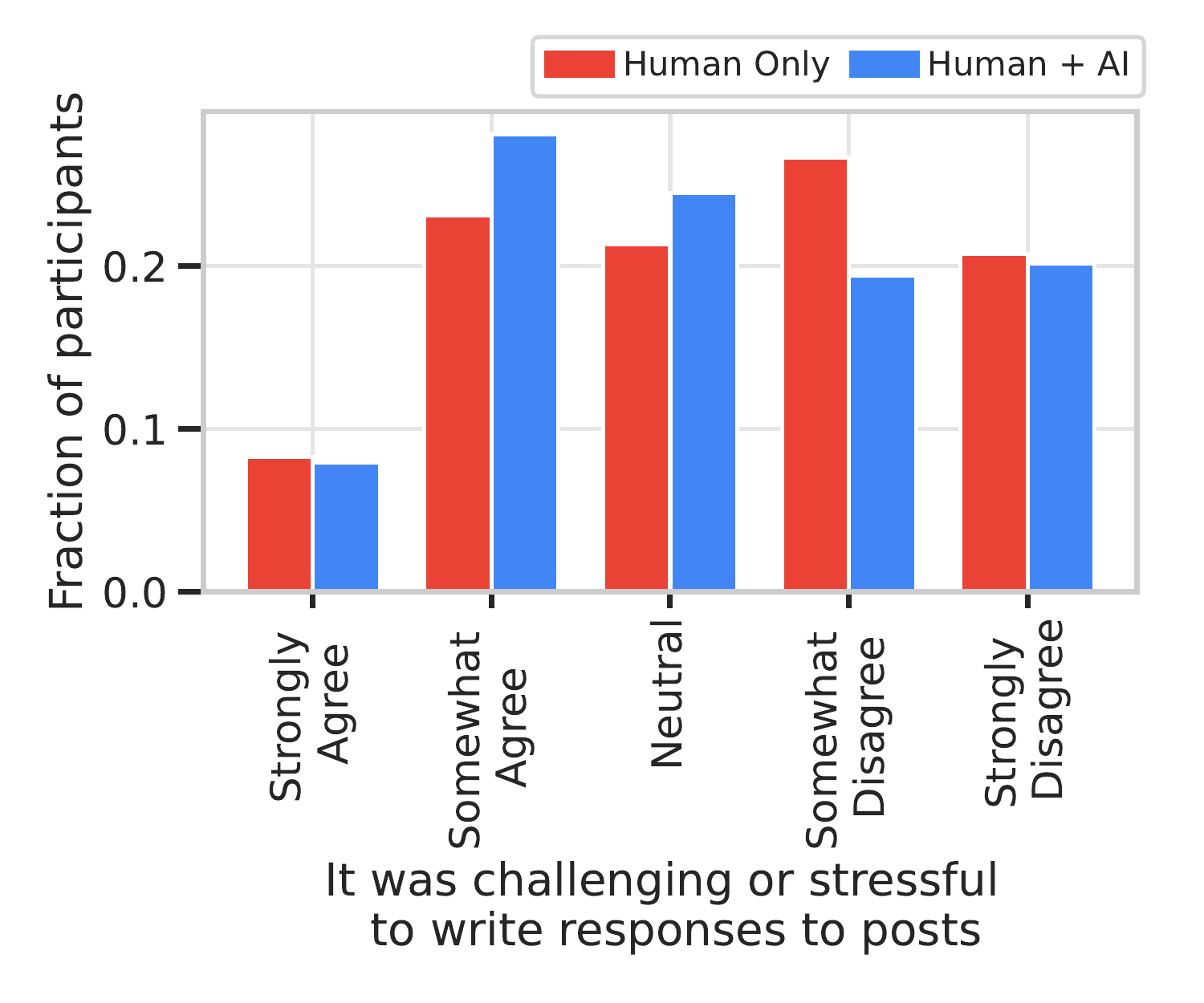}
\label{supp:figure:exit_survey_challenging}
\end{figure} 

%% file: suppFigures/_supp_figure_exit_survey_feedback_would_improve.tex
\begin{figure}
\centering
\caption{Distribution of participants in the Human Only (control) group who indicate that feedback could have improved responses, as reported in phase IV (post-intervention survey).}
	\includegraphics[width=0.45\columnwidth]{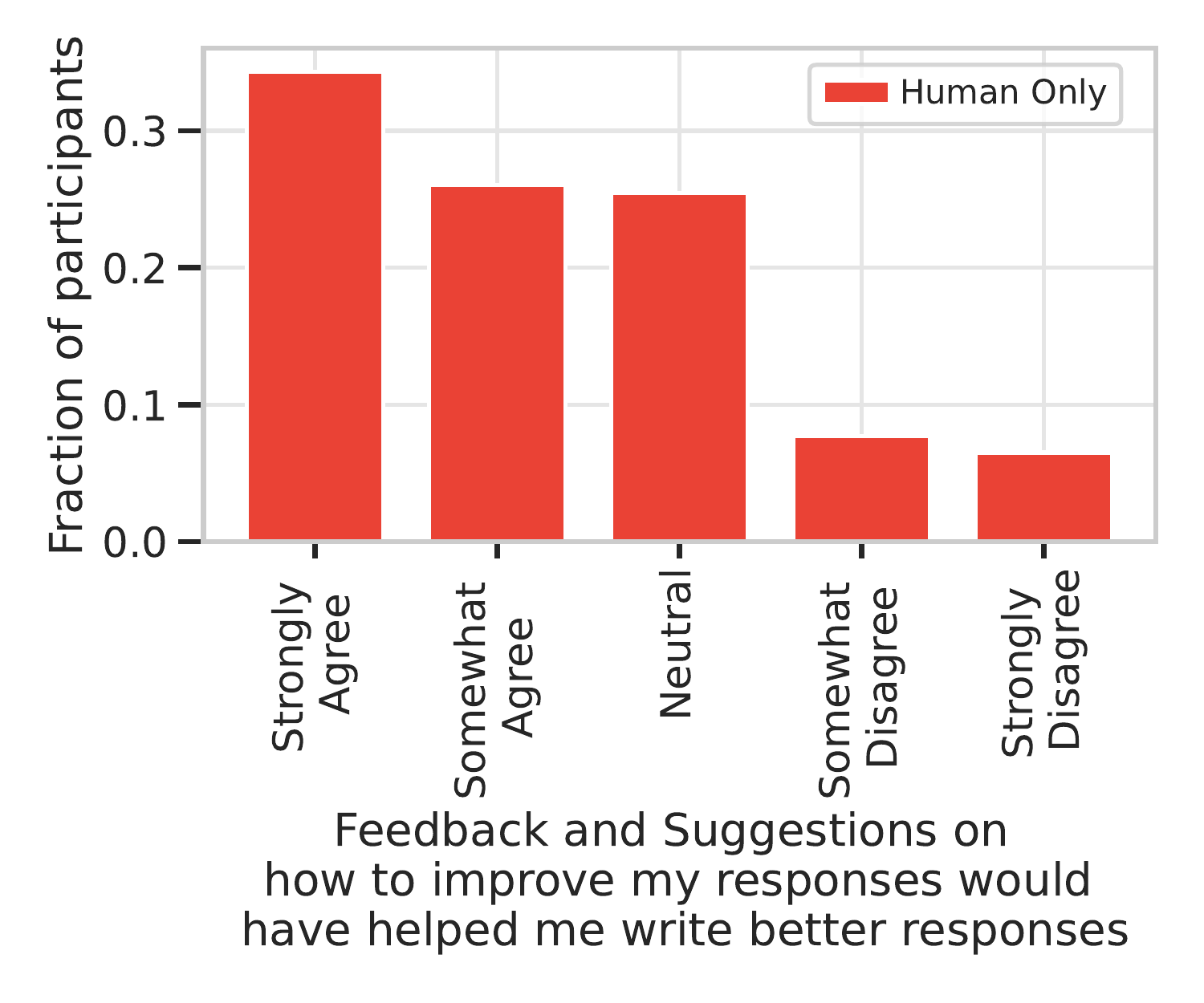}
\label{supp:figure:exit_survey_challenging}
\end{figure} 

%% file: suppFigures/_supp_figure_exit_survey_treatment.tex
\begin{figure}
\centering
\caption{Perceptions of participants in the Human + AI (treatment) group, as reported in phase IV (post-intervention survey).}
\subfloat[Overall helpfulness]{
	\includegraphics[width=0.45\columnwidth]{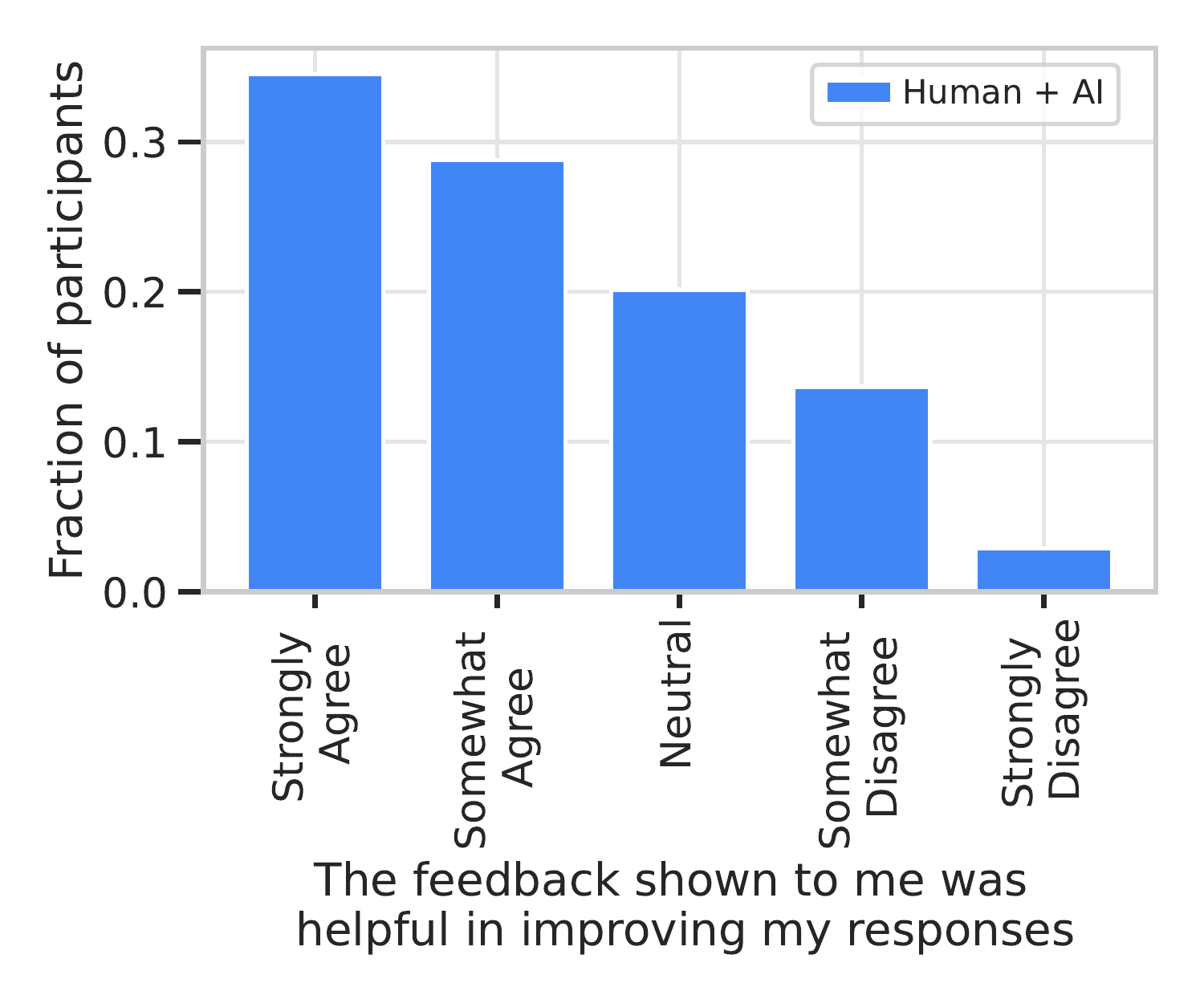} } 
\hfill
\subfloat[Helpfulness for empathy]{
	\includegraphics[width=0.45\columnwidth]{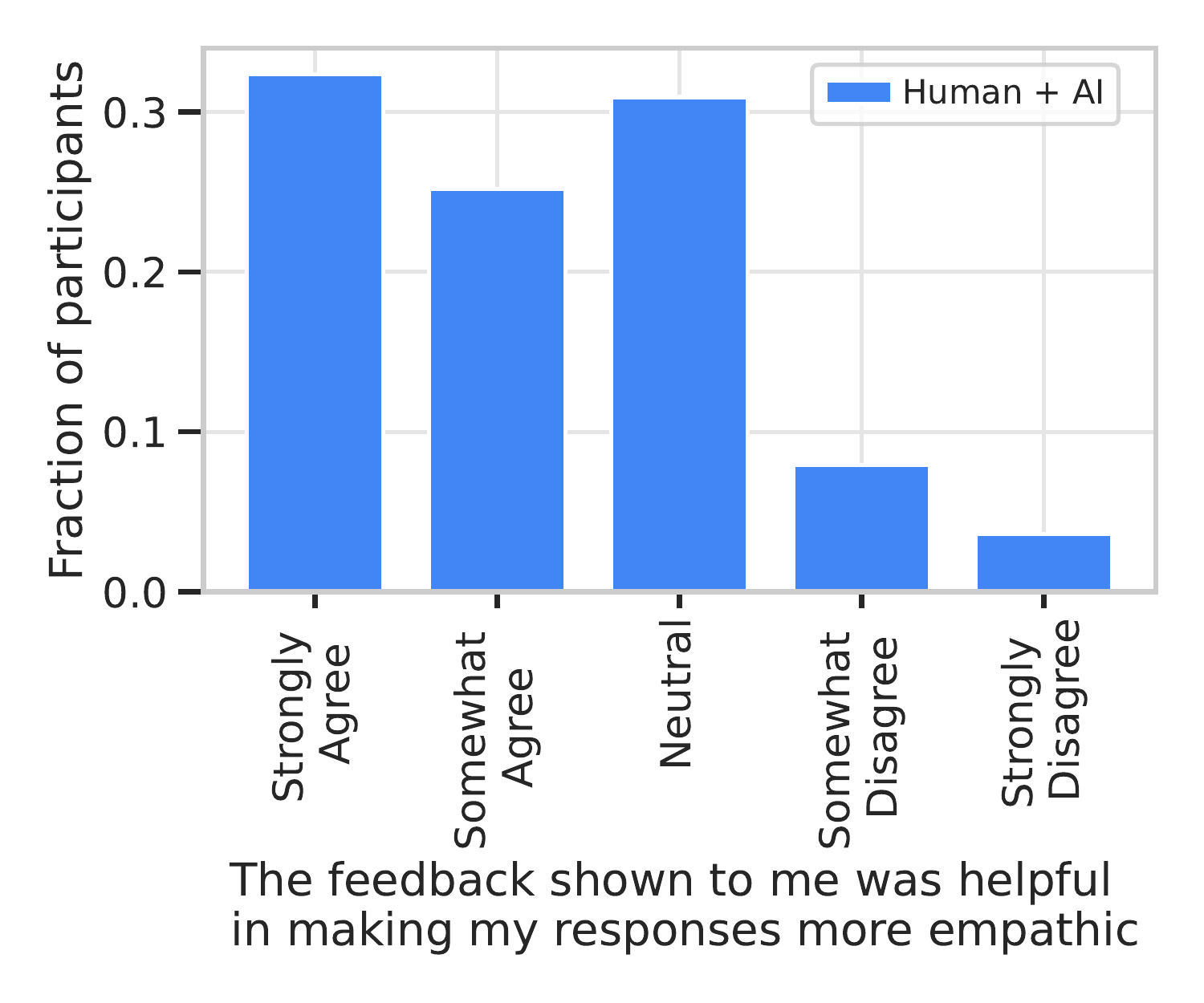} } 
\hfill
\subfloat[Actionability]{
	\includegraphics[width=0.45\columnwidth]{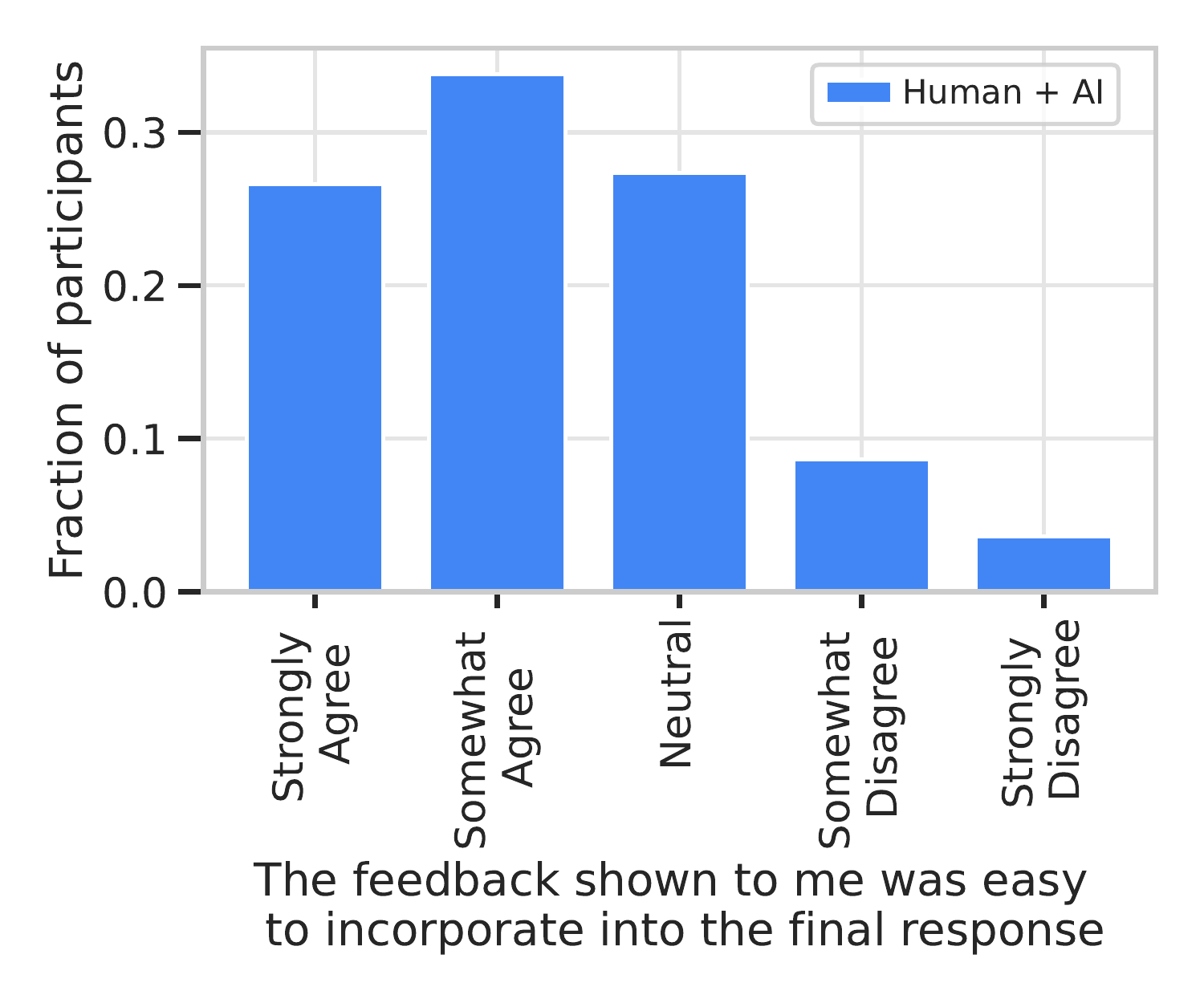} }
\hfill
\subfloat[Self-efficacy]{
	\includegraphics[width=0.45\columnwidth]{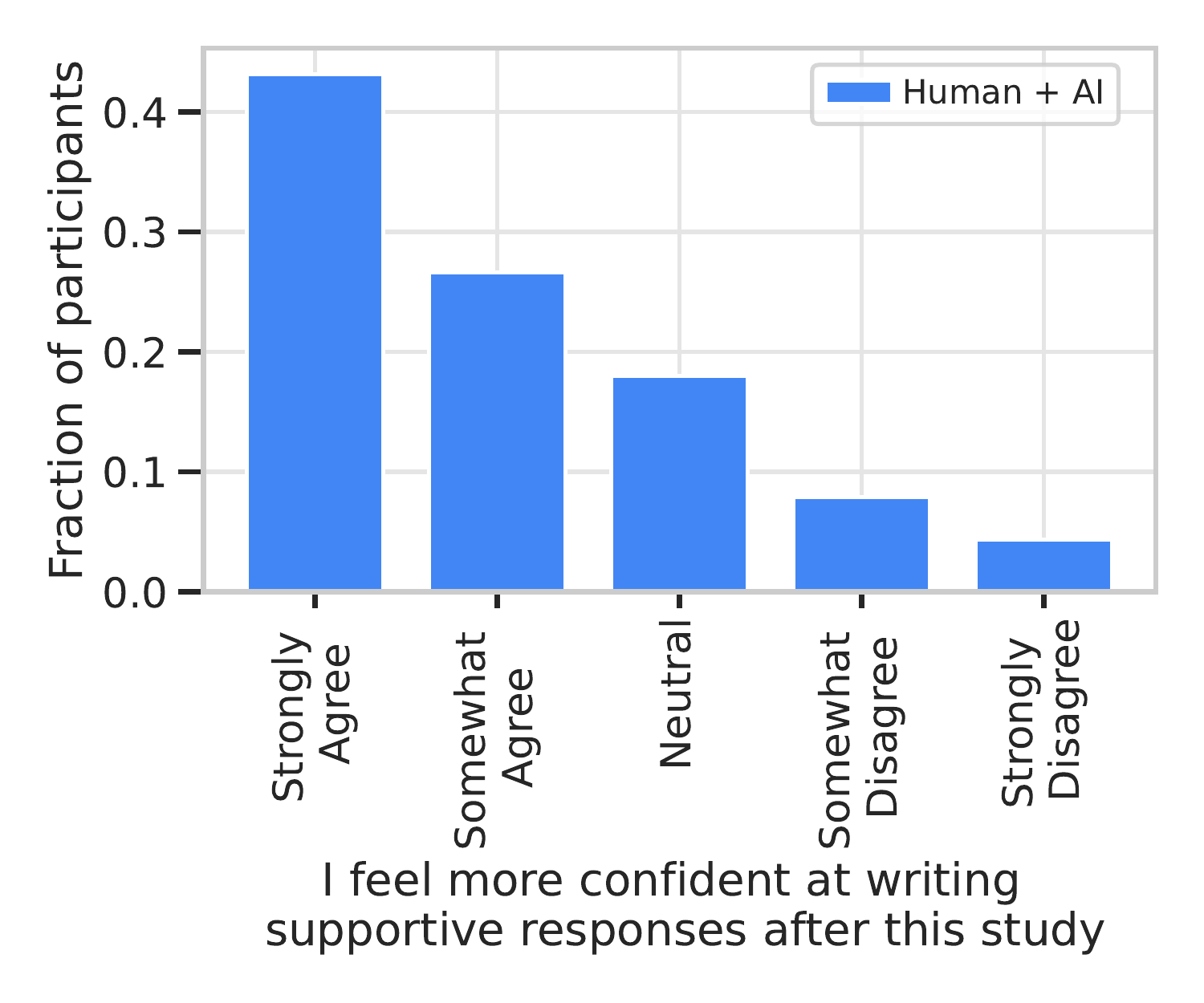} }
\hfill
\subfloat[Intention to adopt]{
	\includegraphics[width=0.45\columnwidth]{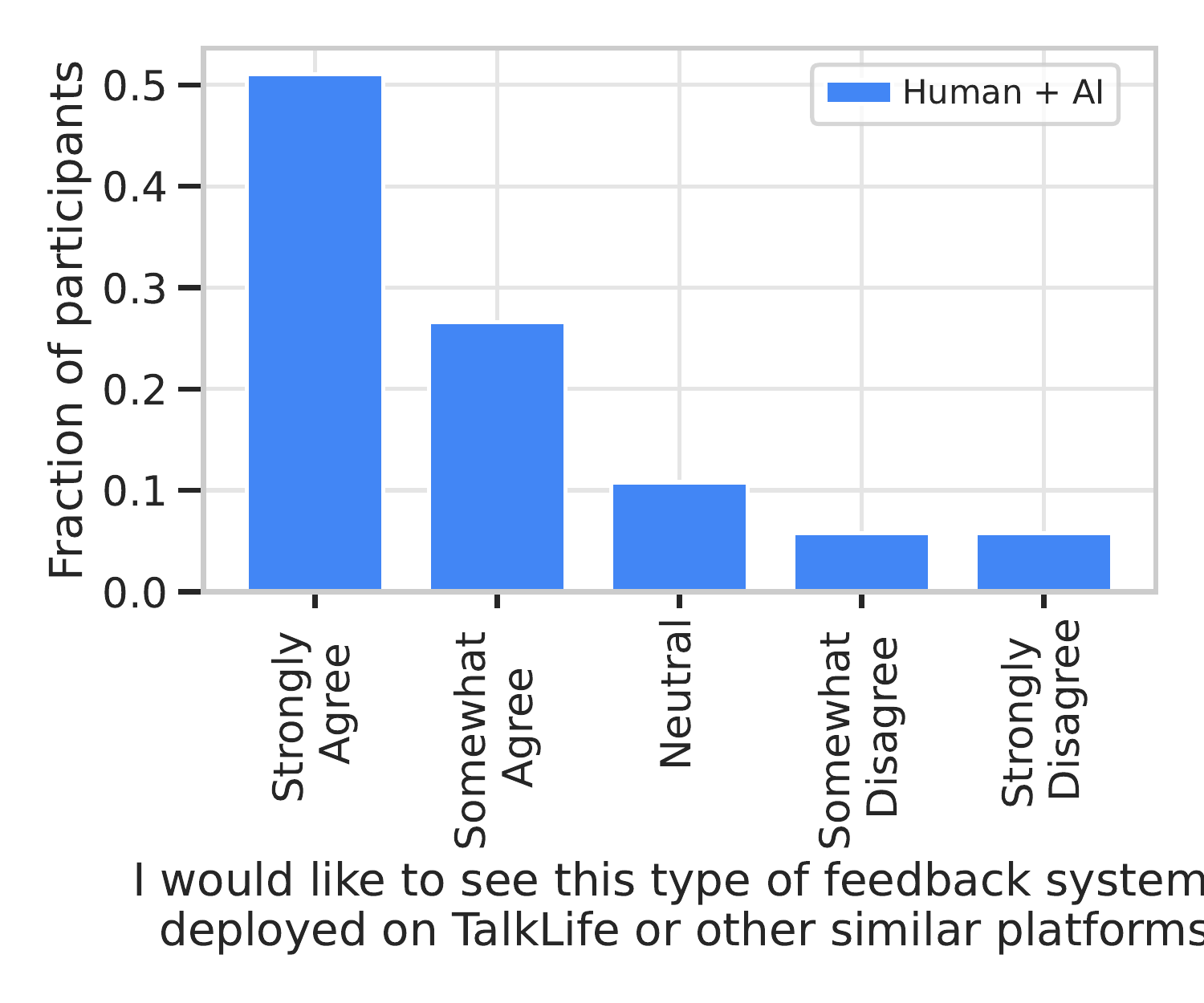} }
\label{supp:figure:exit_survey_treatment}
\end{figure} 

%% file: suppFigures/_supp_figure_exit_survey_vs_empathy.tex
\begin{figure}
\centering
\caption{Expressed empathy levels of responses with perceptions of Human + AI (treatment) group participants, as reported in phase IV (post-intervention survey). The area of the points is proportional to the number of participants with respective perceptions. Error bars indicate bootstrapped 95\% confidence intervals.}
\subfloat[Overall helpfulness]{
	\includegraphics[width=0.4\columnwidth]{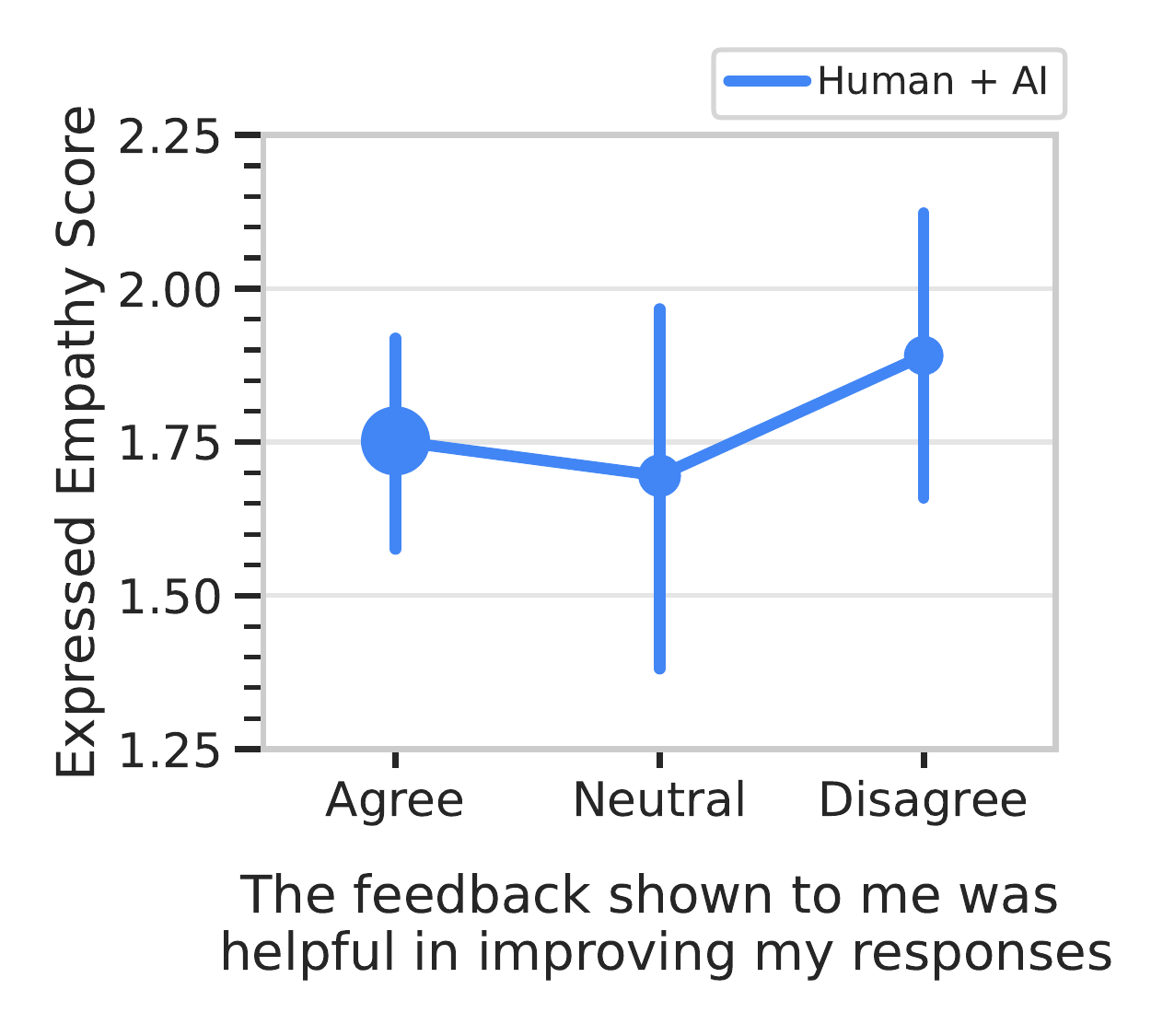} } 
\hfill
\subfloat[Helpfulness for empathy]{
	\includegraphics[width=0.4\columnwidth]{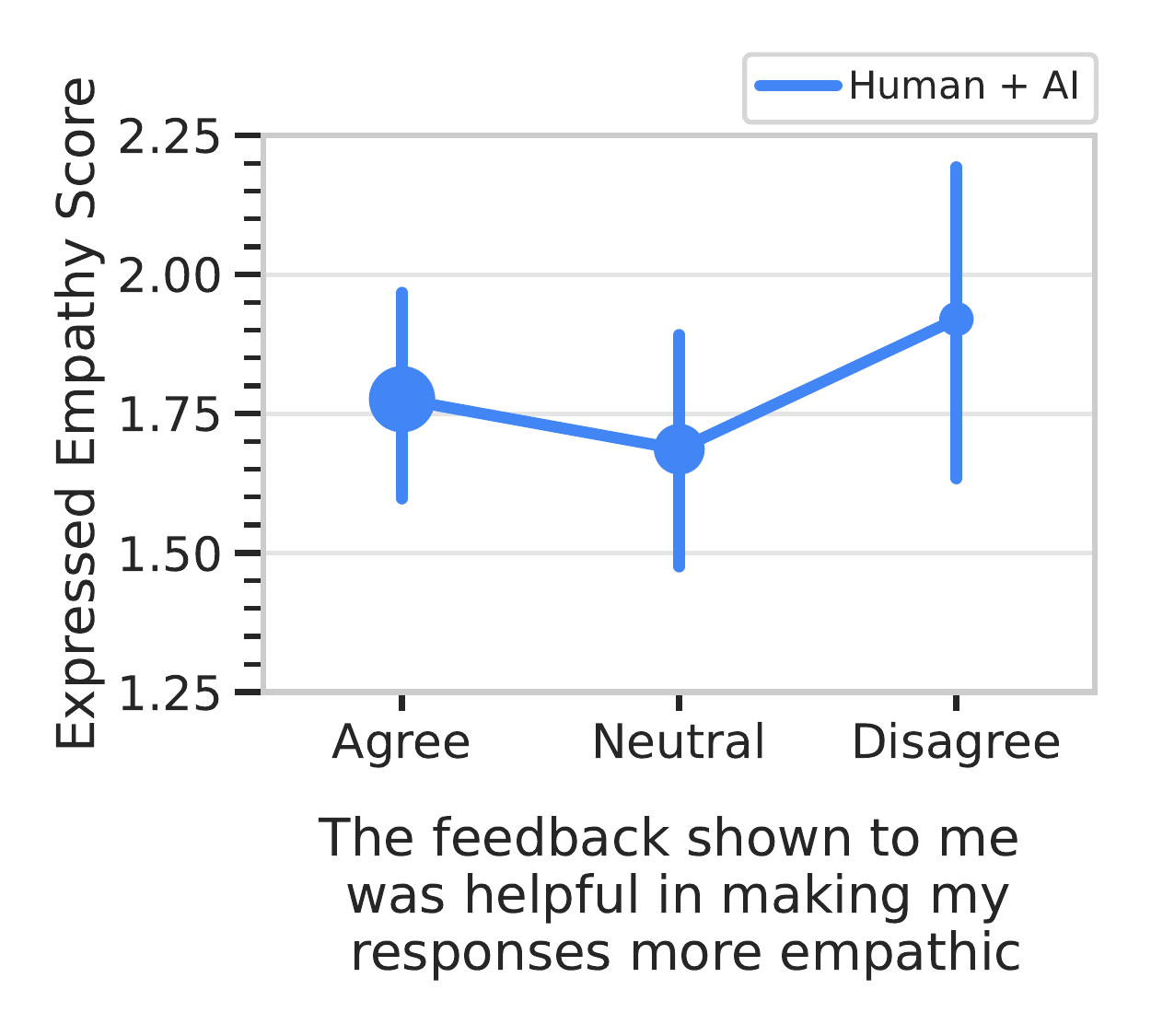} } 
\hfill
\subfloat[Actionability]{
	\includegraphics[width=0.4\columnwidth]{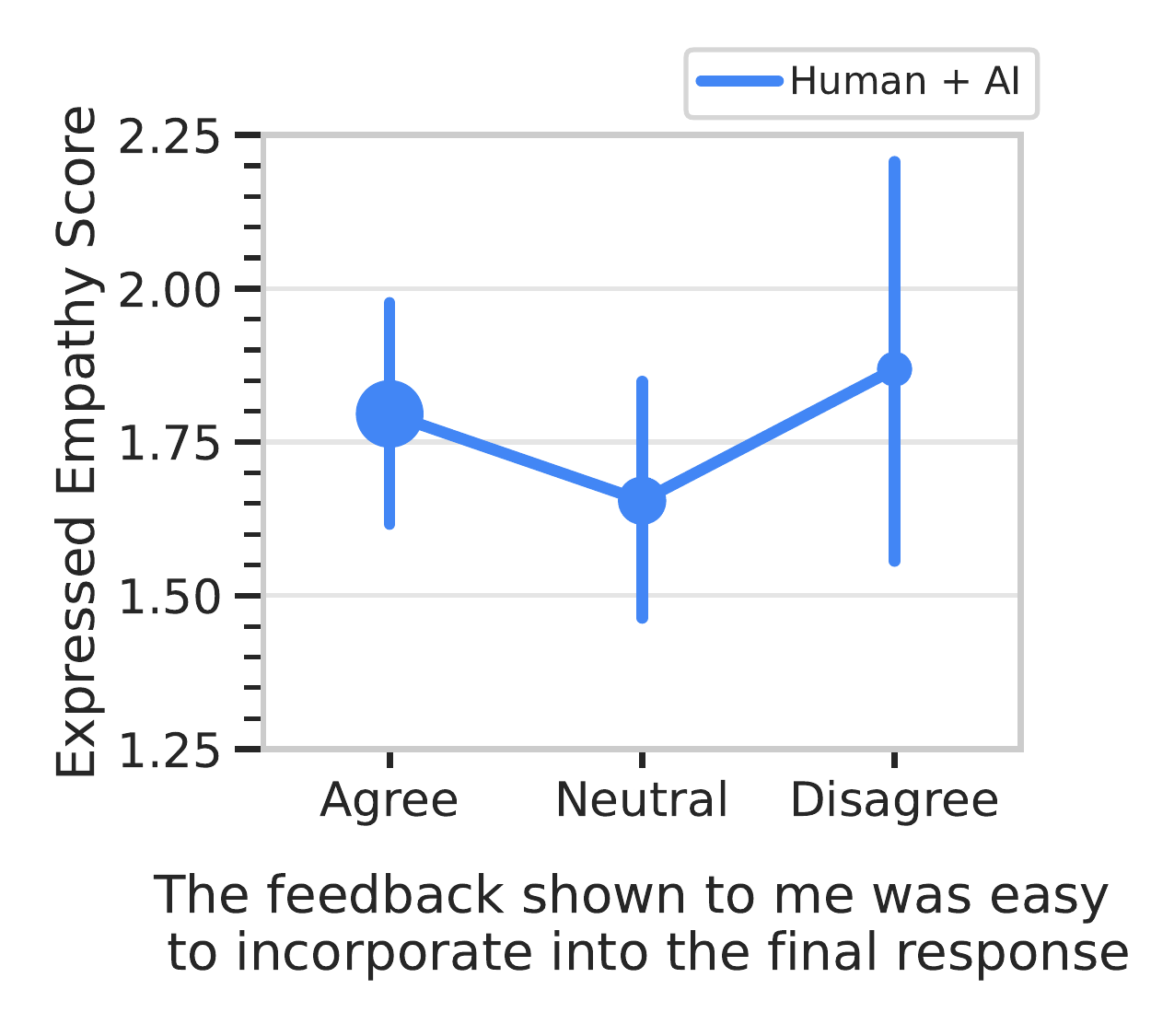} }
\hfill
\subfloat[Self-efficacy]{
	\includegraphics[width=0.4\columnwidth]{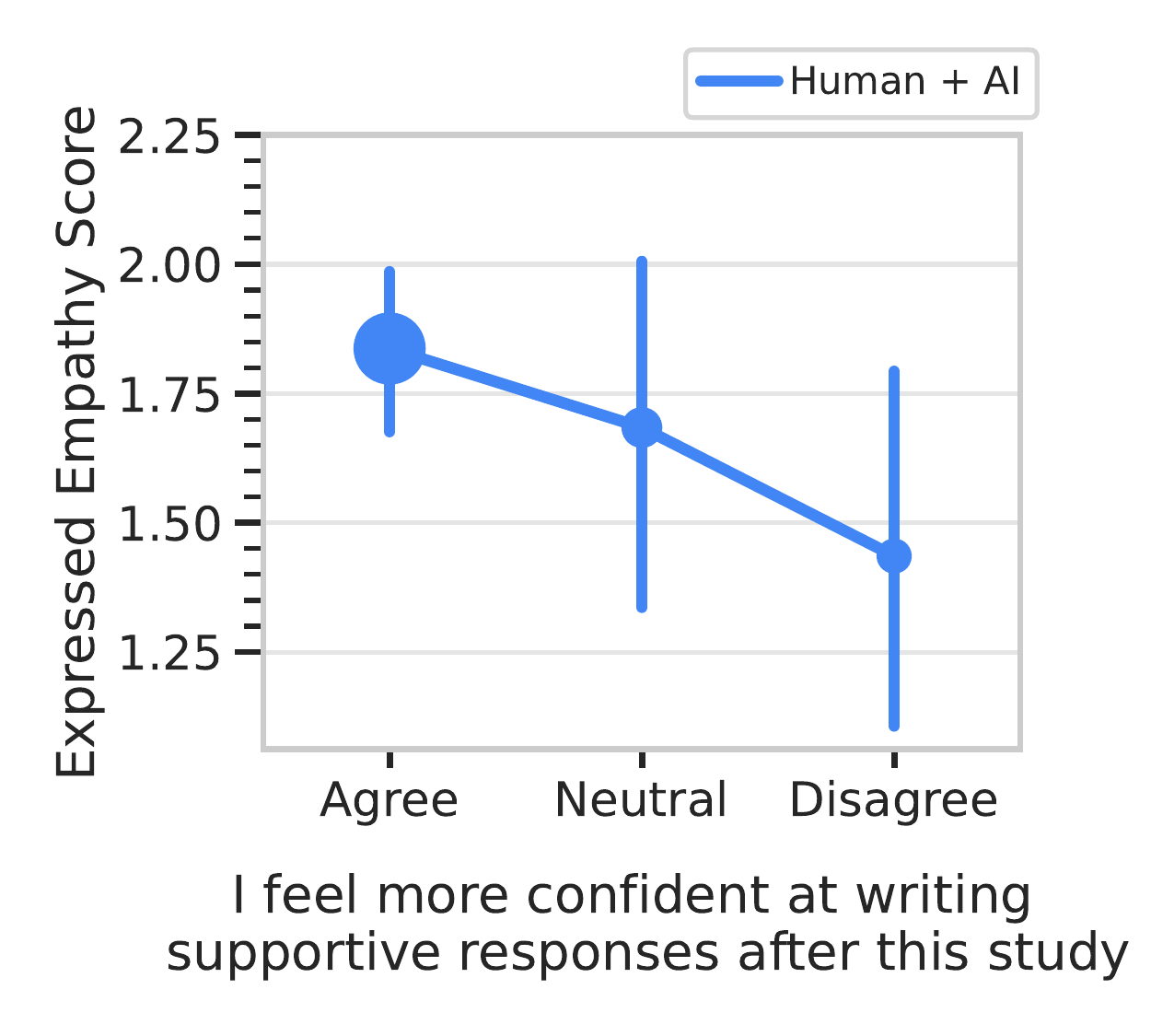} }
\hfill
\subfloat[Intention to adopt]{
	\includegraphics[width=0.4\columnwidth]{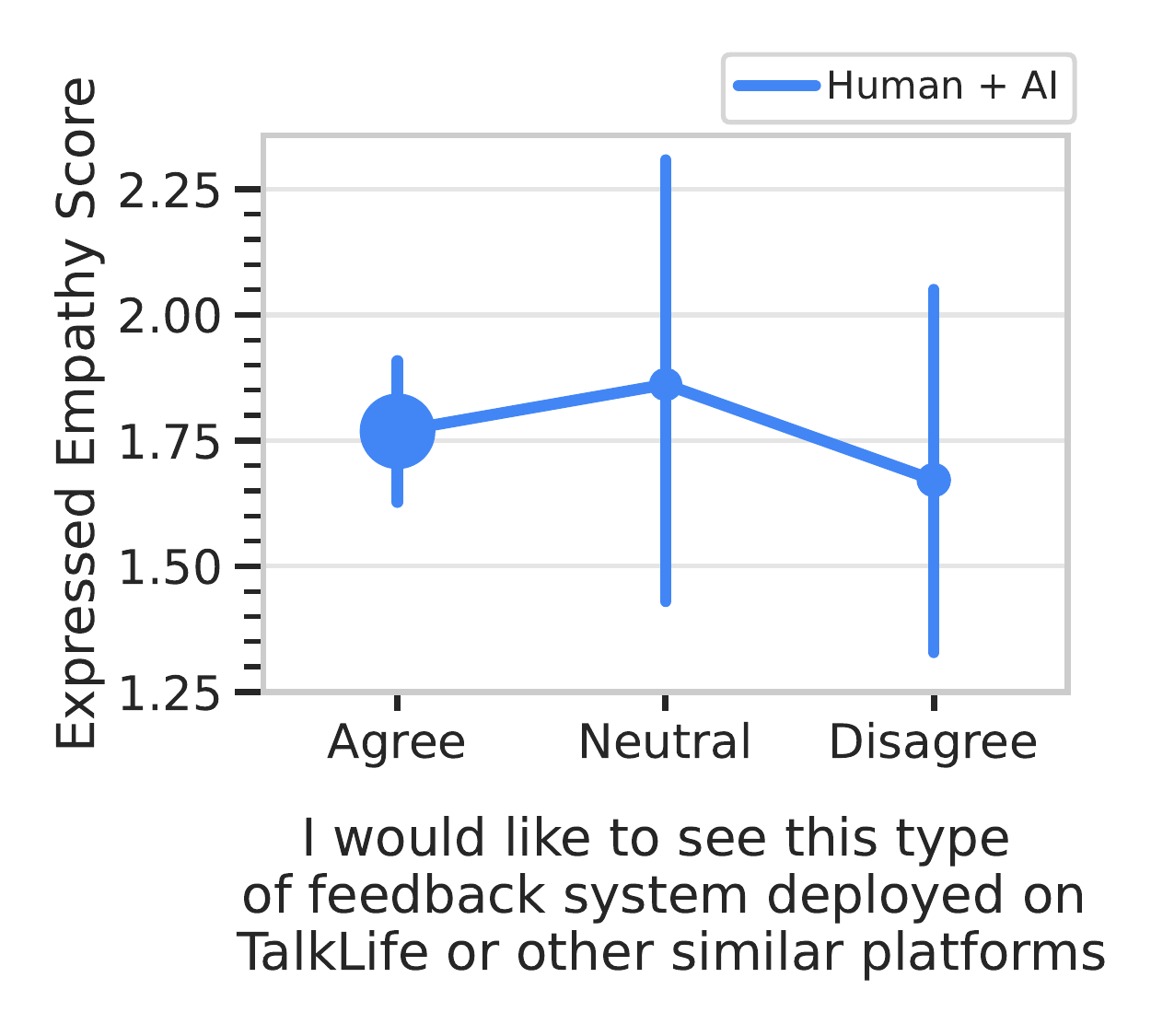} }
\label{supp:figure:exit_survey_vs_empathy}
\end{figure} 

%% file: suppFigures/_supp_figure_hai_perceptions.tex
\begin{figure}
\centering
\caption{Participant perceptions, as reported in phase IV (post-intervention survey), with different human-AI collaboration categories.}
\subfloat[Challenges: Feedback and Suggestions on how to improve my responses would have helped me write better responses]{
	\includegraphics[width=0.47\columnwidth]{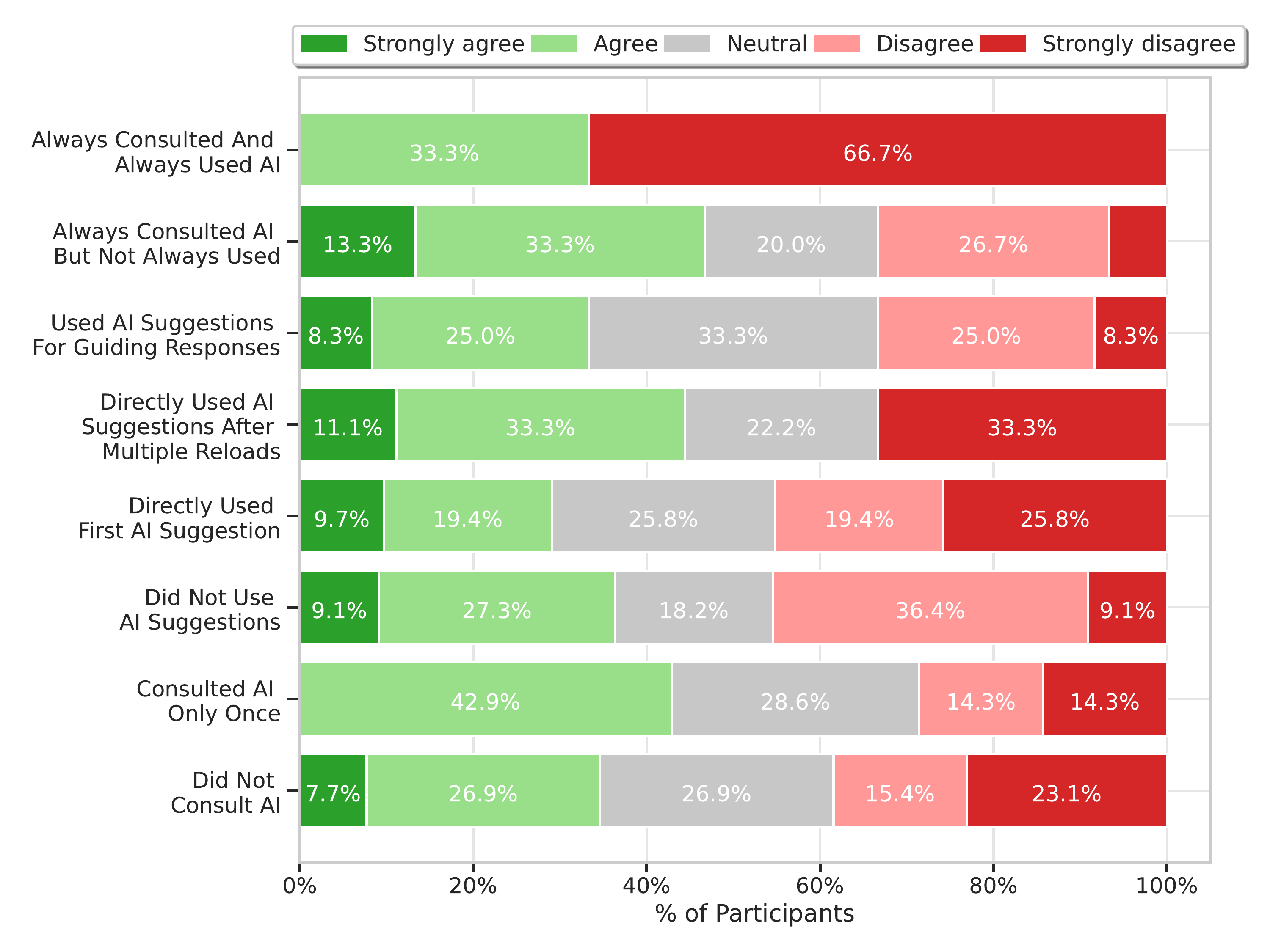} } 
\hfill
\subfloat[Overall helpfulness: The feedback shown to me was helpful in improving my responses]{
	\includegraphics[width=0.47\columnwidth]{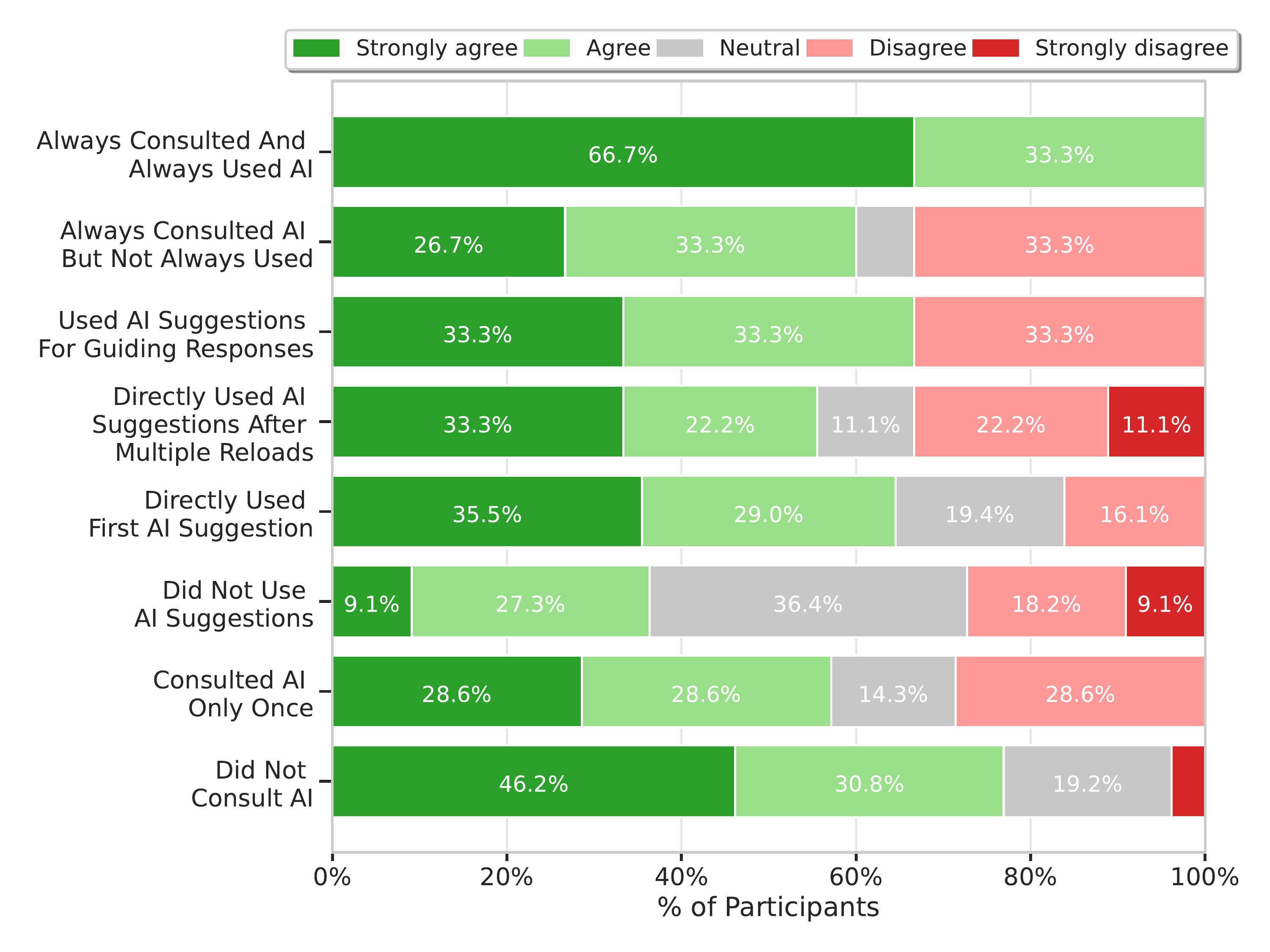} } 
\hfill
\subfloat[Helpfulness for empathy: The feedback shown to me was helpful in making my responses more empathic]{
	\includegraphics[width=0.47\columnwidth]{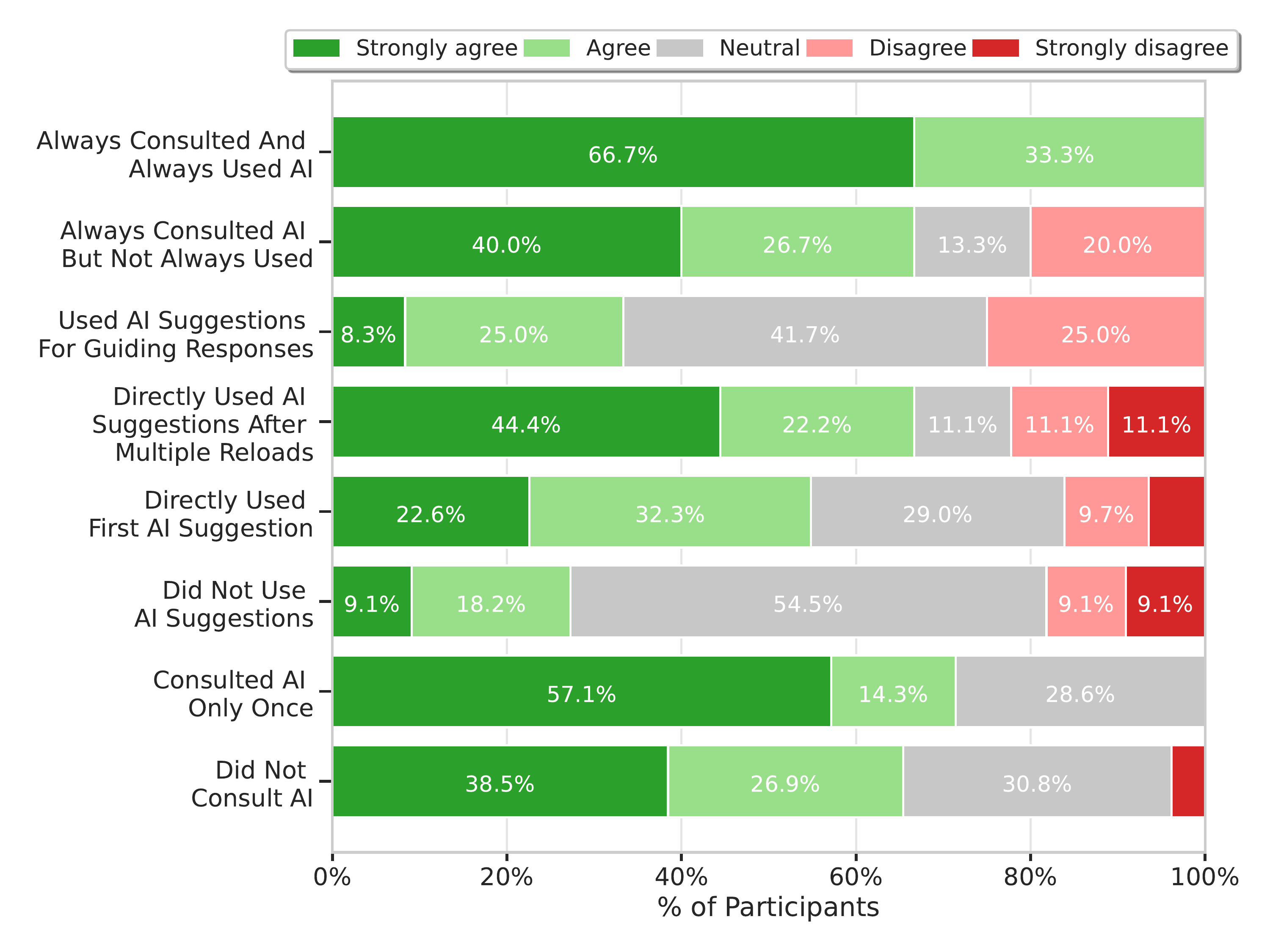} } 
\hfill
\subfloat[Actionability: The feedback shown to me was easy to incorporate into the final response]{
	\includegraphics[width=0.47\columnwidth]{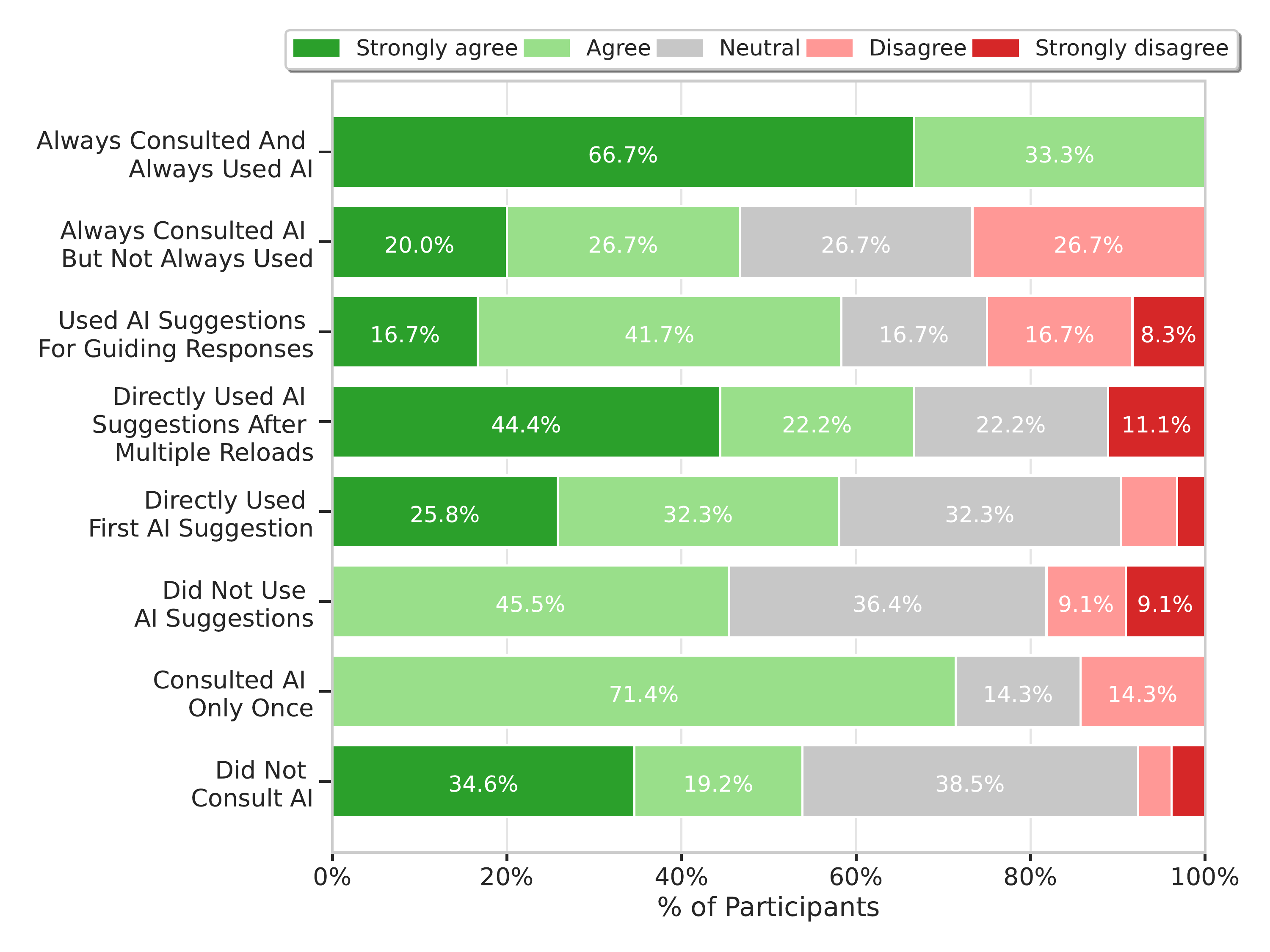} }
\hfill
\subfloat[Self-efficacy: I feel more confident at writing supportive responses after this study]{
	\includegraphics[width=0.47\columnwidth]{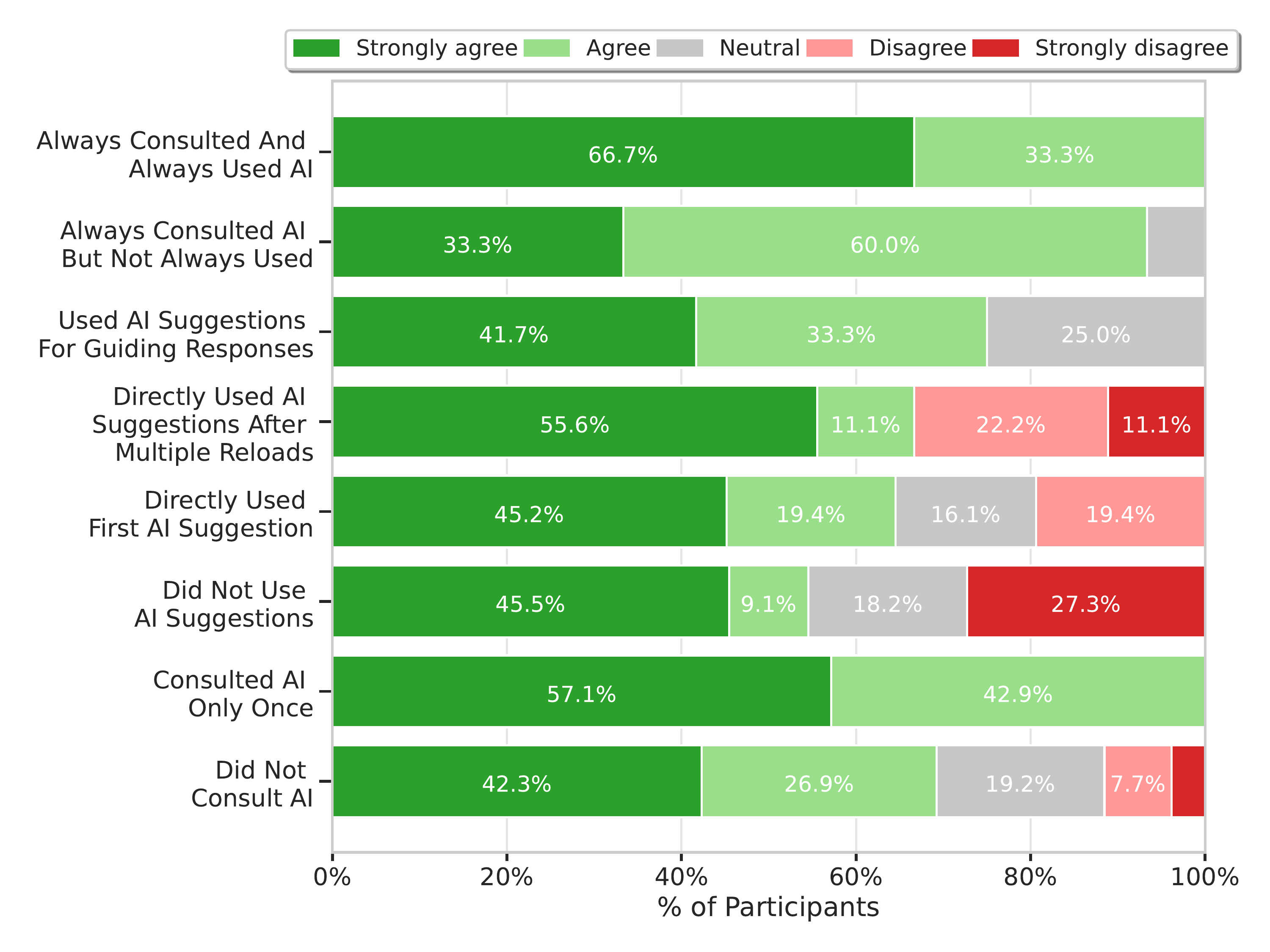} }
\hfill
\subfloat[Intention to adopt: I  would like to see this type of feedback system deployed on TalkLife or other similar platforms]{
	\includegraphics[width=0.47\columnwidth]{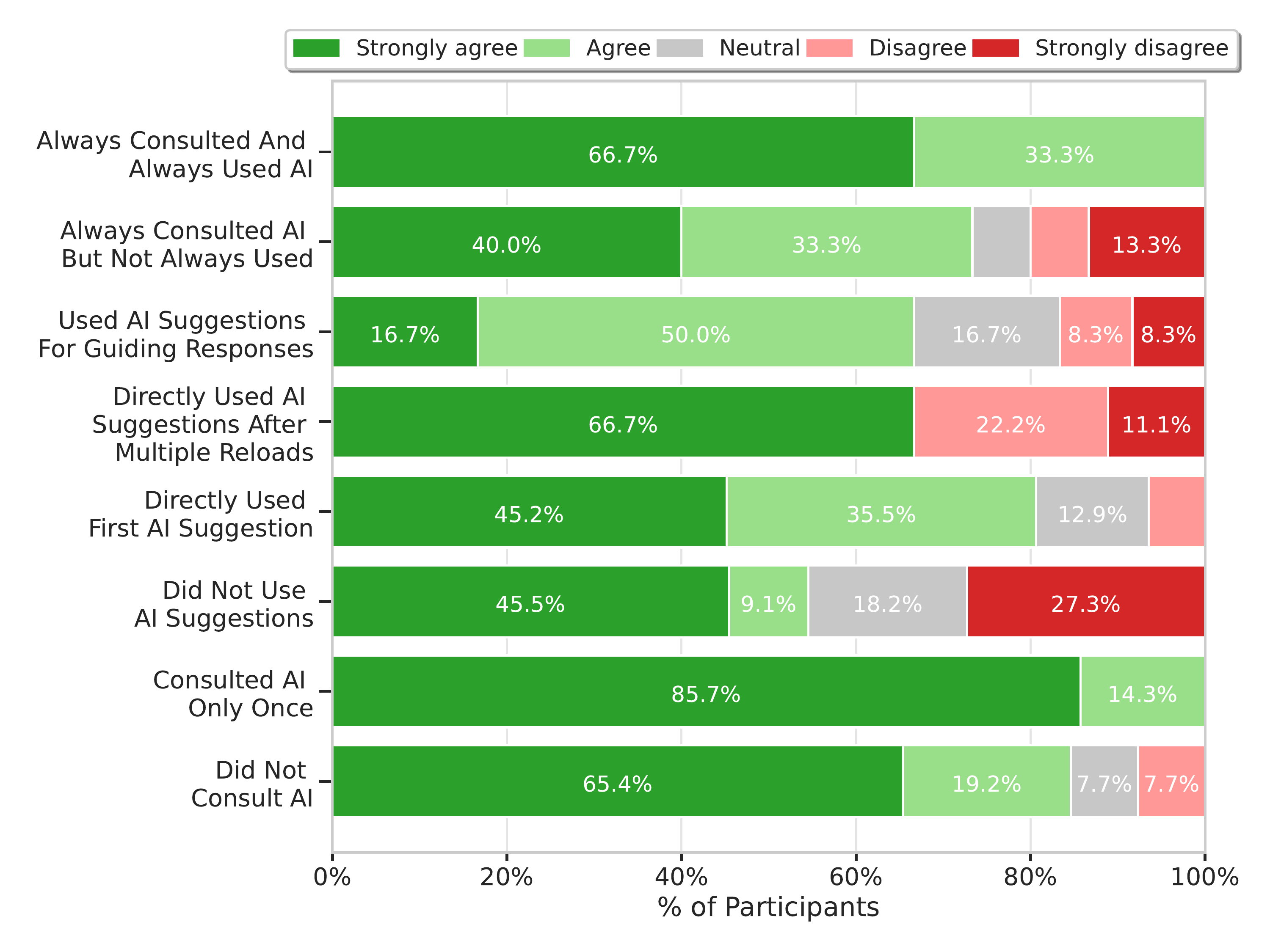} }
\label{supp:figure:hai_perceptions}
\end{figure} 

%% file: suppFigures/_supp_figure_advertisement.tex
\begin{figure}
    \caption{Pop-up used to advertise our study on TalkLife. This pop-up is shown to TalkLife users after they submit a response on the TalkLife platform, with an aim of targeting active peer supporters.}
    \centering
         \includegraphics[width=0.6\textwidth]{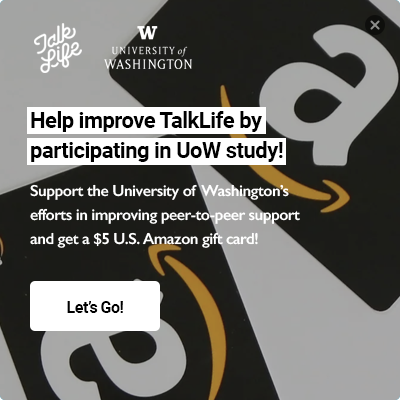}
    \label{supp:figure:advertisement}
\end{figure}

%% file: suppFigures/_supp_figure_study_design.tex
\begin{figure}
    \caption{Consent form used in our study.}
    \centering
         \includegraphics[width=\textwidth]{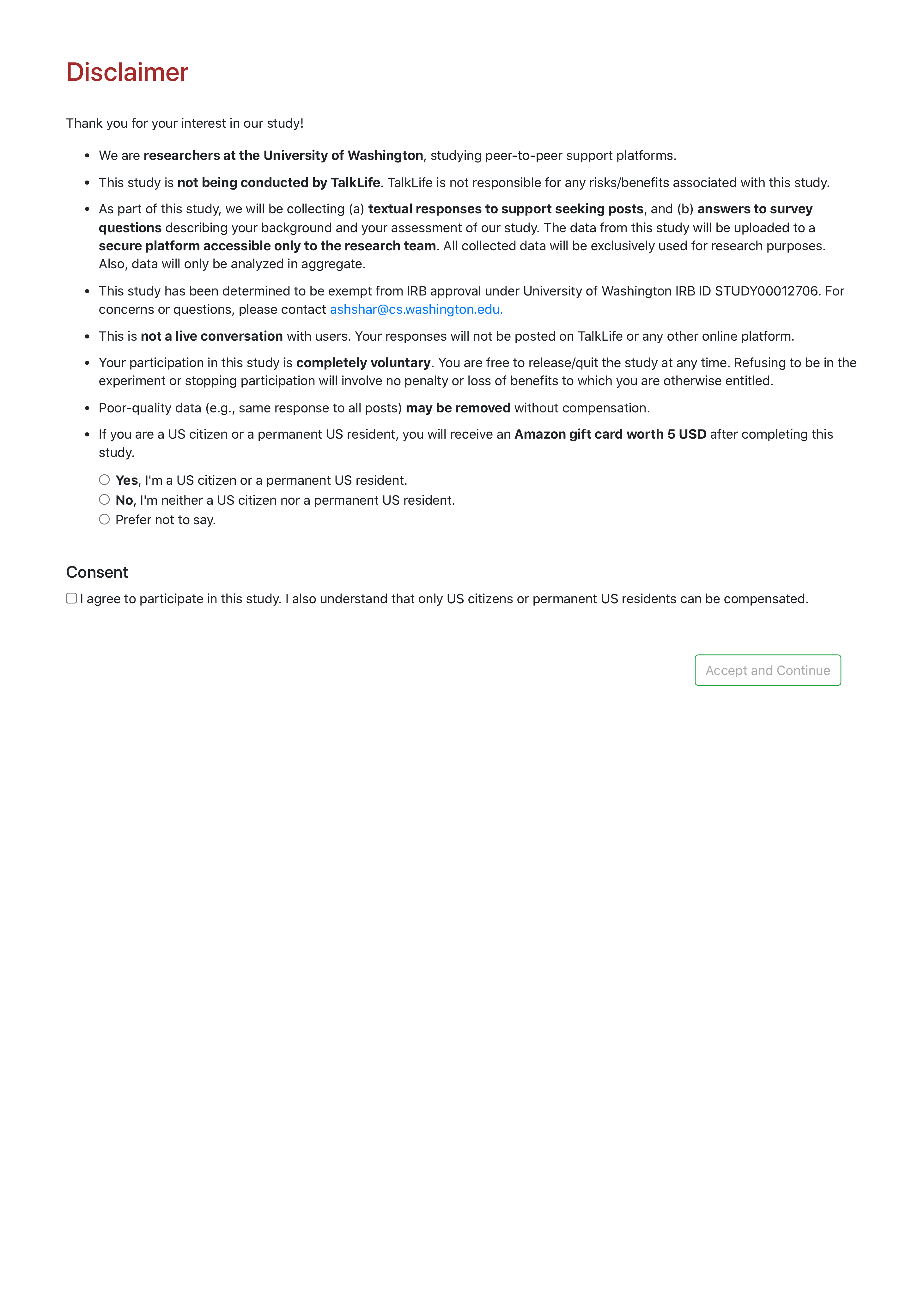}
    \label{supp:figure:study_design:consent}
\end{figure}

\begin{figure}
    \caption{Form used for collecting demographics and background of participants [phase I: pre-intervention survey].}
    \centering
         \includegraphics[width=\textwidth]{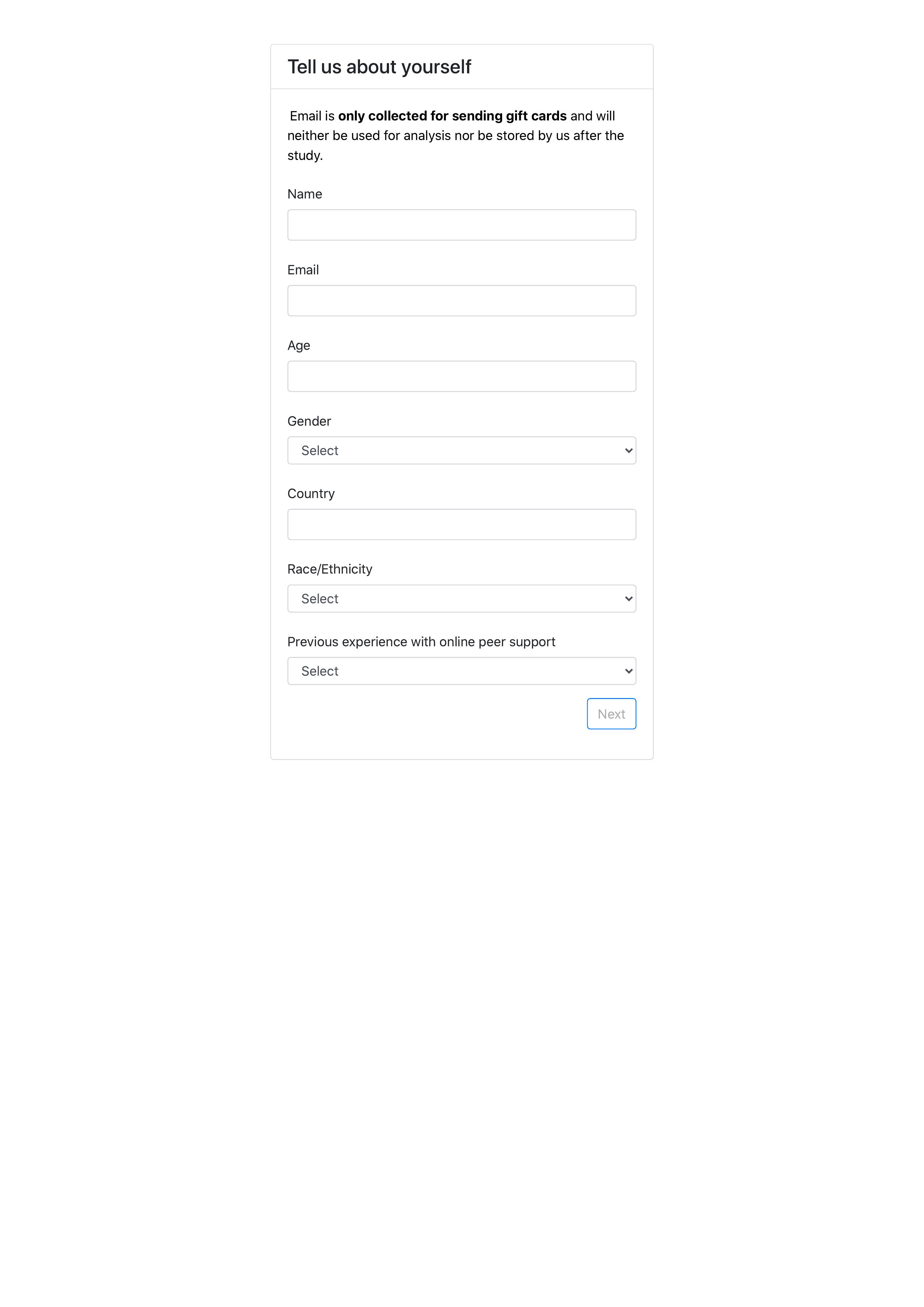}
    \label{supp:figure:study_design:background}
\end{figure}

\begin{figure}
    \caption{Onboarding survey used for collecting perceptions of participants [phase I: pre-intervention survey].}
    \centering
         \includegraphics[width=\textwidth]{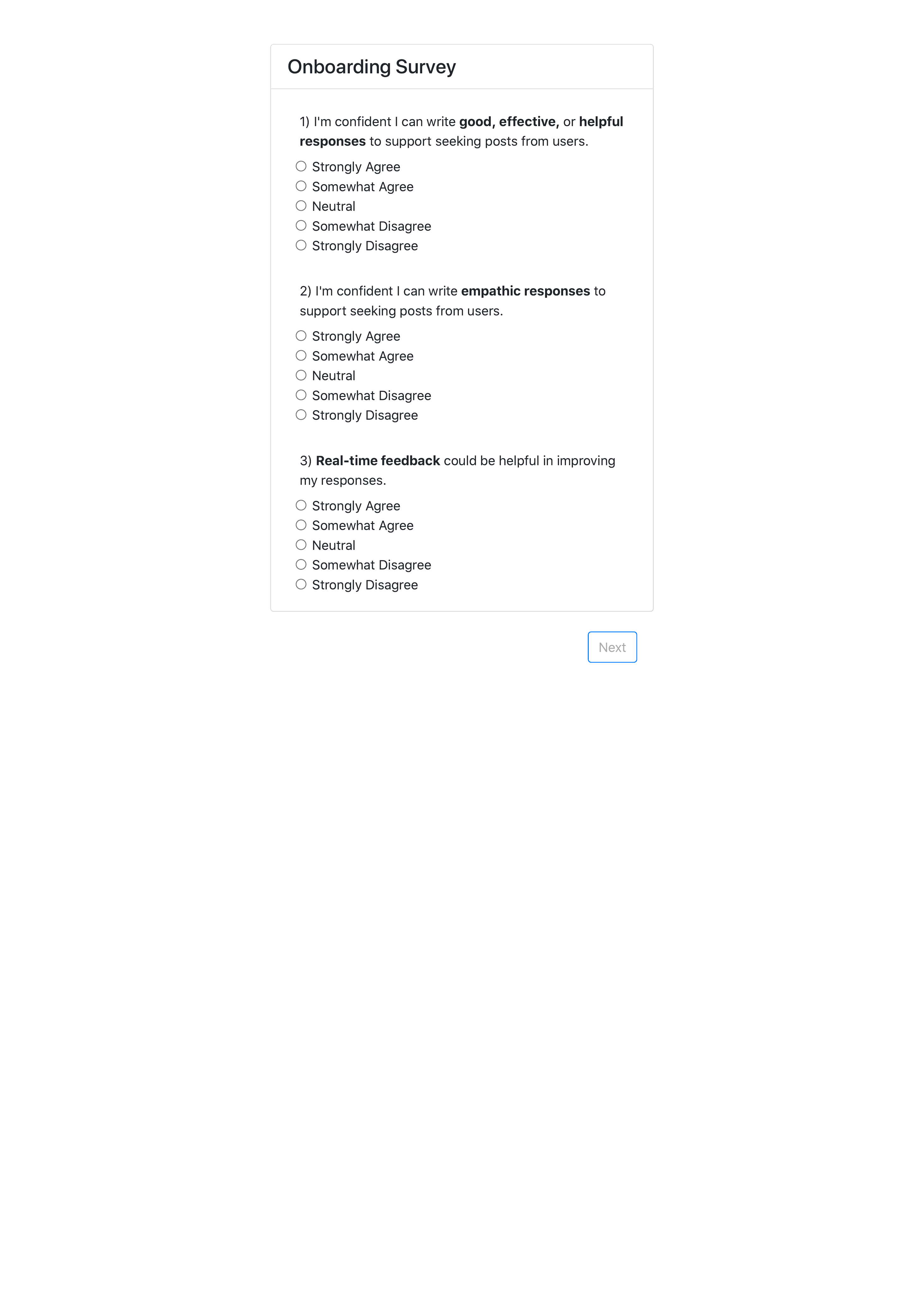}
    \label{supp:figure:study_design:onboarding}
\end{figure}

\begin{figure}
    \caption{Instructions shown to the control group participants [phase II: empathy training and instructions]. Continued on the next page (1/2).}
    \centering
         \includegraphics[width=\textwidth]{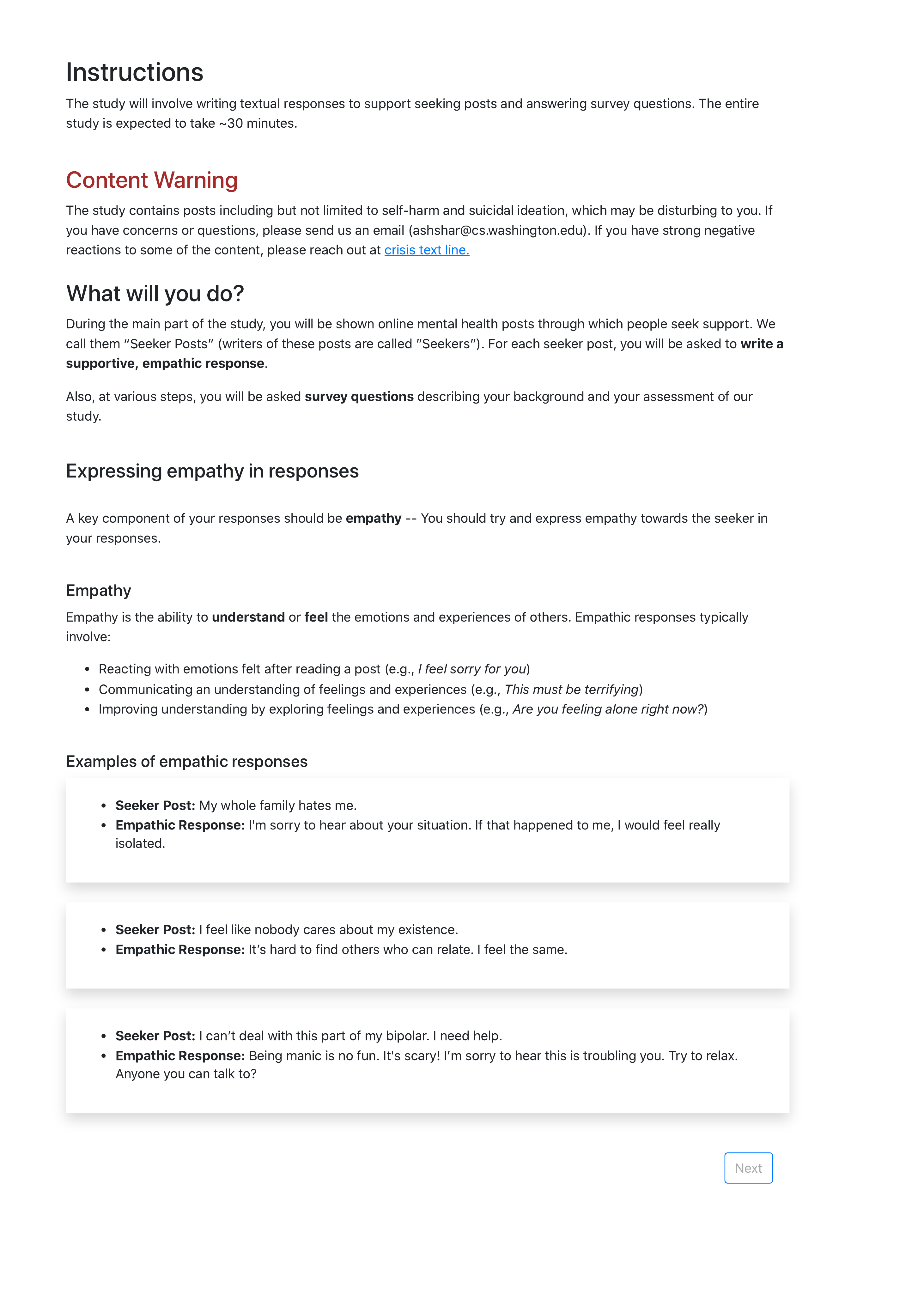}
    \label{supp:figure:study_design:instructions_control_1}
\end{figure}

\begin{figure}
    \caption{Instructions shown to the control group participants [phase II: empathy training and instructions] (2/2).}
    \centering
         \includegraphics[width=\textwidth]{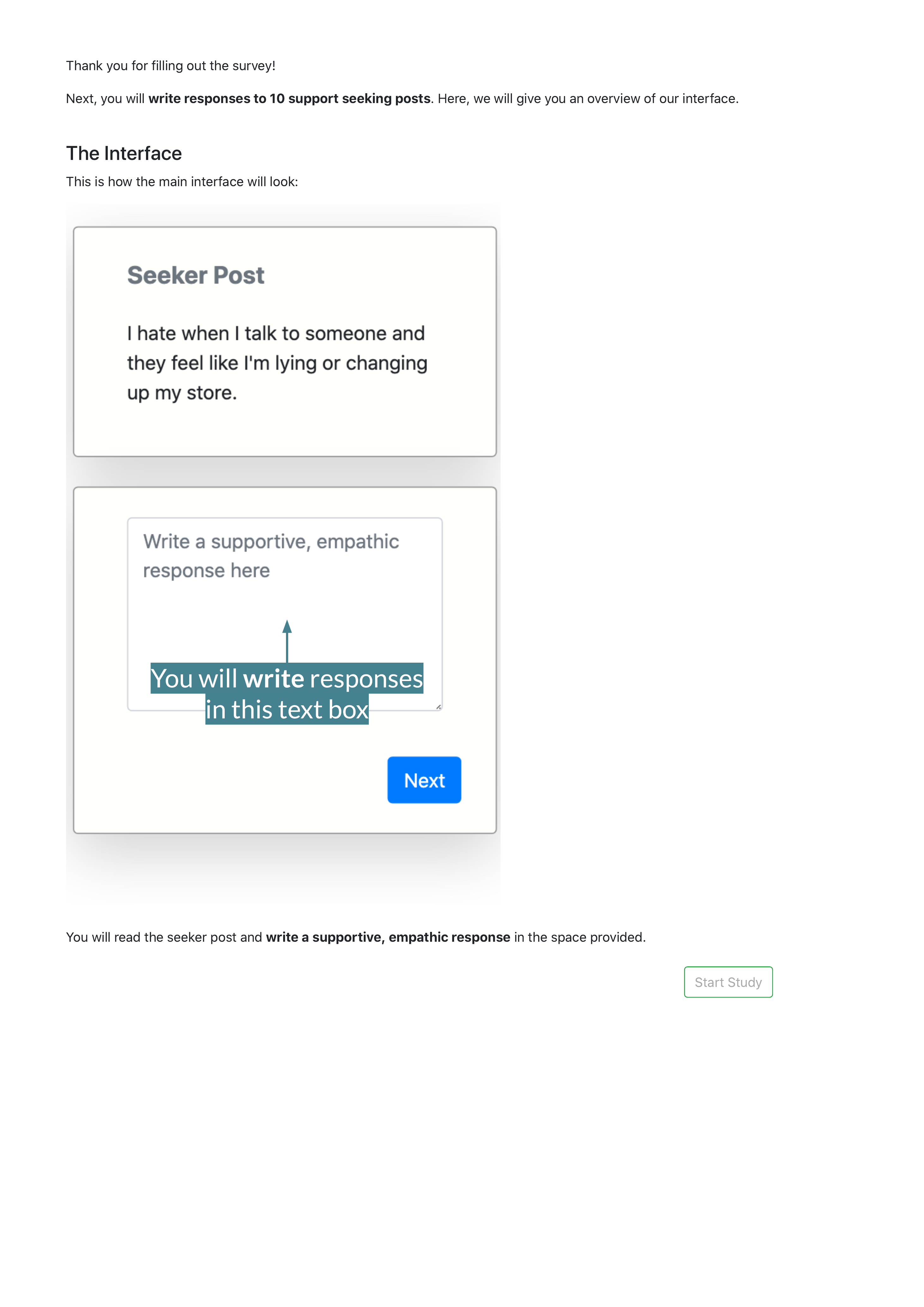}
    \label{supp:figure:study_design:instructions_control_2}
\end{figure}

\begin{figure}
\vspace{-10pt}
    \caption{Instructions shown to the treatment group participants [phase II: empathy training and instructions]. Continued on the next page (1/6).}
    \centering
         \includegraphics[width=\textwidth]{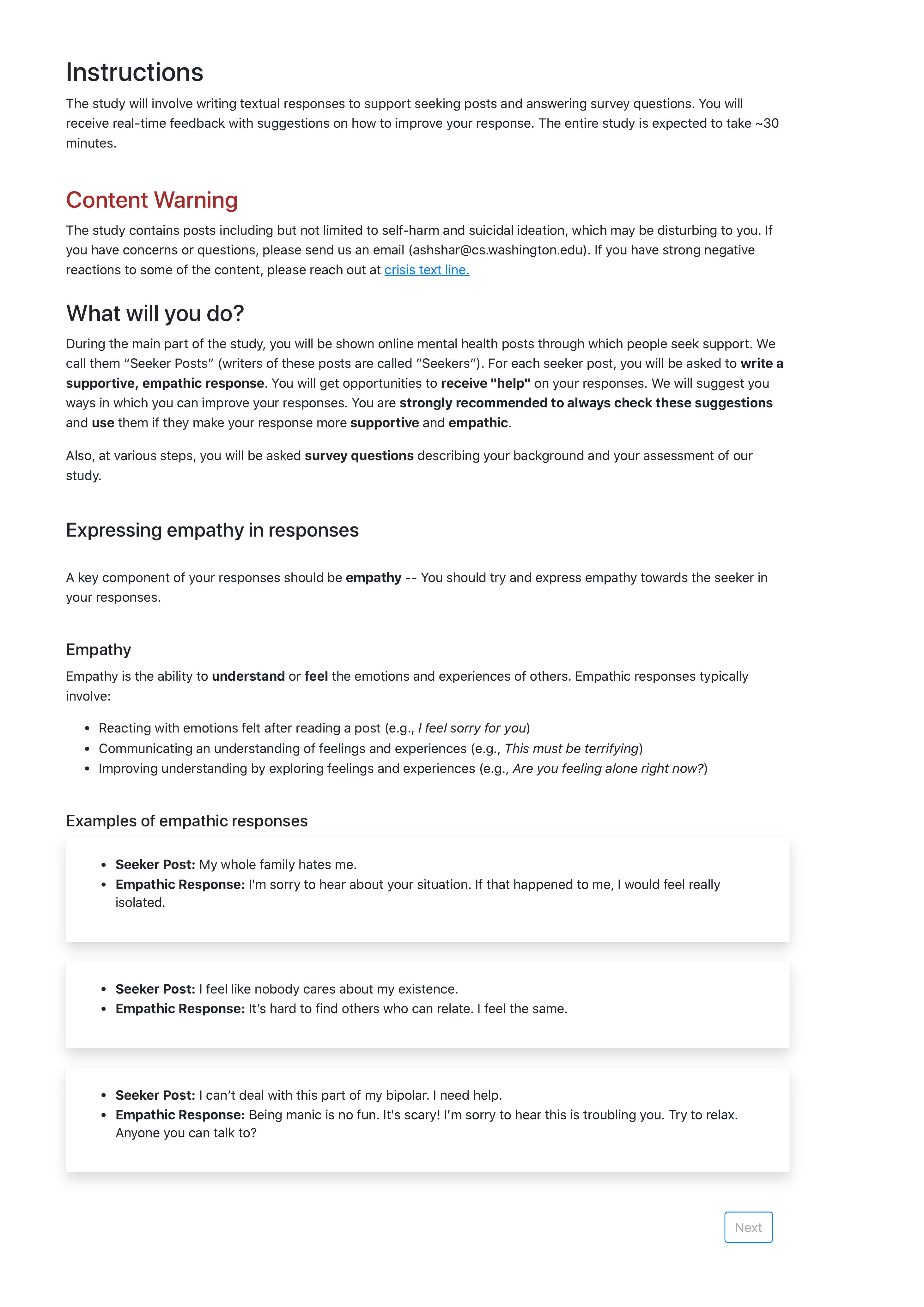}
    \label{supp:figure:study_design:instructions_treatment_1}
\end{figure}

\begin{figure}
    \caption{Instructions shown to the treatment group participants [phase II: empathy training and instructions]. Continued on the next page (2/6).}
    \centering
         \includegraphics[width=\textwidth]{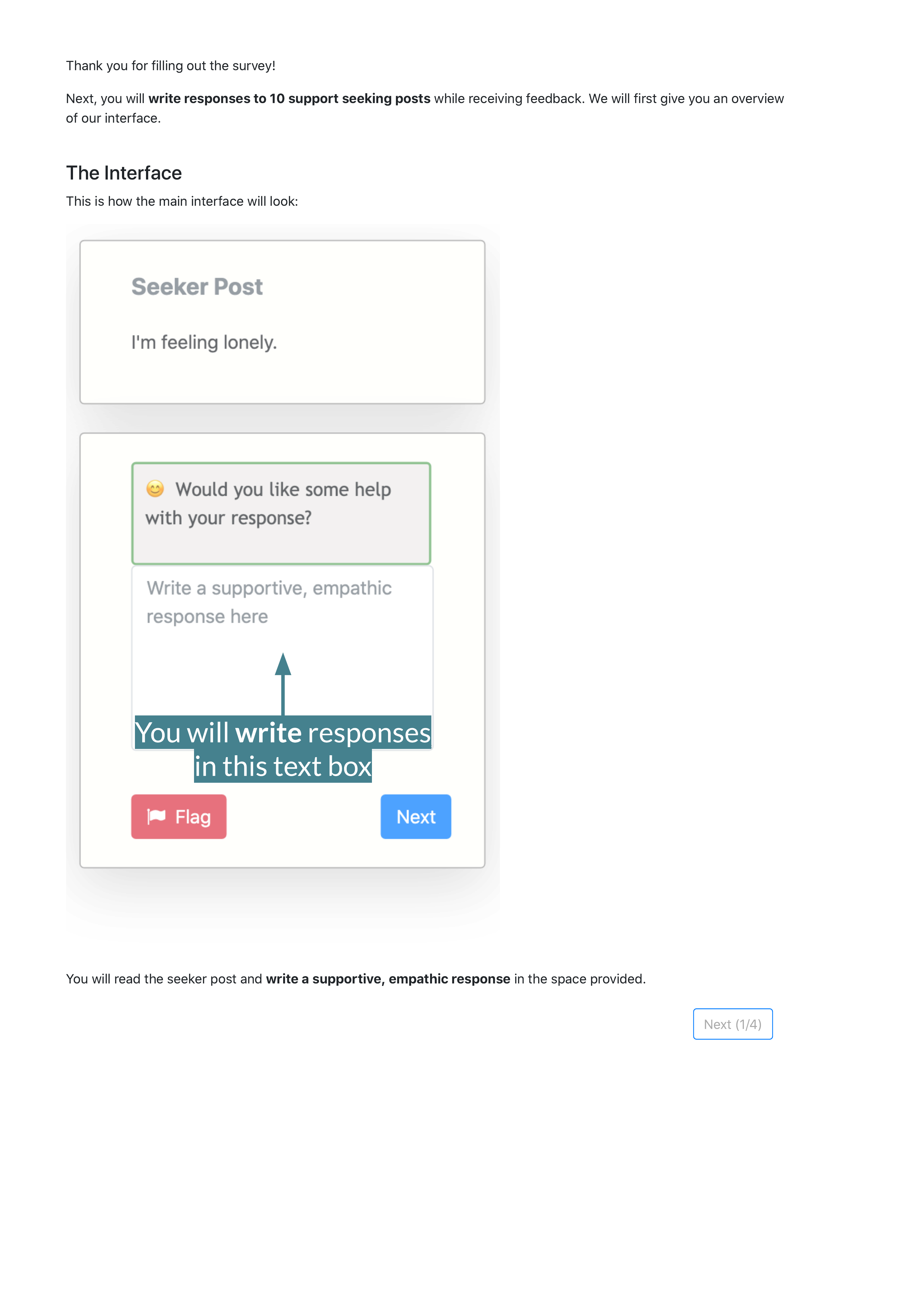}
    \label{supp:figure:study_design:instructions_treatment_2}
\end{figure}

\begin{figure}
    \caption{Instructions shown to the treatment group participants [phase II: empathy training and instructions]. Continued on the next page (3/6).}
    \centering
         \includegraphics[width=\textwidth]{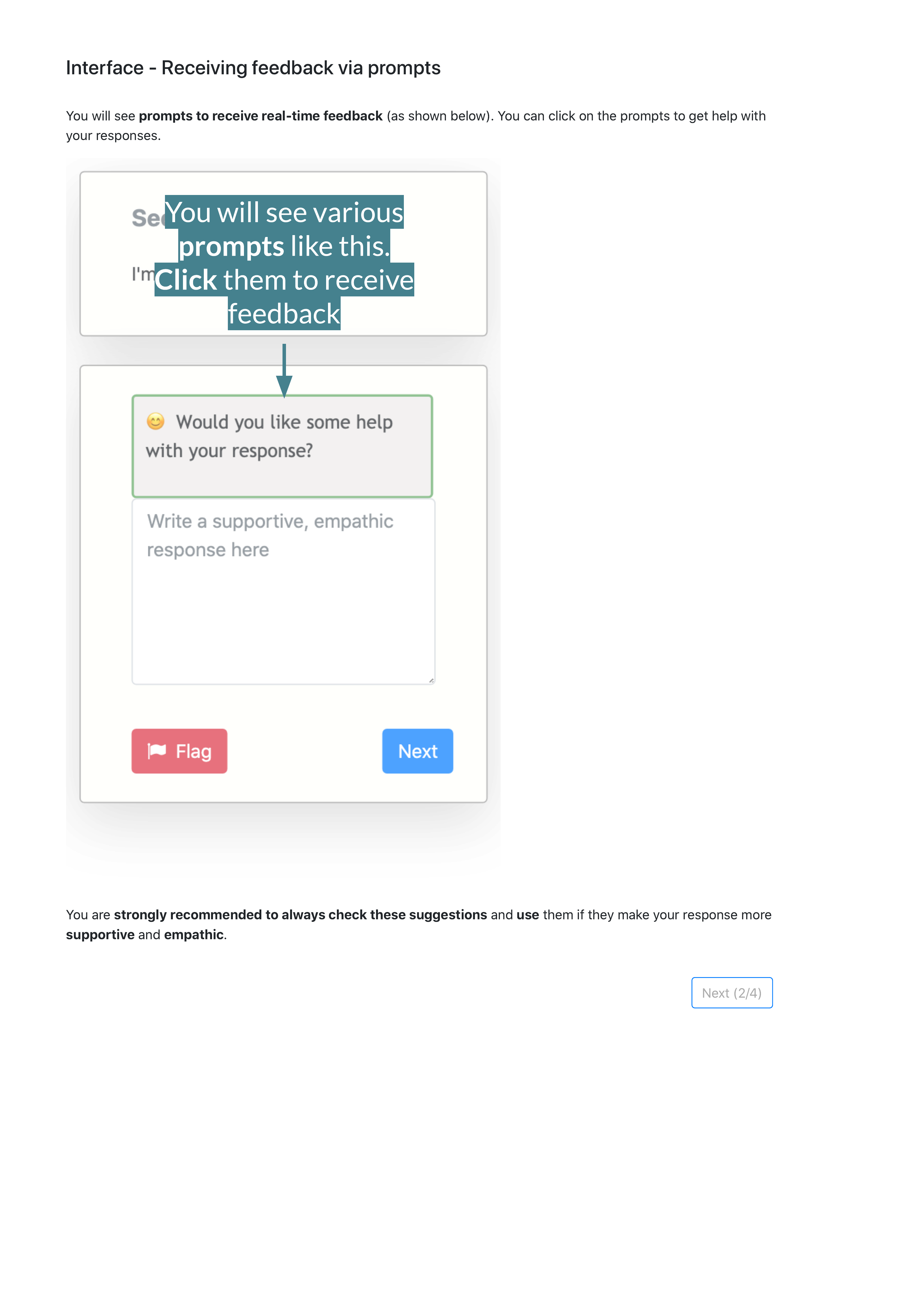}
    \label{supp:figure:study_design:instructions_treatment_3}
\end{figure}

\begin{figure}
    \caption{Instructions shown to the treatment group participants [phase II: empathy training and instructions]. Continued on the next page (4/6).}
    \centering
         \includegraphics[page=1,width=\textwidth]{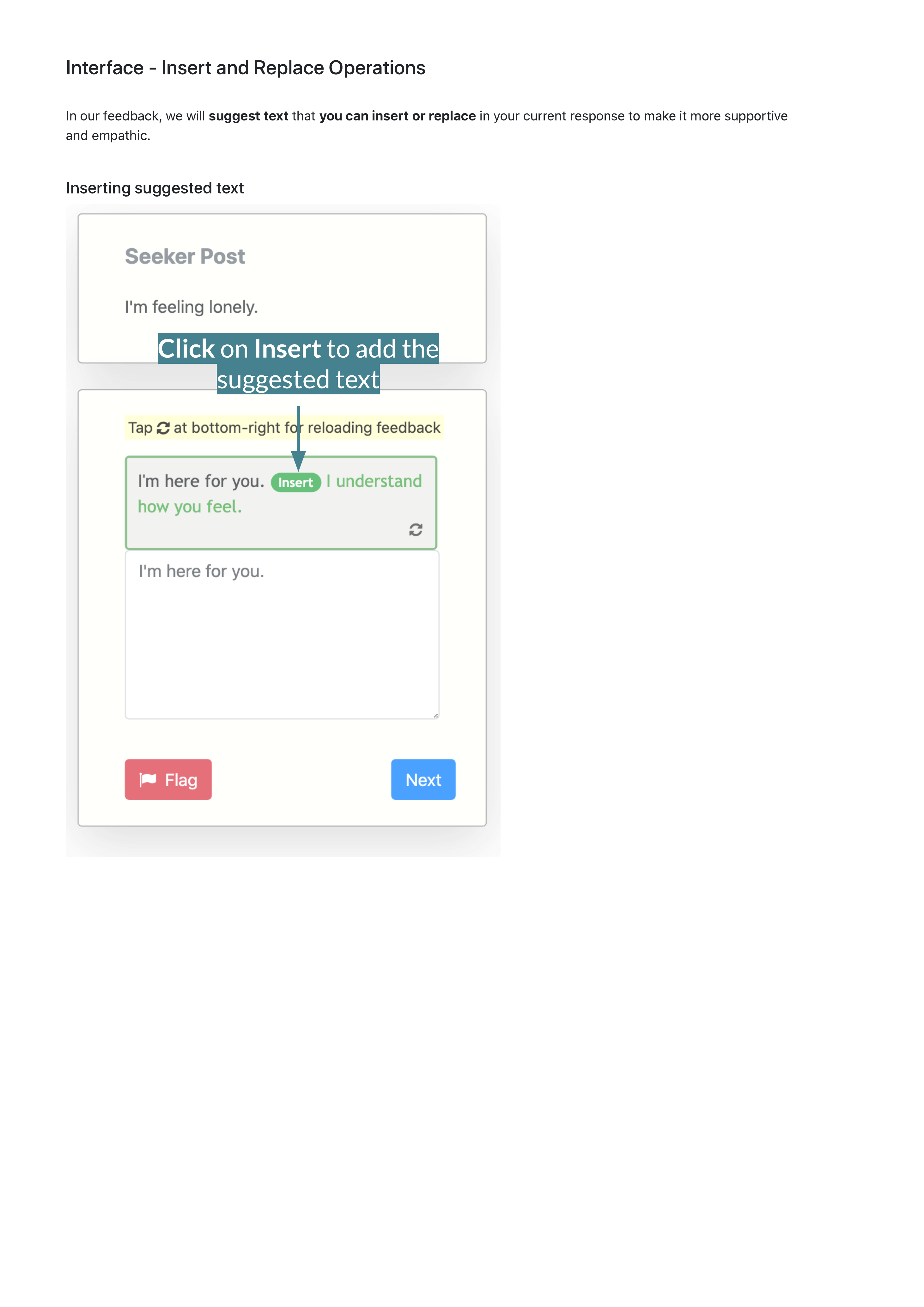}
    \label{supp:figure:study_design:instructions_treatment_4}
\end{figure}

\begin{figure}
    \caption{Instructions shown to the treatment group participants [phase II: empathy training and instructions]. Continued on the next page (5/6).}
    \centering
         \includegraphics[page=2,width=\textwidth]{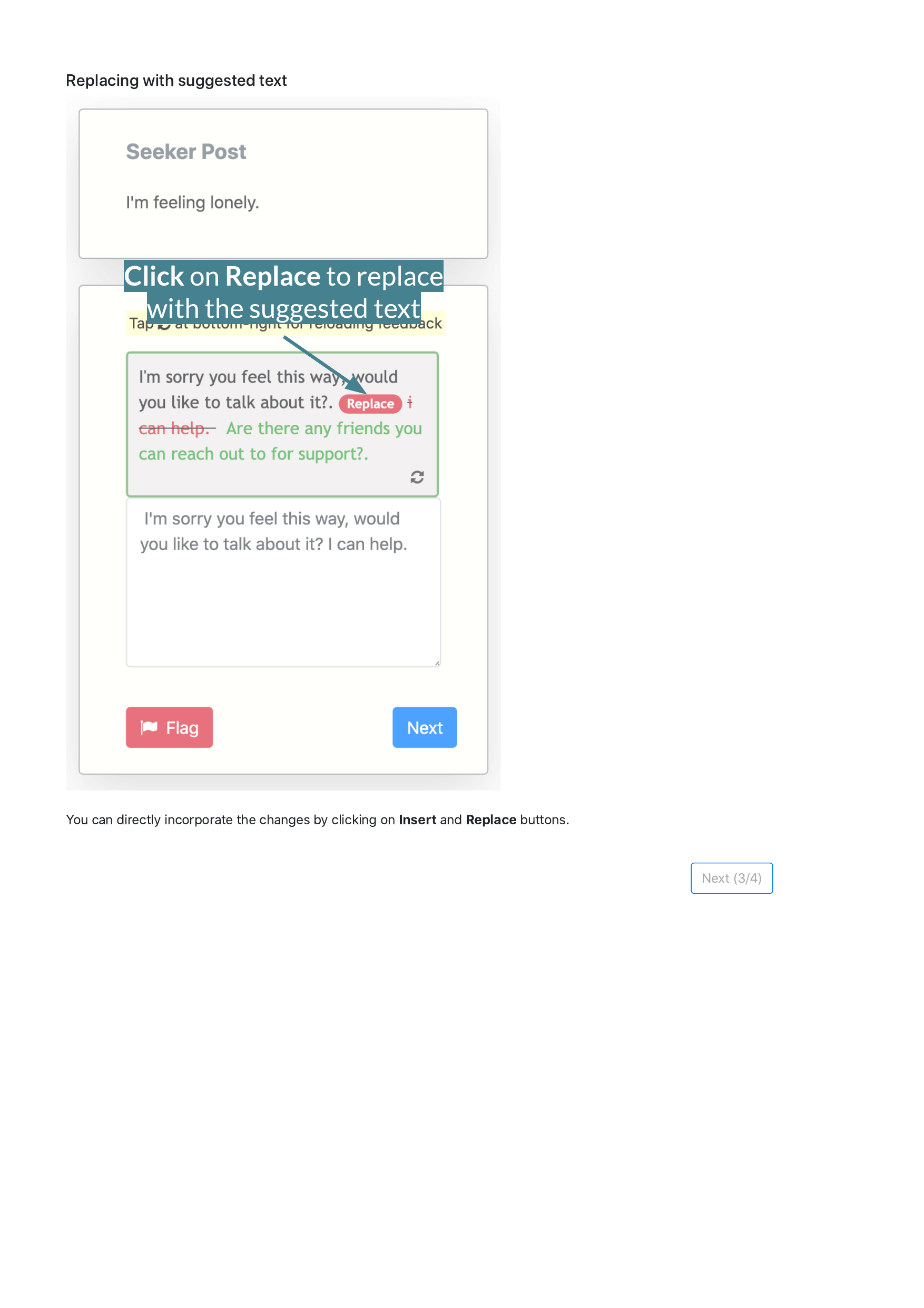}
    \label{supp:figure:study_design:instructions_treatment_5}
\end{figure}

\begin{figure}
    \caption{Instructions shown to the treatment group participants [phase II: empathy training and instructions] (6/6).}
    \centering
         \includegraphics[width=\textwidth]{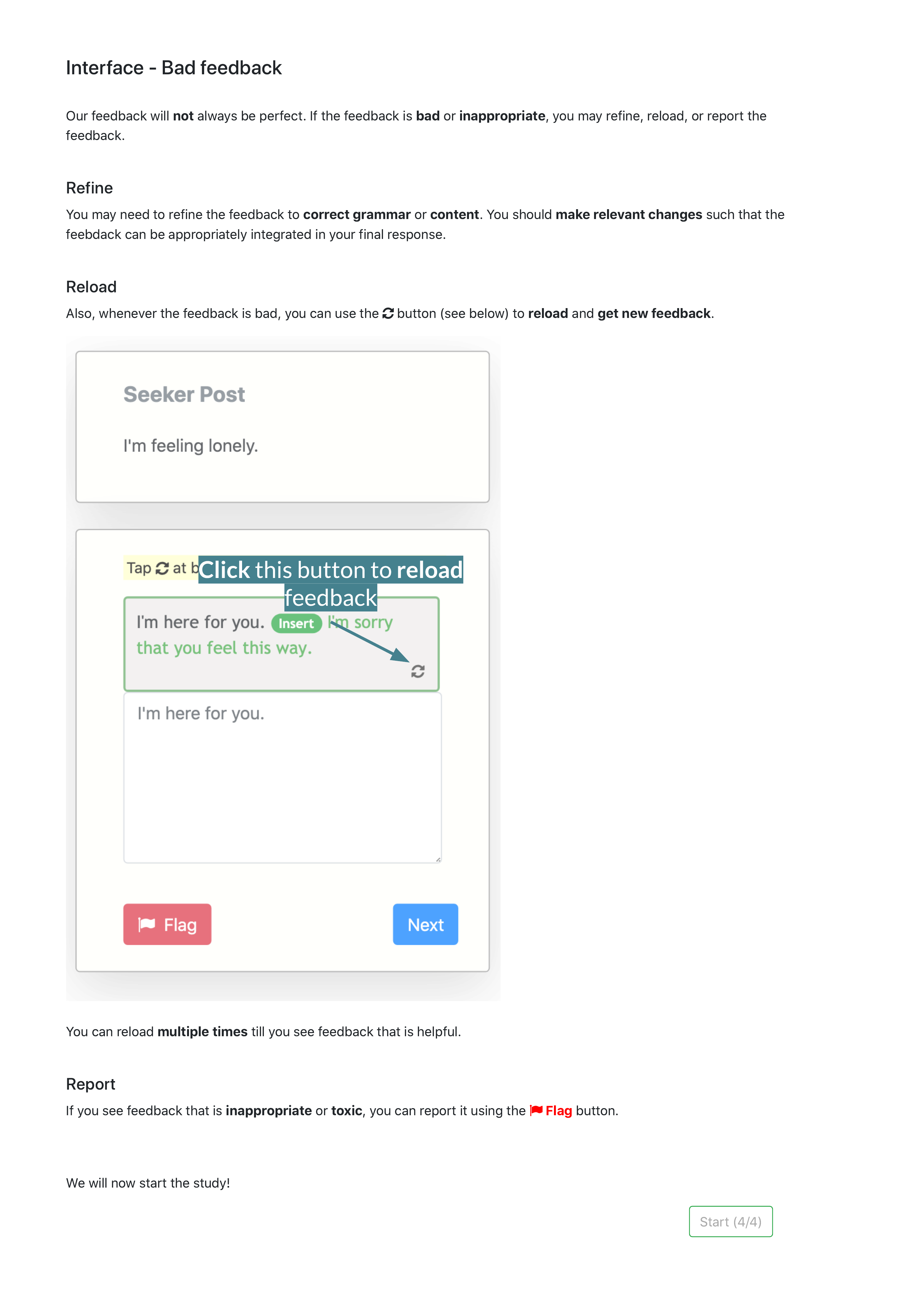}
    \label{supp:figure:study_design:instructions_treatment_6}
\end{figure}

\begin{figure}
\centering
\caption{An example workflow for Human Only (control) participants [phase III: write supportive, empathic responses]. \textbf{(a)} Participant is asked to write a supportive, empathic response. \textbf{(b)} Participant starts writing the response.}
\subfloat[]{
	\includegraphics[width=0.25\columnwidth]{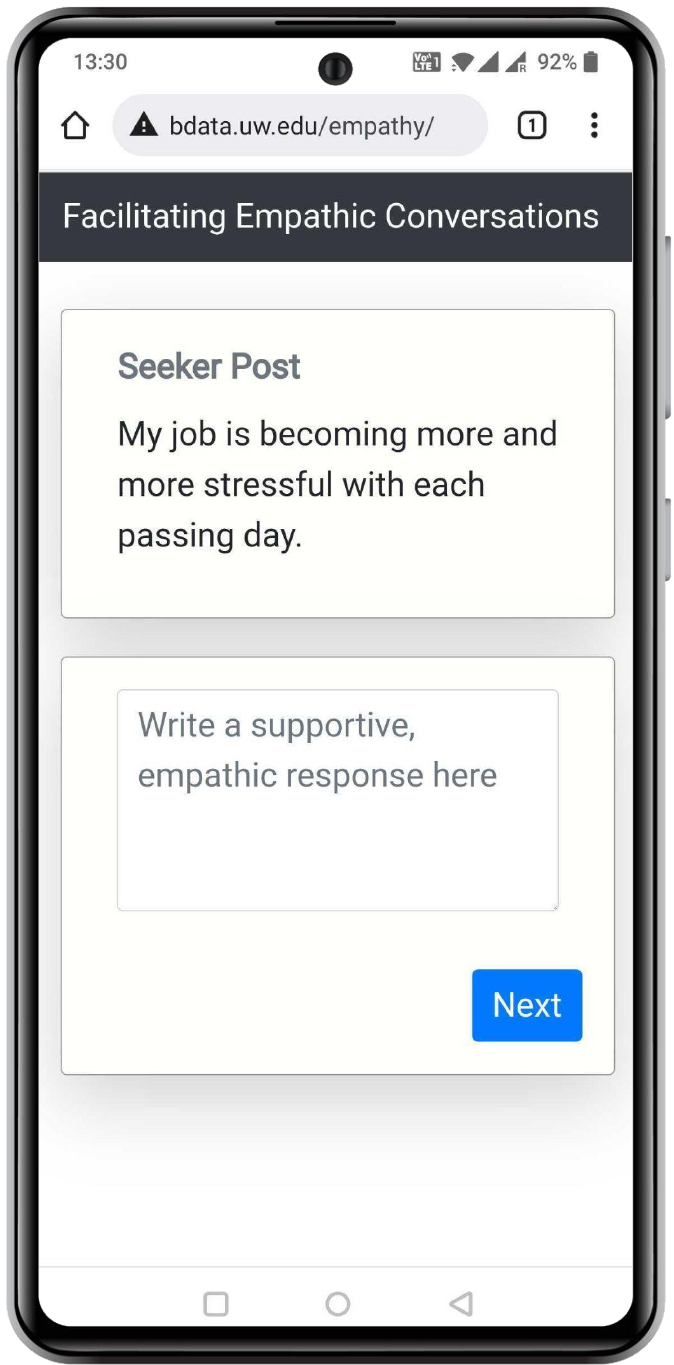} } 
\hspace{50pt}
\subfloat[]{
	\includegraphics[width=0.25\columnwidth]{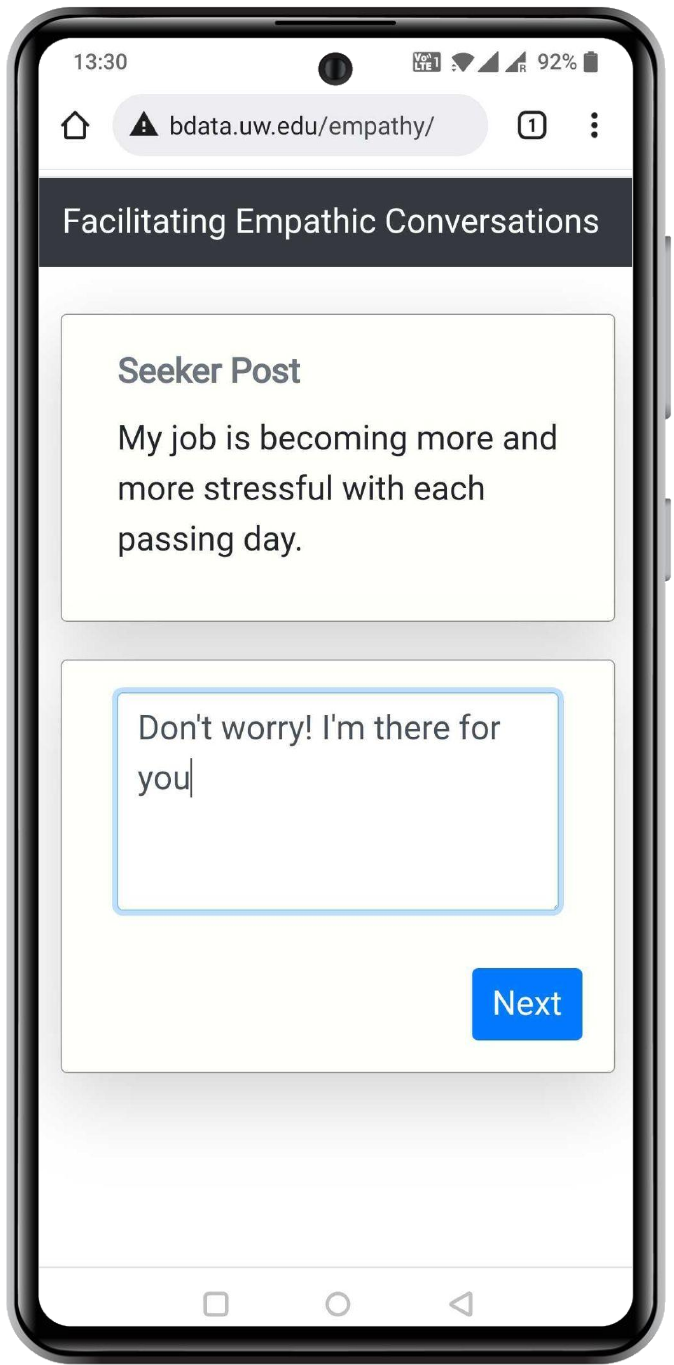} } 
\label{supp:figure:study_design:interface_control}
\end{figure}

\begin{figure}
\centering
\caption{An example workflow for Human + AI (treatment) participants [phase III: write supportive, empathic responses]. \textbf{(a)} Participant is asked to write a supportive, empathic response and given an option to receive feedback. \textbf{(b)} Participant starts writing the response. \textbf{(c)} Participant clicks on the prompt to request feedback from \oursystem. \textbf{(d)} Participant accepts the suggested changes and gets an option to request more feedback. \textbf{(e)} Participant continues editing the response and requests more feedback as needed. \textbf{(f)} When the response is already highly empathic, the participant simply receives a positive feedback.}
\subfloat[]{
	\includegraphics[width=0.25\columnwidth]{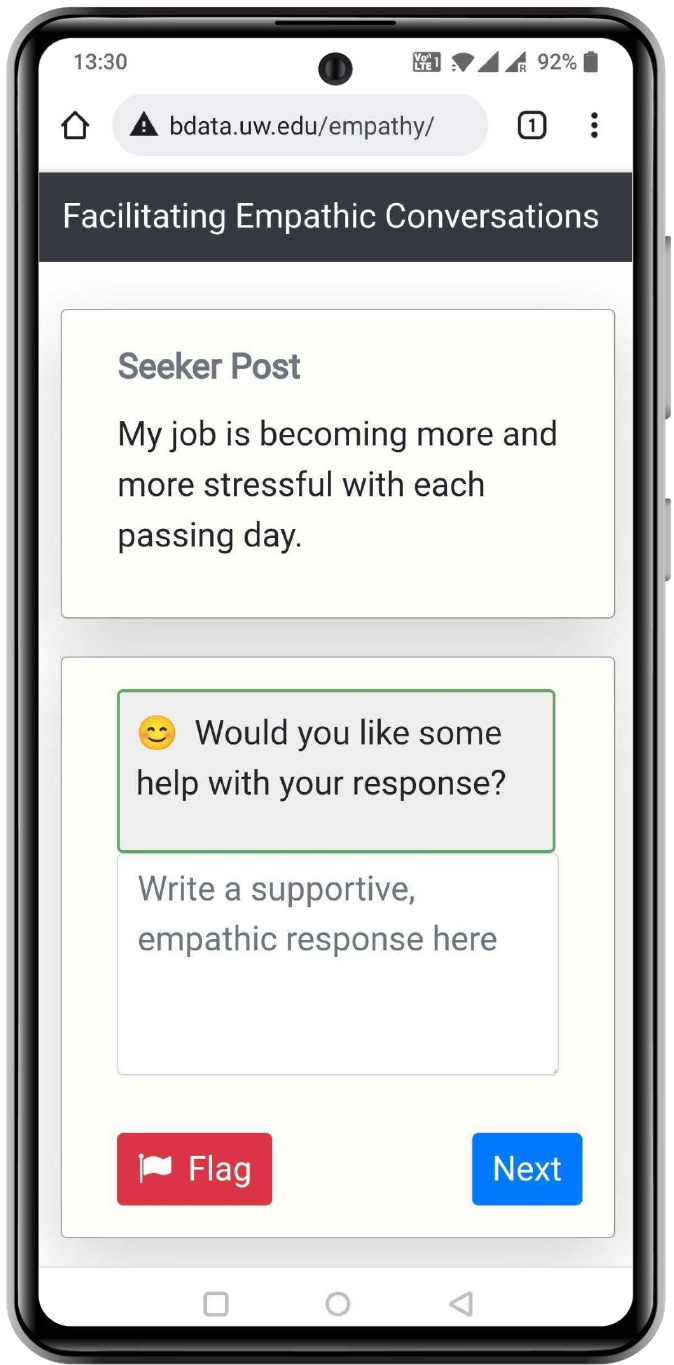} } 
\hfill
\subfloat[]{
	\includegraphics[width=0.25\columnwidth]{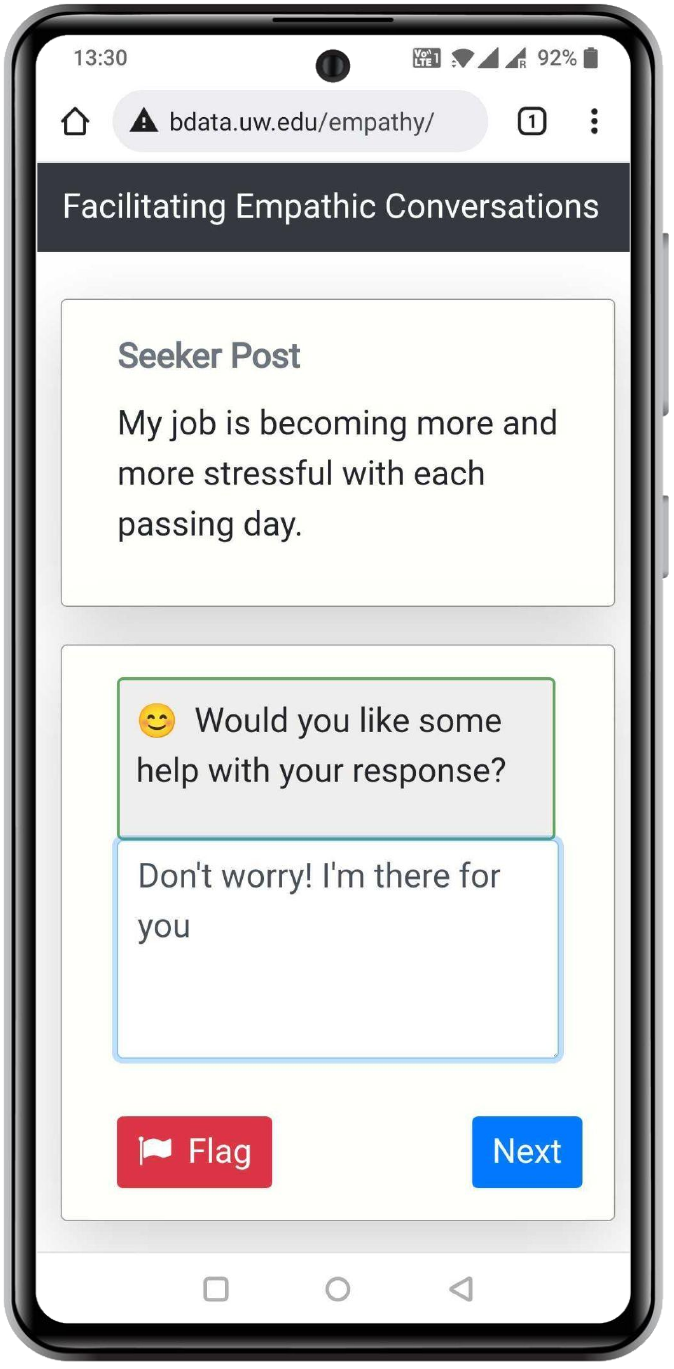} } 
\hfill
\subfloat[]{
	\includegraphics[width=0.25\columnwidth]{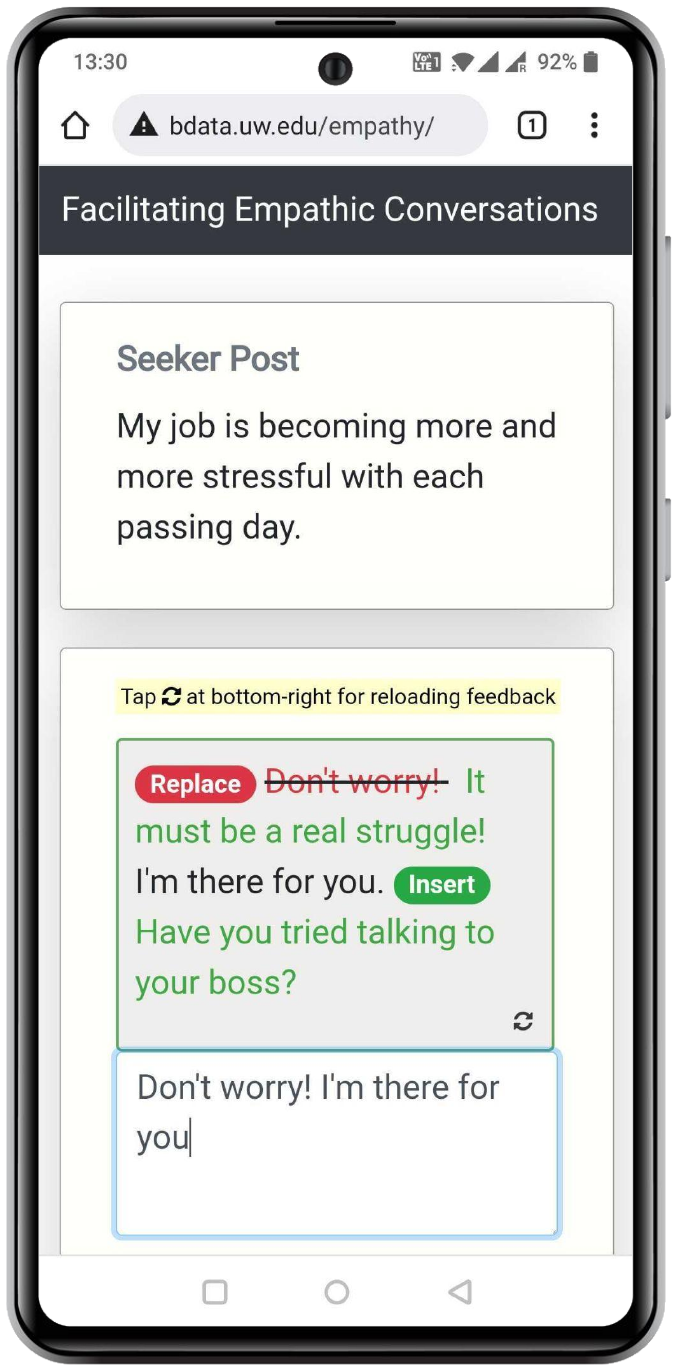} } 
\hfill
\vspace{10pt}
\subfloat[]{
	\includegraphics[width=0.25\columnwidth]{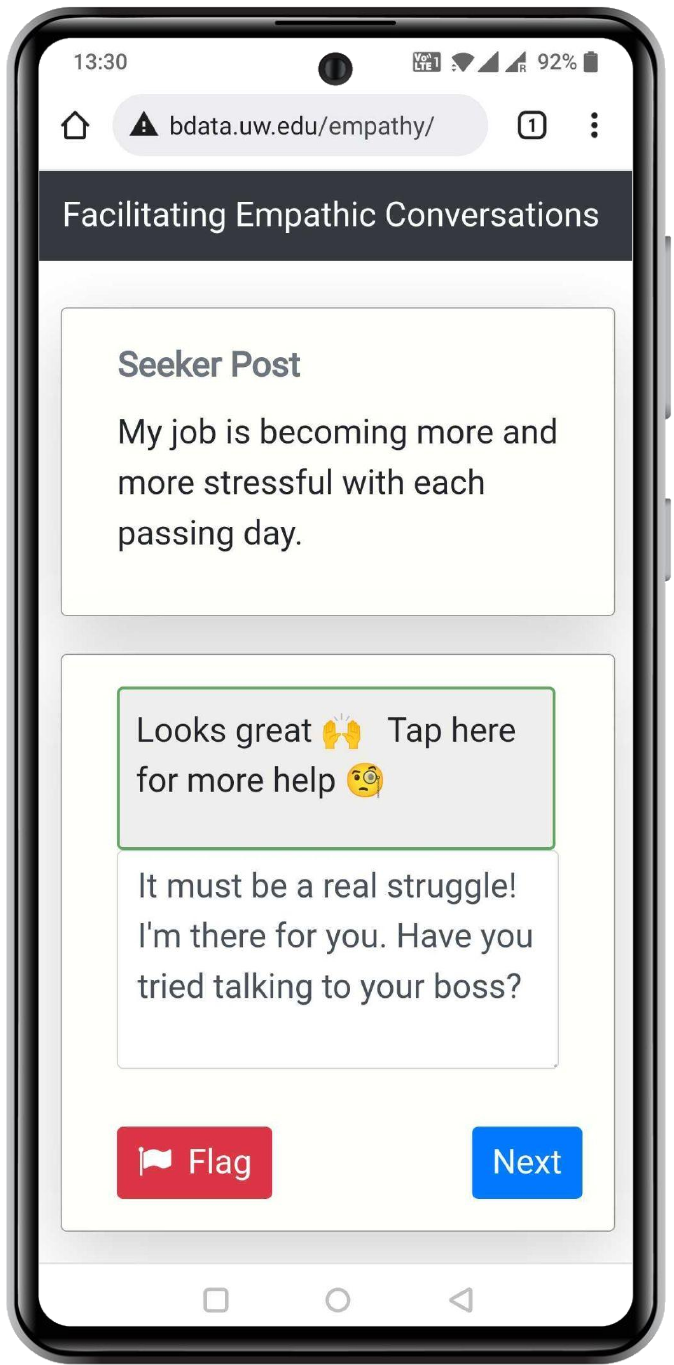} } 
\hfill
\subfloat[]{
	\includegraphics[width=0.25\columnwidth]{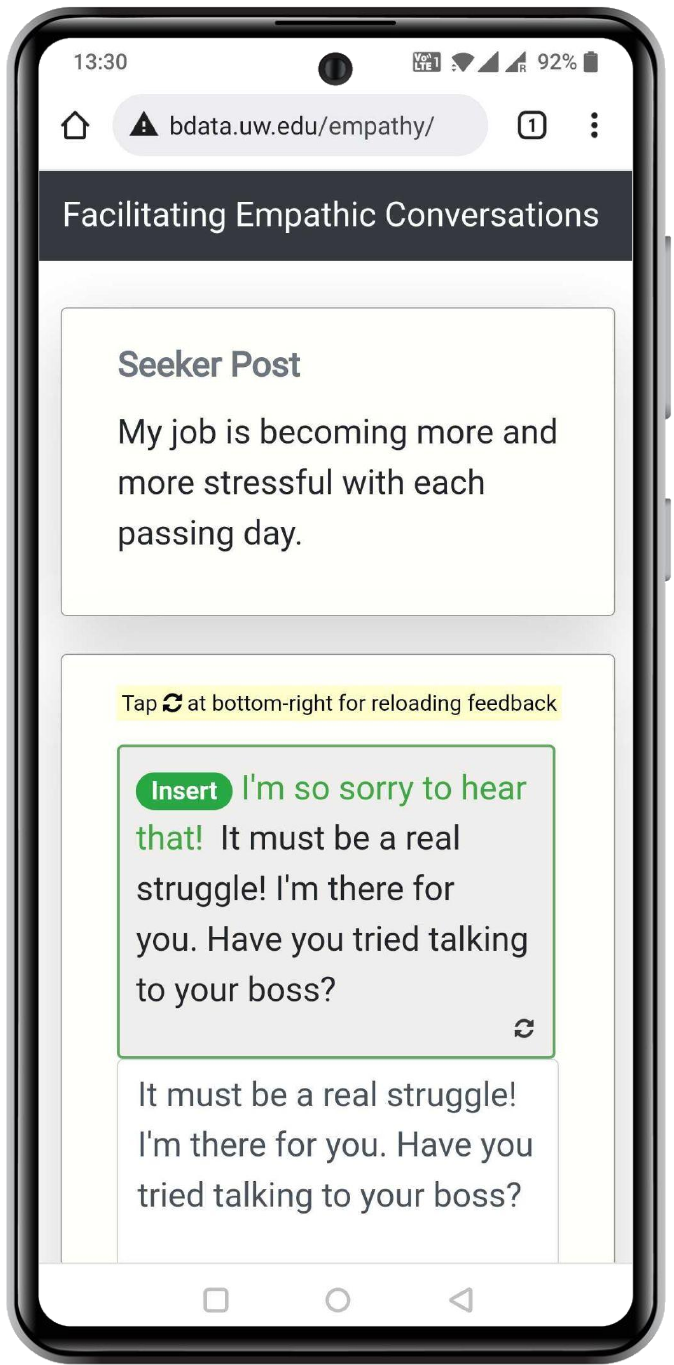} } 
\hfill
\subfloat[]{
	\includegraphics[width=0.25\columnwidth]{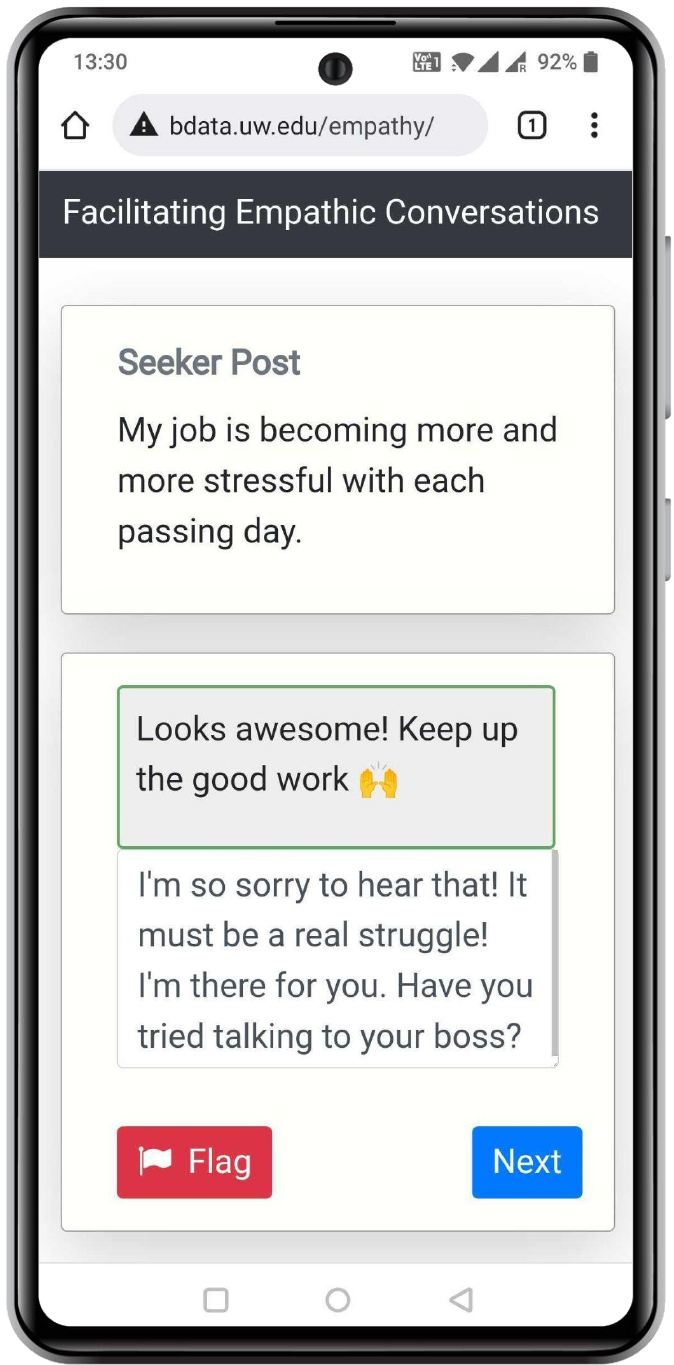} } 
\hfill
\label{supp:figure:study_design:interface_treatment}
\end{figure}

\begin{figure}
    \caption{Interface for flagging feedback [phase III: write supportive, empathic responses].}
    \centering
         \includegraphics[width=0.25\textwidth]{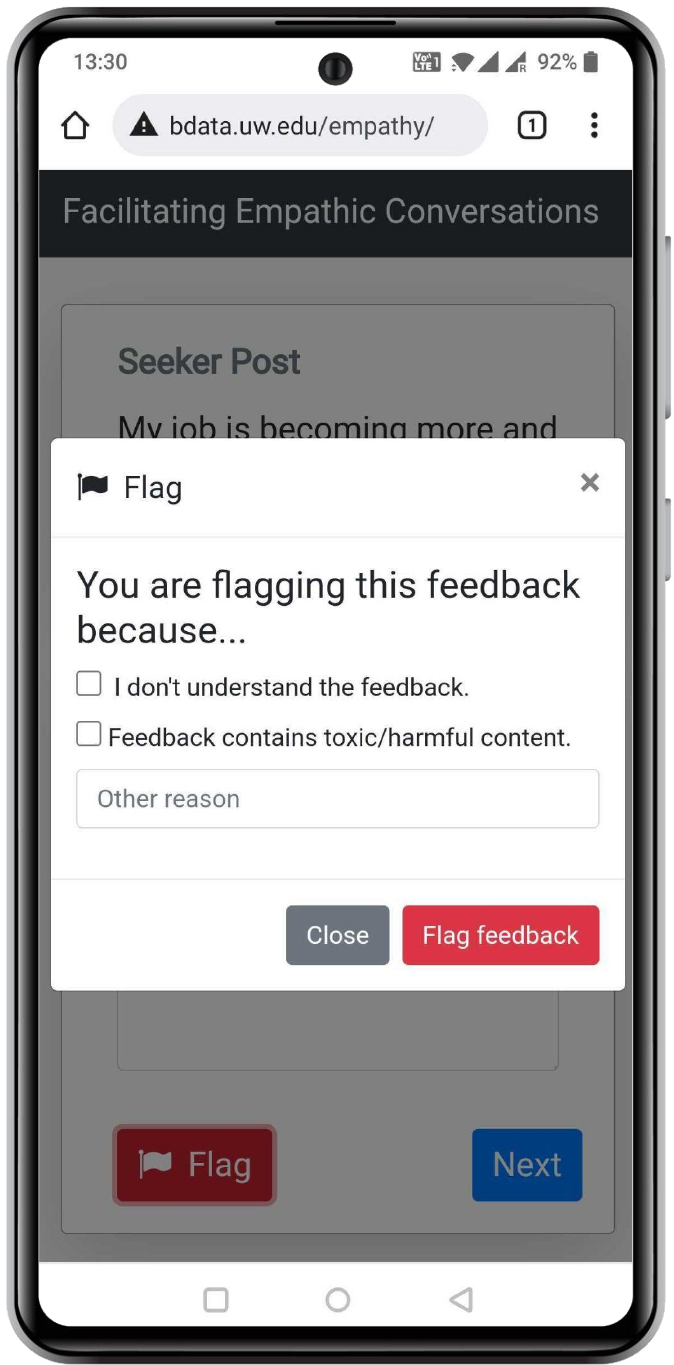}
    \label{supp:figure:study_design:flag}
\end{figure}

\begin{figure}
    \caption{Exit survey used for collecting perceptions of control group participants [phase IV: post-intervention survey].}
    \centering
         \includegraphics[width=\textwidth]{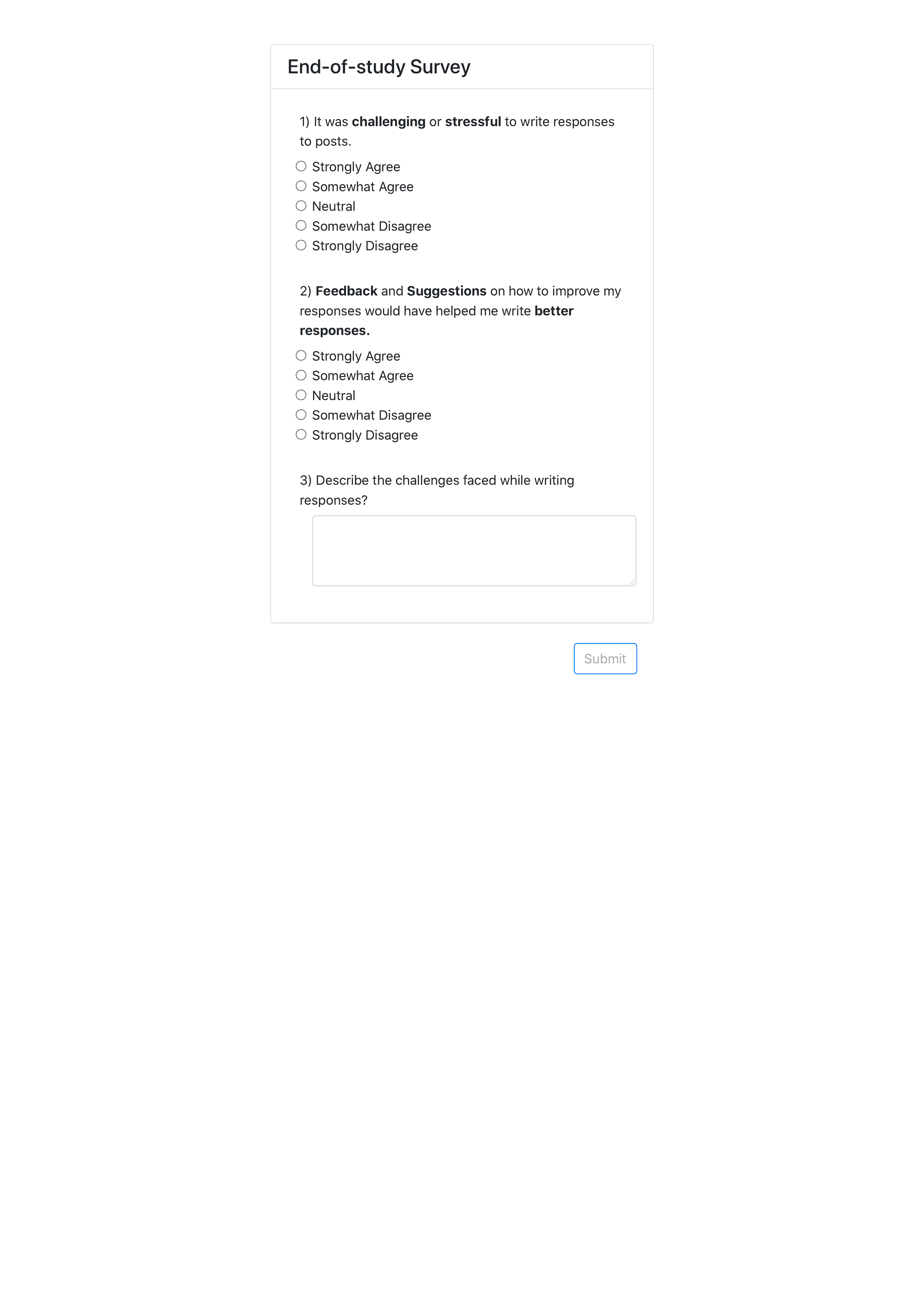}
    \label{supp:figure:study_design:exit_survey_control}
\end{figure}

\begin{figure}
    \caption{Exit survey used for collecting perceptions of treatment group participants [phase IV: post-intervention survey]. Continued on the next page (1/2).}
    \centering
         \includegraphics[page=1,width=\textwidth]{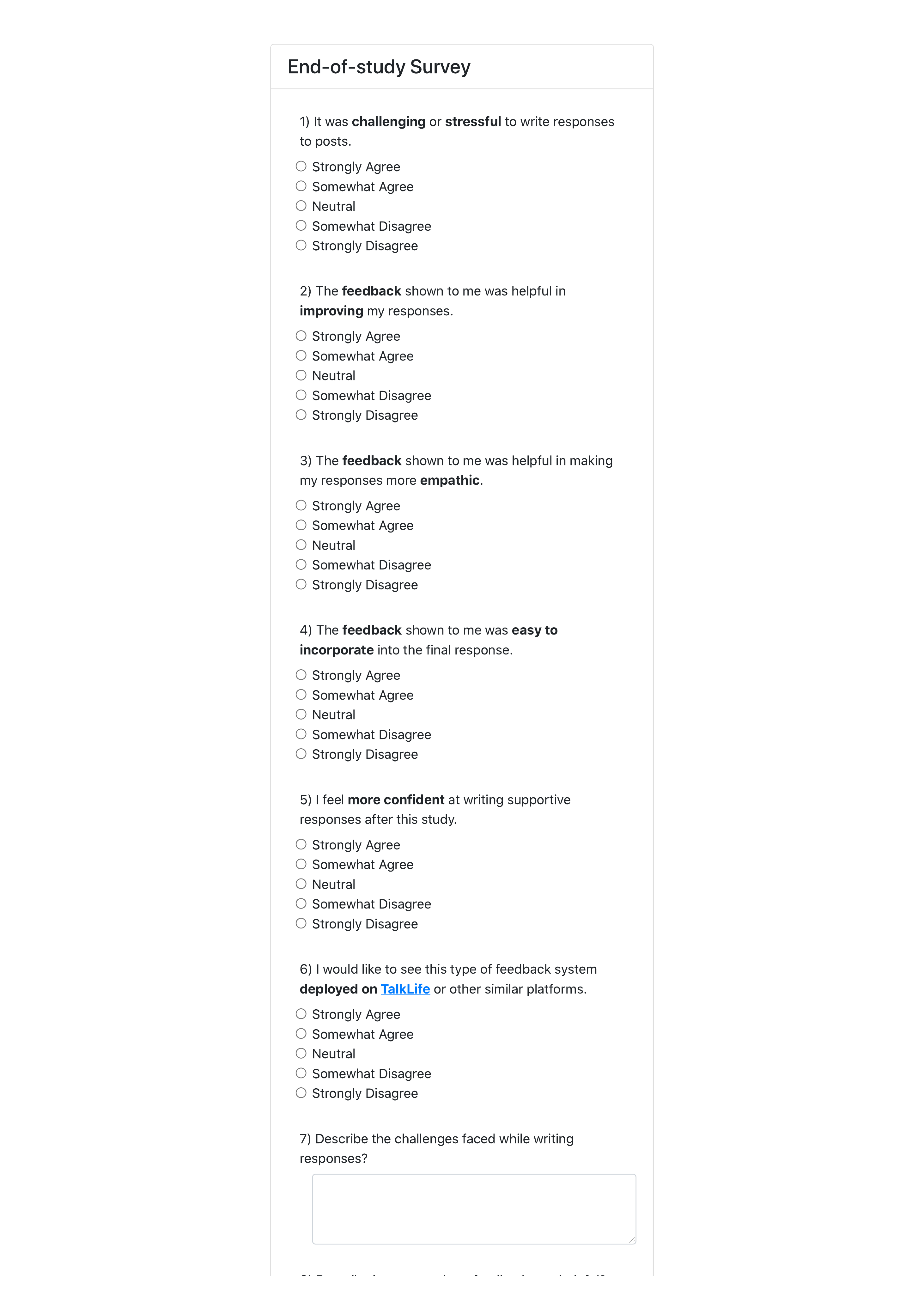}
    \label{supp:figure:study_design:exit_survey_treatment_1}
\end{figure}

\begin{figure}
    \caption{Exit survey used for collecting perceptions of treatment group participants [phase IV: post-intervention survey] (2/2).}
    \centering
         \includegraphics[page=2,width=\textwidth]{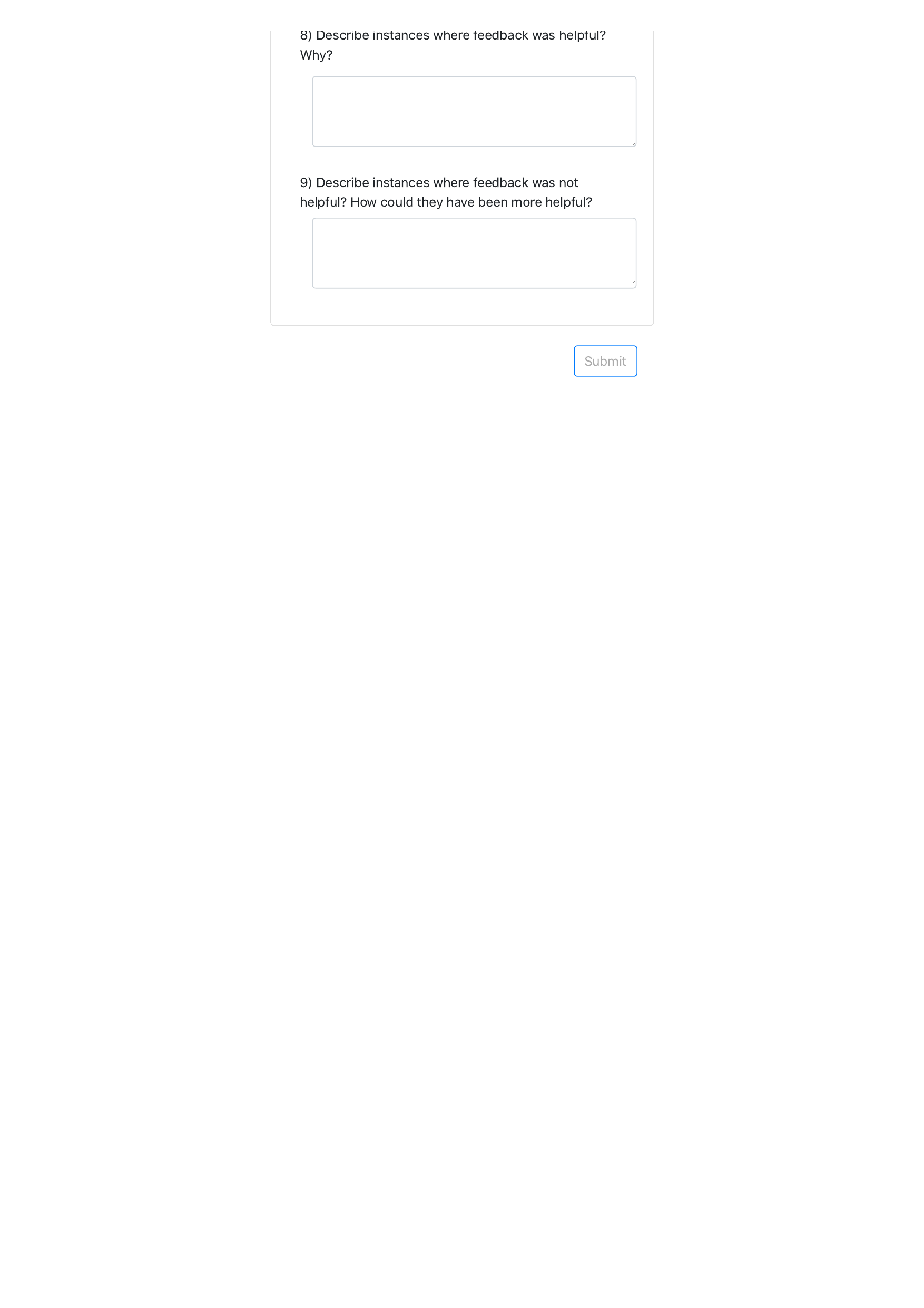}
    \label{supp:figure:study_design:exit_survey_treatment_2}
\end{figure}


%% file: suppFigures/_supp_figure_feedback_eval_design.tex
\begin{figure}
    \caption{Consent form used for human evaluation of responses.}
    \centering
         \includegraphics[width=\textwidth]{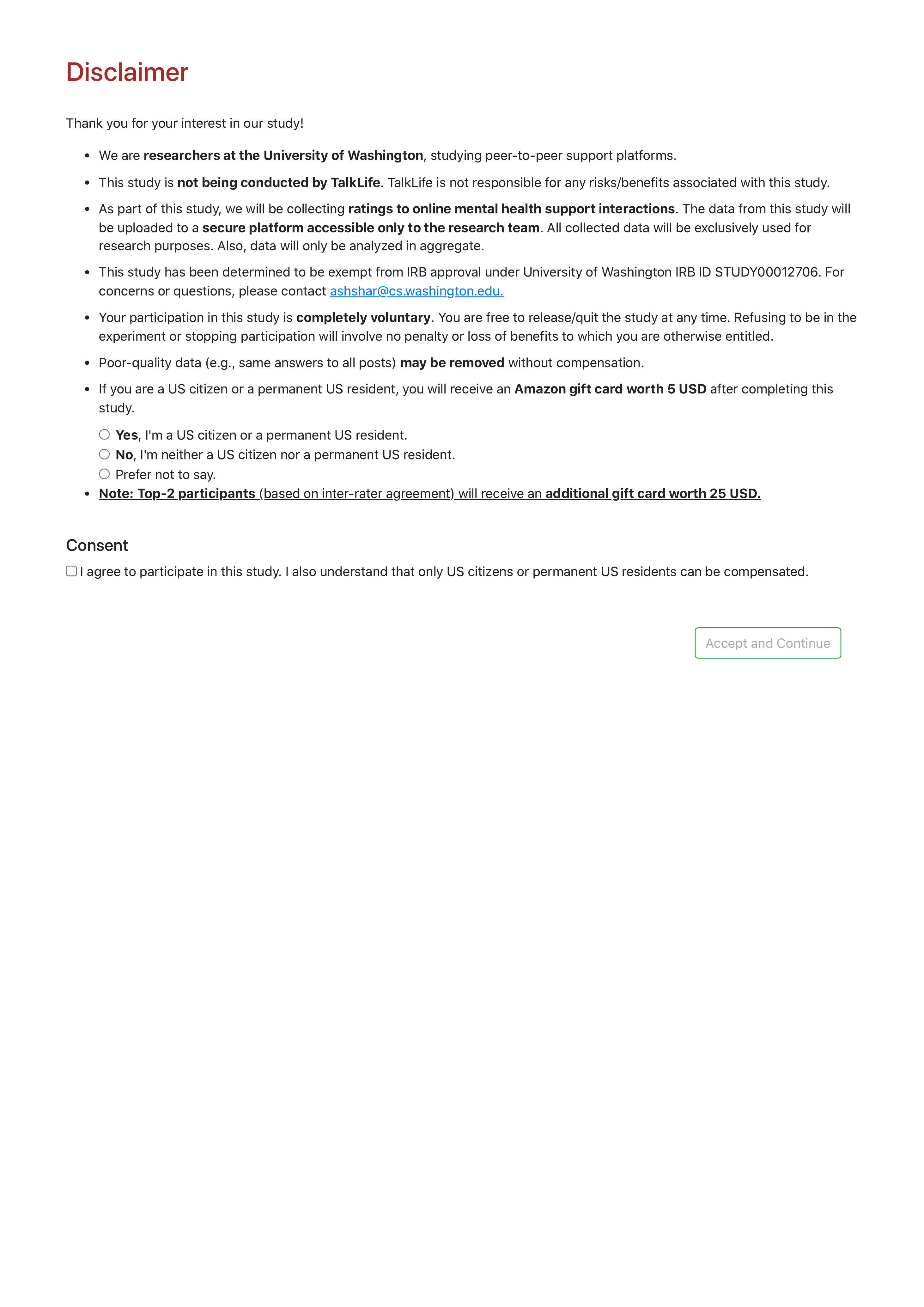}
    \label{supp:figure:study_design:feedback_eval:consent}
\end{figure}

\begin{figure}
    \caption{Instructions for human evaluation of responses.}
    \centering
         \includegraphics[width=\textwidth]{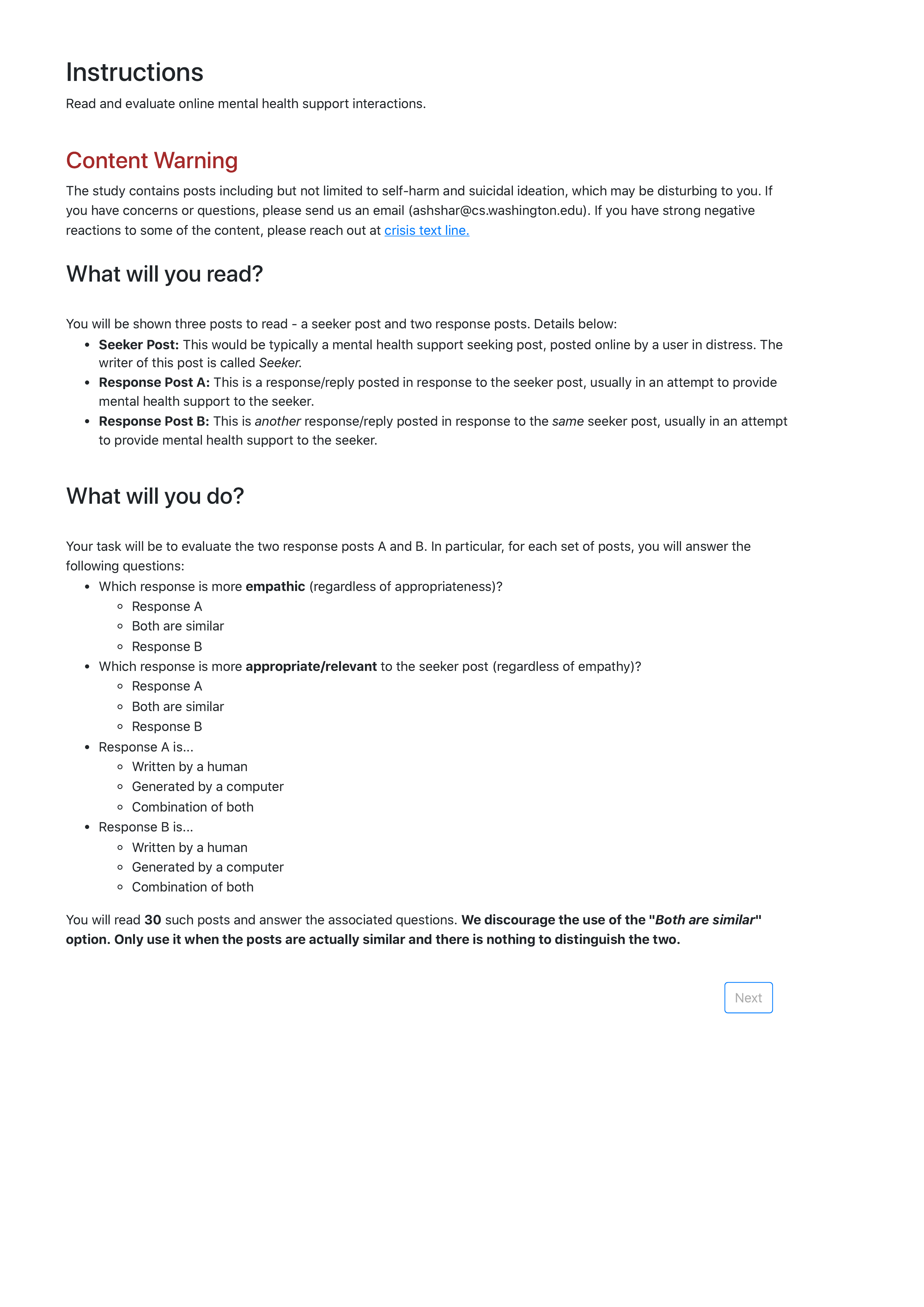}
    \label{supp:figure:study_design:feedback_eval:instructions}
\end{figure}

\begin{figure}
    \caption{Interface for human evaluation of responses.}
    \centering
         \includegraphics[width=\textwidth]{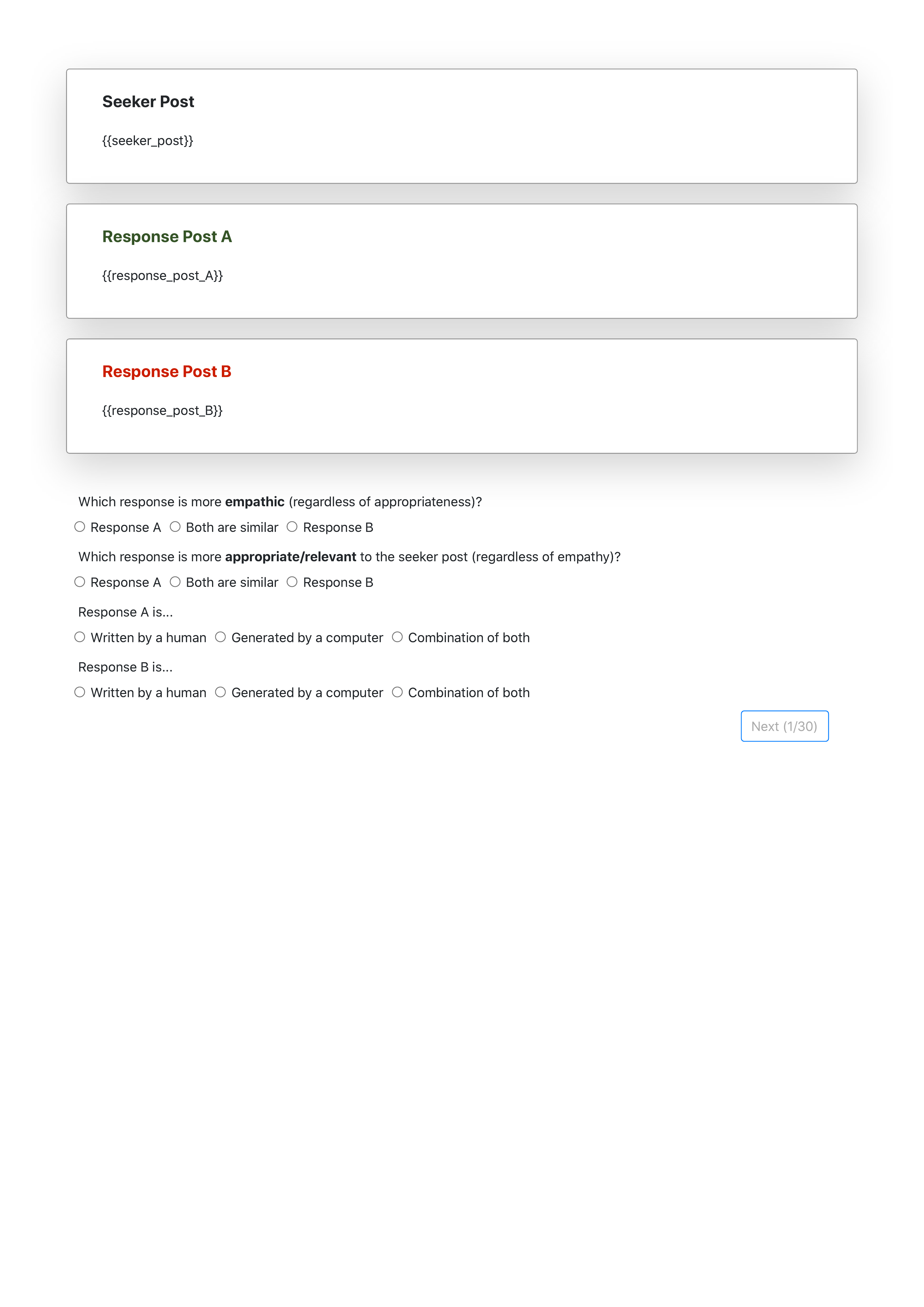}
    \label{supp:figure:study_design:feedback_eval:interface}
\end{figure}

%% file: suppFigures/_supp_figure_empathic_rewriting_model.tex
\begin{figure}
    \caption{An overview of \textsc{Partner}, the deep reinforcement learning model that \oursystem~uses. Figure adapted from Sharma et al.\cite{Sharma2021-rq}}
    \centering
         \includegraphics[width=\textwidth]{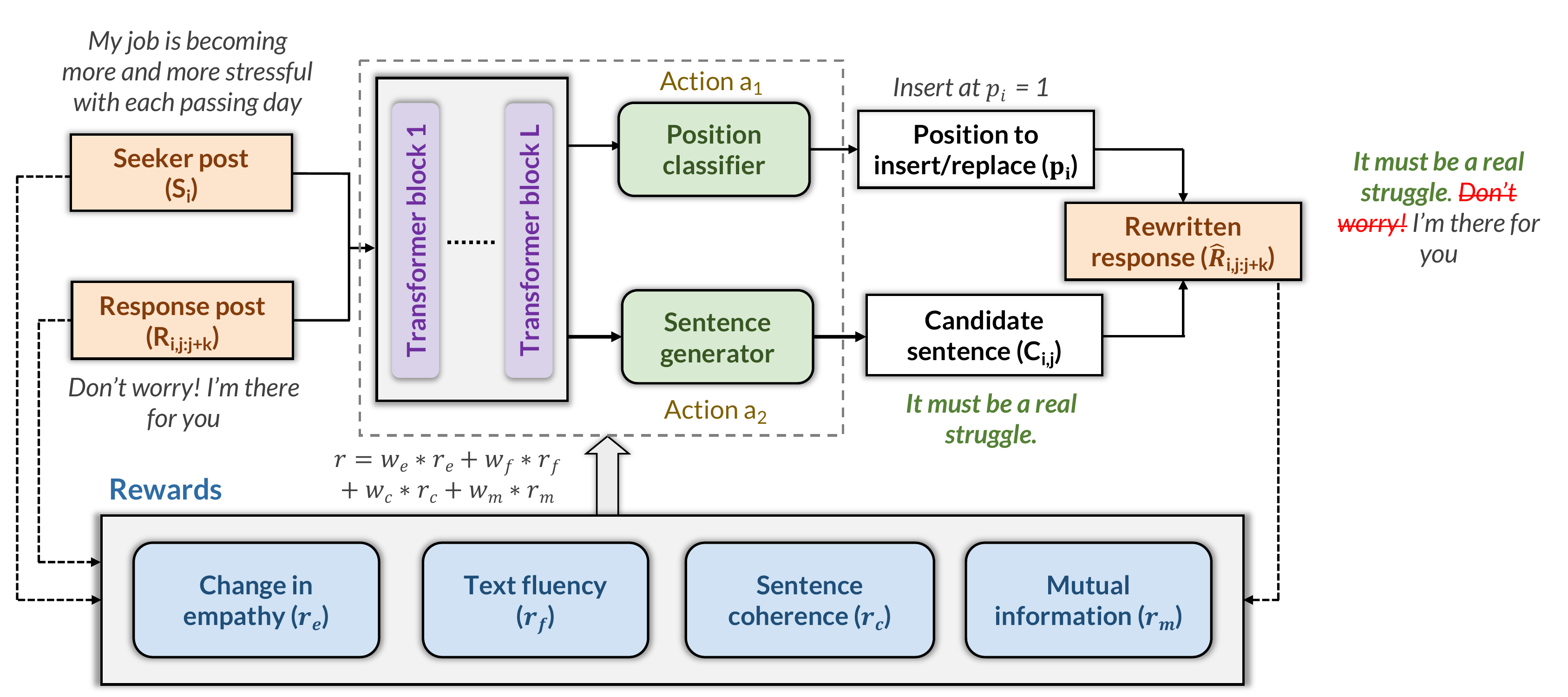}
    \label{supp:figure:empathic_rewriting_model}
\end{figure}